%% file: gradients-jmlr.tex
\documentclass{article}
\pdfoutput=1

\usepackage[arxiv]{jmlr2e2021}

\usepackage{natbib}

\usepackage{bm}
\usepackage{hyperref}       
\usepackage{url}            
\usepackage{booktabs}       
\usepackage{amsfonts}       
\usepackage{nicefrac}       
\usepackage{microtype}      
\usepackage{subfig}
\usepackage{tabularx}
\usepackage{amsmath}
\usepackage{enumerate}
\usepackage{xcolor}
\usepackage{graphicx}
\usepackage[font=small]{caption}
\usepackage[export]{adjustbox}
\usepackage{algorithm}
\usepackage{algorithmic}
\usepackage{float}
\usepackage{placeins}
\usepackage{multirow}

\include{gradients-commands}

\newlength{\llw}
\newlength{\picwi}

\begin{document}

\title{Manifold Coordinates with Physical Meaning}

%

\author{\name Samson J. Koelle${}^1$ \email sjkoelle@uw.edu \\
       \name Hanyu Zhang${}^1$ \email hanyuz6@uw.edu \\
       \name Marina Meil\u{a}${}^{1,2}$ \email mmp@stat.washington.edu \\
       \name Yu-Chia Chen${}^2$ \email yuchaz@uw.edu \\
       \addr ${}^1$Department of Statistics\\
       University of Washington\\
       Seattle, WA 98195-4322, USA   \\
       ${}^2$Department of Electrical and Computer Engineering \\
       University of Washington \\
       Seattle, WA 98195, USA
       }
       
\editor{Francis Bach, David Blei, and Bernhard Sch{\"o}lkopf}

\maketitle
\begin{abstract}
Manifold embedding algorithms map high-dimensional data down to
coordinates in a much lower-dimensional space. One of the aims of
dimension reduction is to find {\em intrinsic coordinates} that
describe the data manifold. The coordinates returned by the
embedding algorithm are abstract, and finding their physical or
domain-related meaning is not formalized and often left to domain
experts. This paper studies the problem of recovering the
meaning of the new low-dimensional representation in an
 automatic, principled fashion.  We propose a method to explain
embedding coordinates of a manifold as {\em non-linear} compositions
of functions from a user-defined dictionary. We show that this problem
can be set up as a sparse {\em linear Group Lasso} recovery problem,
find sufficient recovery conditions, and demonstrate its effectiveness
on data.

\end{abstract}


\input{gradients-jmlr-intro}

\input{gradients-jmlr-problem}
\input{gradients-jmlr-background}

\input{gradients-jmlr-pullback}

\input{gradients-jmlr-related}

\input{gradients-jmlr-theory}

\input{gradients-jmlr-exp}
\input{gradients-jmlr-conclusion}

\medskip
\section*{Acknowledgement}
Samson Koelle was funded by NSF DMS 2015272, NSF DMS 1810975.  Hanyu Zhang was funded by NSF DMS 2015272, NSF DMS 1810975  Yu-Chia Chen was funded by NSF DMS 2015272 and the U.S. Department of
Energy, Solar Energy Technology Office award DE-EE0008563.
\begin{small}
\bibliography{mmp,materials,gradients,randos,manifolds2}
\bibliography{mmp}
\end{small}

\input{gradients-jmlr-appendix}

\end{document}

%% file: gradients-commands.tex
\newenvironment{itemize*}{
\begin{itemize}
\setlength{\parskip}{0em}
\setlength{\topparskip}{0em}
}
{\end{itemize}}

\newenvironment{enumerate*}{
\begin{enumerate}
\setlength{\parskip}{0em}
\setlength{\topparskip}{0em}
}
{\end{enumerate}}

\newcommand{\comment}[1]{}

\newcommand{\mmp}[1]{}
\newcommand{\hanyuz}[1]{}
\newcommand{\skcomment}[1]{}

\newcommand{\beq}{\begin{equation}}
\newcommand{\eeq}{\end{equation}}
\newcommand{\beqa}{\begin{eqnarray}}
\newcommand{\eeqa}{\end{eqnarray}}
\newcommand{\beqas}{\begin{eqnarray*}}
\newcommand{\eeqas}{\end{eqnarray*}}

\newcommand{\bit}{\begin{itemize}}
\newcommand{\eit}{\end{itemize}}
\newcommand{\bits}{\begin{itemize*}}
\newcommand{\eits}{\end{itemize*}}
\newcommand{\benum}{\begin{enumerate}}
\newcommand{\eenum}{\end{enumerate}}
\newcommand{\benums}{\begin{enumerate*}}
\newcommand{\eenums}{\end{enumerate*}}

\newcommand{\dpullalg}{{\sc PullBackDPhi}}

\newcommand{\glassoalg}{{\sc GroupLasso}}
\newcommand{\embedalg}{{\sc EmbeddingAlg}}
\newcommand{\flassoalg}{{\sc GroupLasso}}
\newcommand{\flasso}{{\sc GroupLasso}}
\newcommand{\svd}{{\sc SVD}} 

\newcommand{\rmalg}{{\sc RMetric}}
\newcommand{\lapalg}{{\sc Laplacian}}
\newcommand{\tppalg}{{\sc LocalPCA}}
\newcommand{\ouralg}{{\sc ManifoldLasso}}

\newcommand{\srdata}{{\bf SwissRoll}}
\newcommand{\redata}{{\bf RigidEthanol}}
\newcommand{\ethdata}{{\bf Ethanol}}
\newcommand{\toldata}{{\bf Toluene}}
\newcommand{\maldata}{{\bf Malonaldehyde}}

\newcommand{\fracpartial}[2]{\frac{\partial #1}{\partial  #2}}

\newcommand{\bigOO}{{\cal O}}
\newcommand{\dataset}{{\cal D}}

\newcommand{\rrr}{{\mathbb R}}

\newcommand{\M}{{\cal M}}
\newcommand{\T}{{\cal T}}
\newcommand{\txim}{\ensuremath{\T_{\xi_i}\M}} 
\newcommand{\tphim}{\ensuremath{\T_{\phi(\xi_i)}\phi(\M)}} 
\newcommand{\I}{{\cal I}} 

\newcommand{\G}{{\cal G}} 
\newcommand{\neigh}{{\cal N}}

\newcommand{\xb}{\mathbf{X}}  

\newcommand{\g}{\mathbf{g}} 
\newcommand{\id}{\mathbf{id}} 

\newcommand{\proj}{\operatorname{Proj}}

\newcommand{\rank}{\operatorname{rank}}
\newcommand{\range}{\operatorname{span}}
\newcommand{\rowrange}{\operatorname{rowspan}}
\newcommand{\diag}{\operatorname{diag}}
\newcommand{\grad}{\operatorname{grad}}
\newcommand{\supp}{\operatorname{supp}}

\newcommand{\barbx}{\bar{\bar{X}}}
\newcommand{\bhat}{\hat{\beta}}
\newcommand{\zhat}{\hat{z}}
\newcommand{\barw}{\bar{\epsilon}}
\newcommand{\bbar}{\bar{\beta}}
\newcommand{\gammanorm}{\zeta}  
\newcommand{\gammamax}{\gamma_{\rm max}} 



%% file: gradients-jmlr-intro.tex
Manifold learning (ML) algorithms, also known as embedding or unsupervised learning algorithms, map data from high or infinite-dimensional spaces to coordinates of a much lower-dimensional space.
In the sciences, one of the motivating goals of dimension reduction is the discovery of descriptors of the data generating process.
Linear dimension reduction algorithms like Principal Component Analysis (PCA) and non-linear algorithms such as Diffusion Maps \citep{coifman:06} are used in applications from genomics to astronomy to uncover the variables describing large-scale properties of the interrogated system.

\begin{figure}[htb]
\subfloat[][Toluene]{\includegraphics[width=4cm, height=4cm, trim={-.5cm, -.5cm,
-.5cm, -.5cm}, clip]{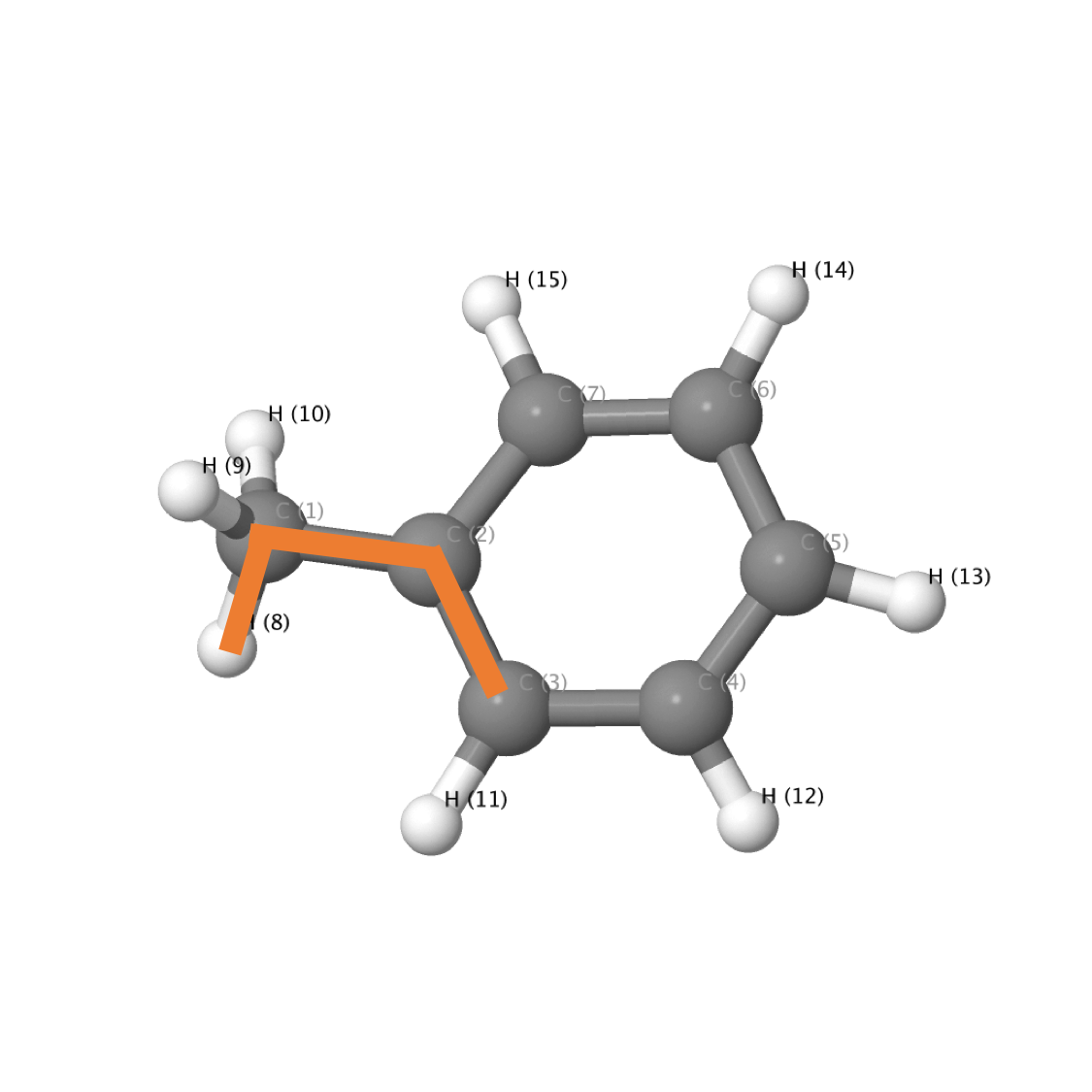}\label{fig:toluene-bonds}}\hfill
\subfloat[][Ethanol]{\includegraphics[width=4cm, height=4cm, trim={-1cm, -1cm,
-1cm, -1cm}, clip]{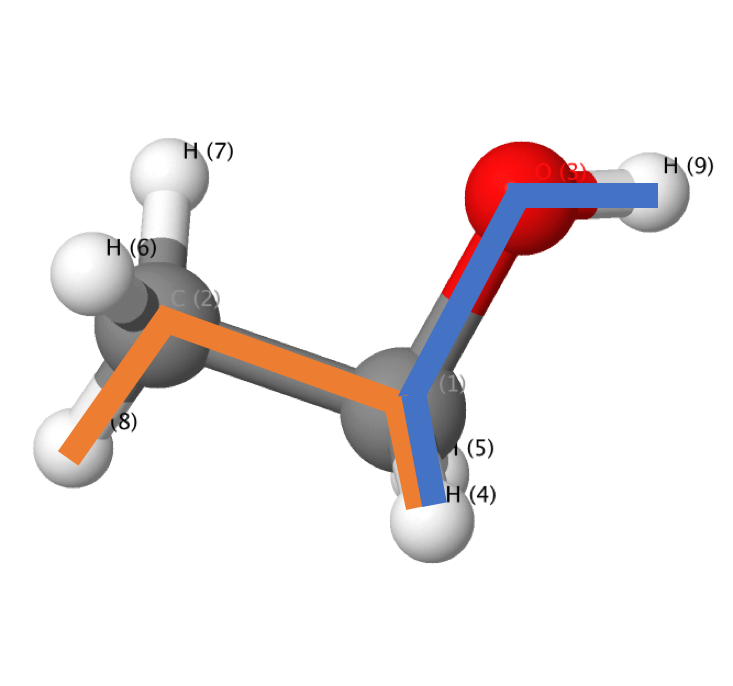}\label{fig:ethanol-bonds}}\hfill
\subfloat[][Malonaldehyde]{\includegraphics[width=4cm, trim={0, -3cm, 0, 0},
clip]{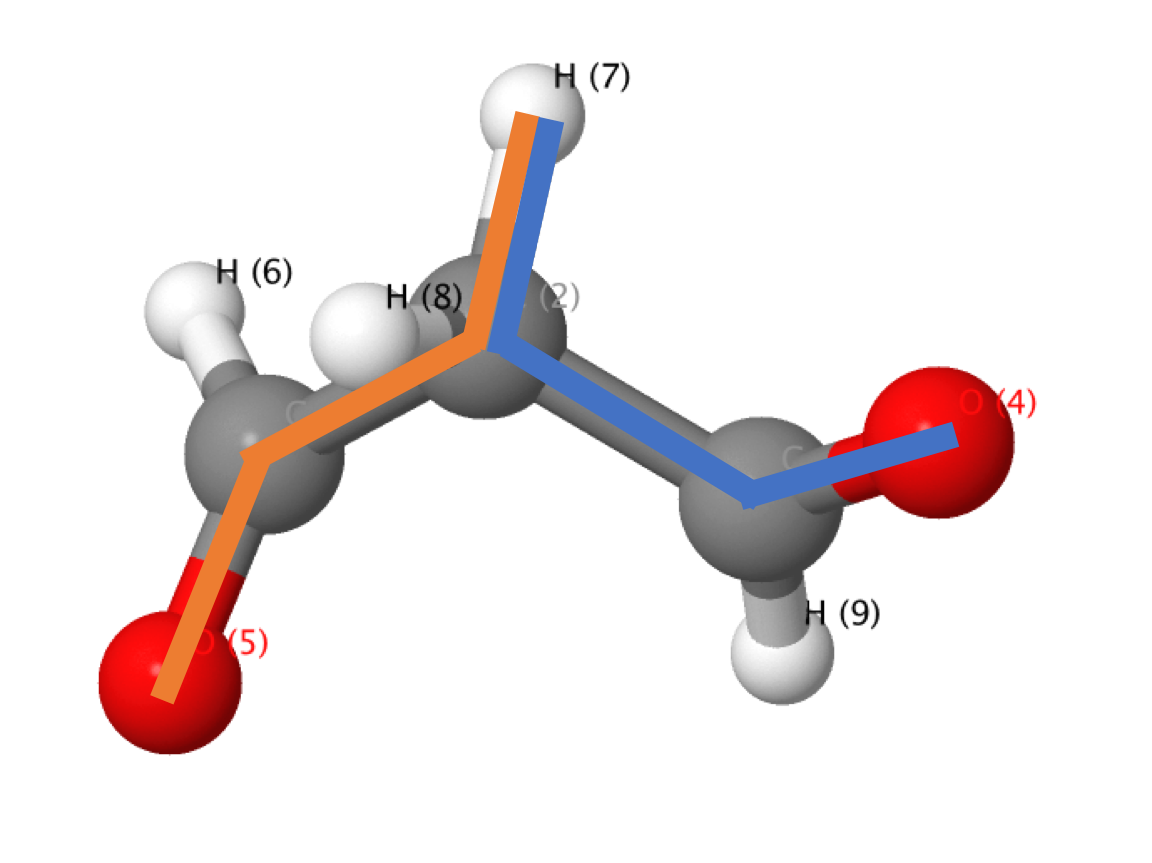}\label{fig:malonaldehyde-bonds}}\hfill \newline
\subfloat[]{\includegraphics[width=4cm,  trim={-1cm, -1cm,
-1cm, -1cm}, clip]{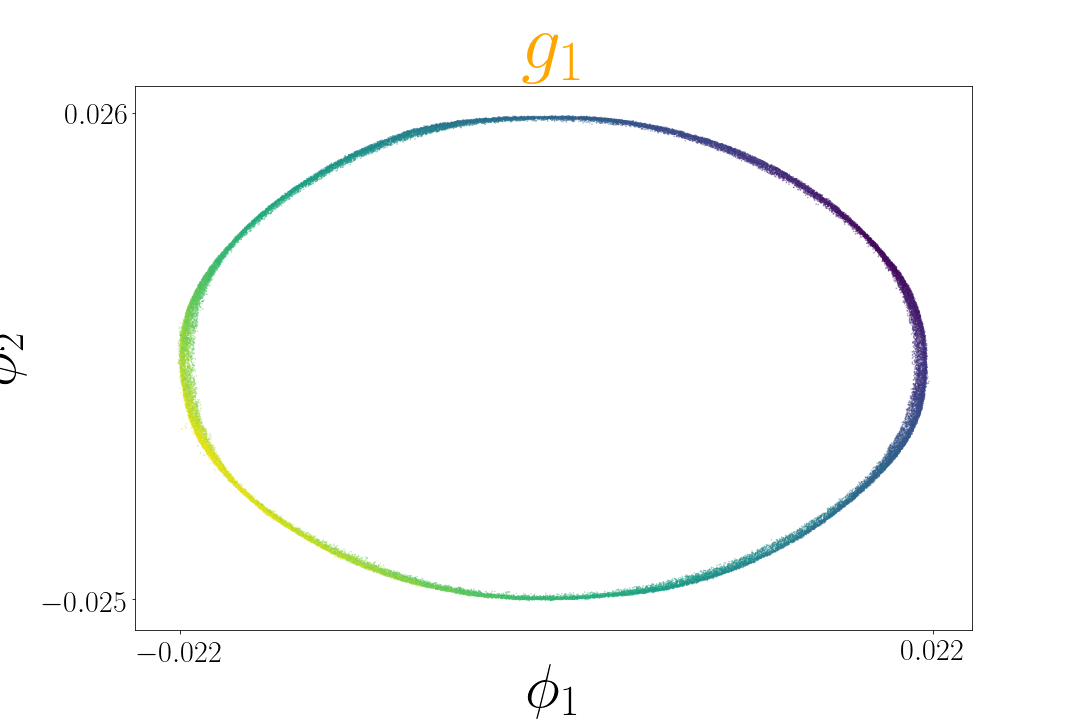}\label{fig:tol_circ}}\hfill
\subfloat[]{\includegraphics[width=5cm, trim={-1cm, -1cm,
-1cm, -1cm}, clip]{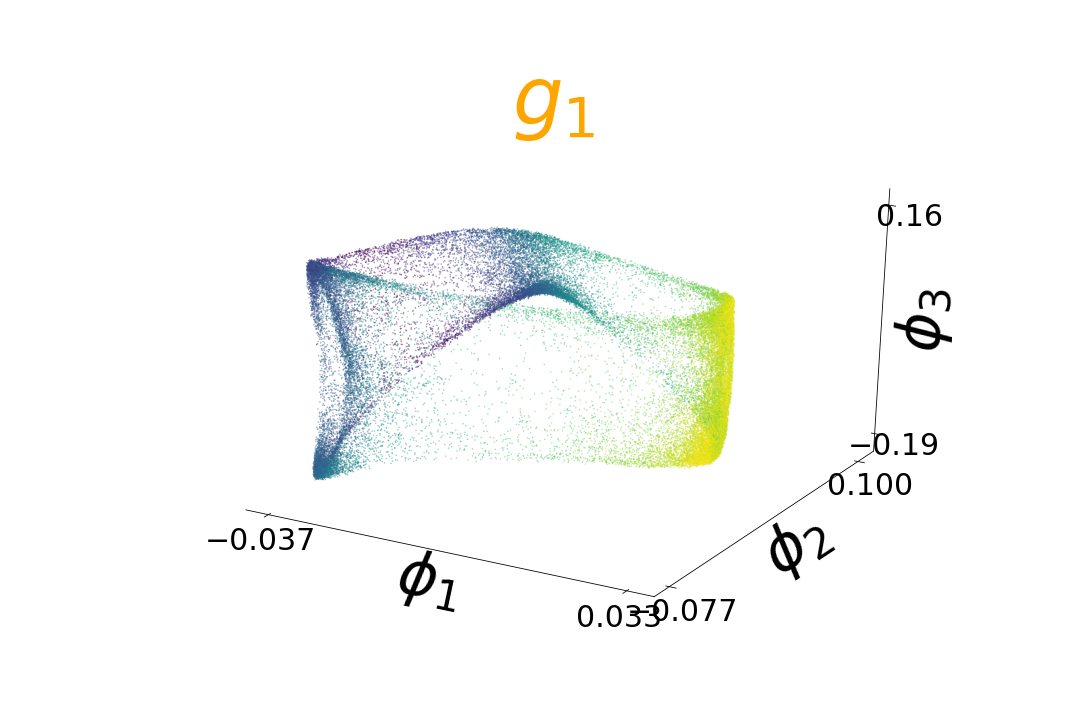}\label{fig:eth_tor1}}\hfill
\subfloat[]{\includegraphics[width=5cm, trim={.5cm, .5cm,
.5cm, .5cm},
clip]{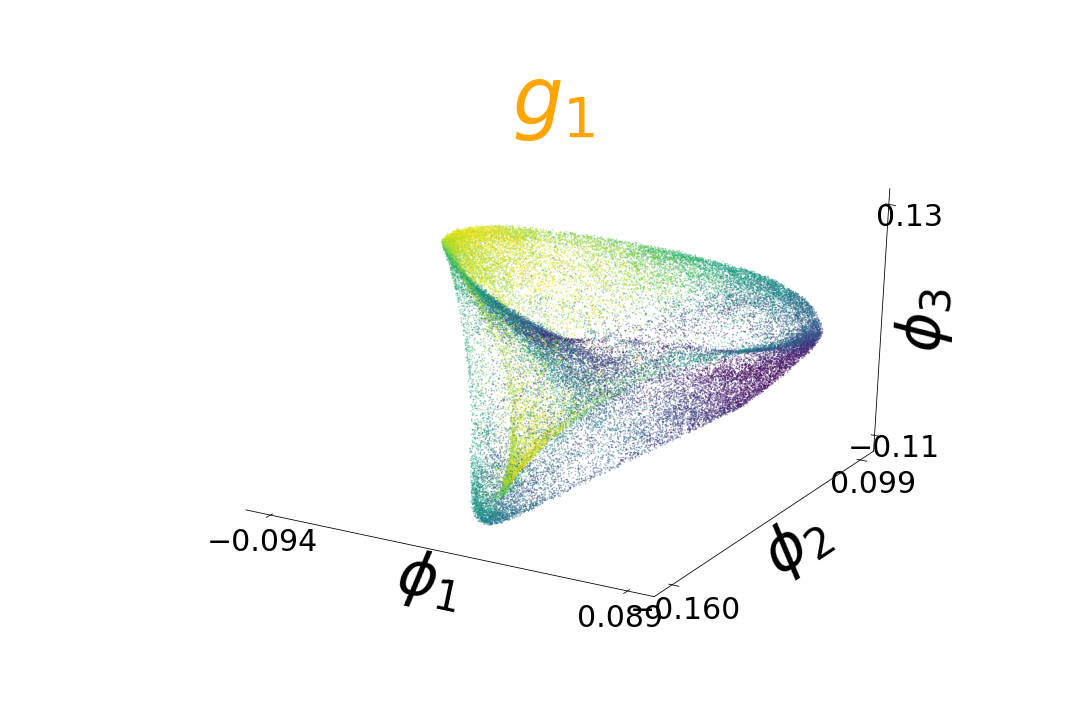}\label{fig:mal_tor1}}\hfill \newline
\subfloat[][Torsion example]{\includegraphics[width=3.5cm, trim={.5cm, .5cm,
.5cm, .5cm},
clip]{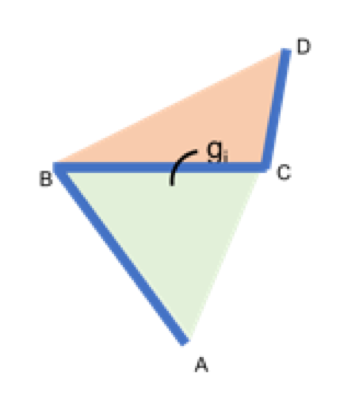}\label{fig:tor_explain}}\hfill 
\subfloat[]{\includegraphics[width=5cm, height=4cm, trim={-1cm, -1cm,
-1cm, -1cm}, clip]{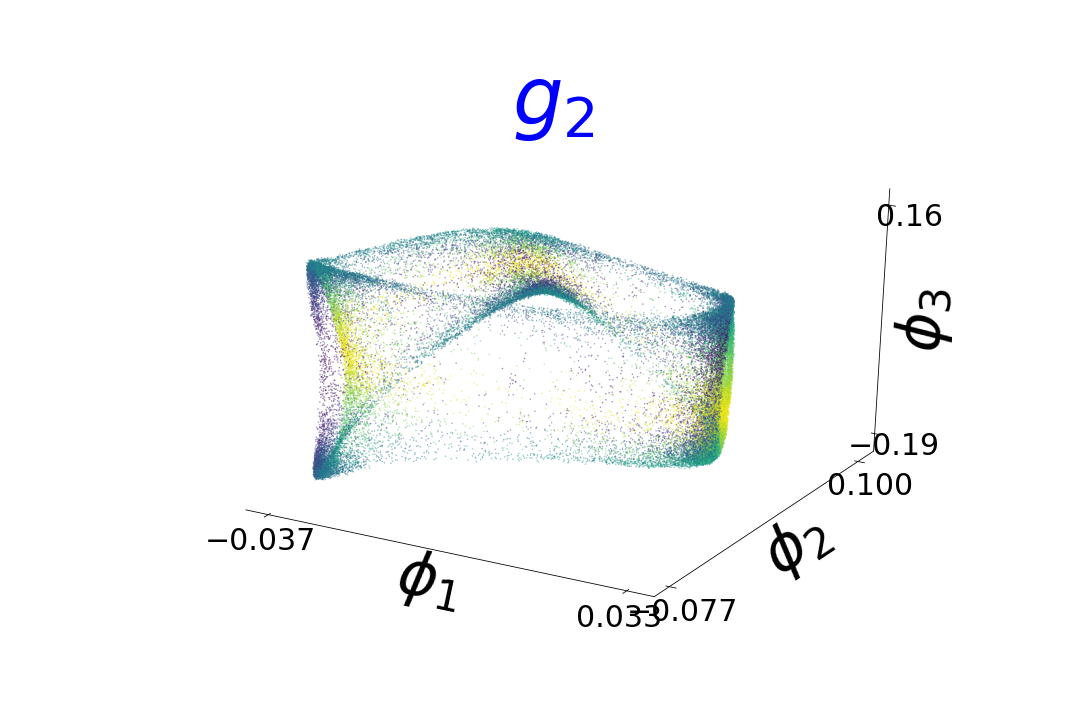}\label{fig:eth_tor2}}\hfill
\subfloat[]{\includegraphics[width=5cm, trim={0, -3cm, 0, 0},
clip]{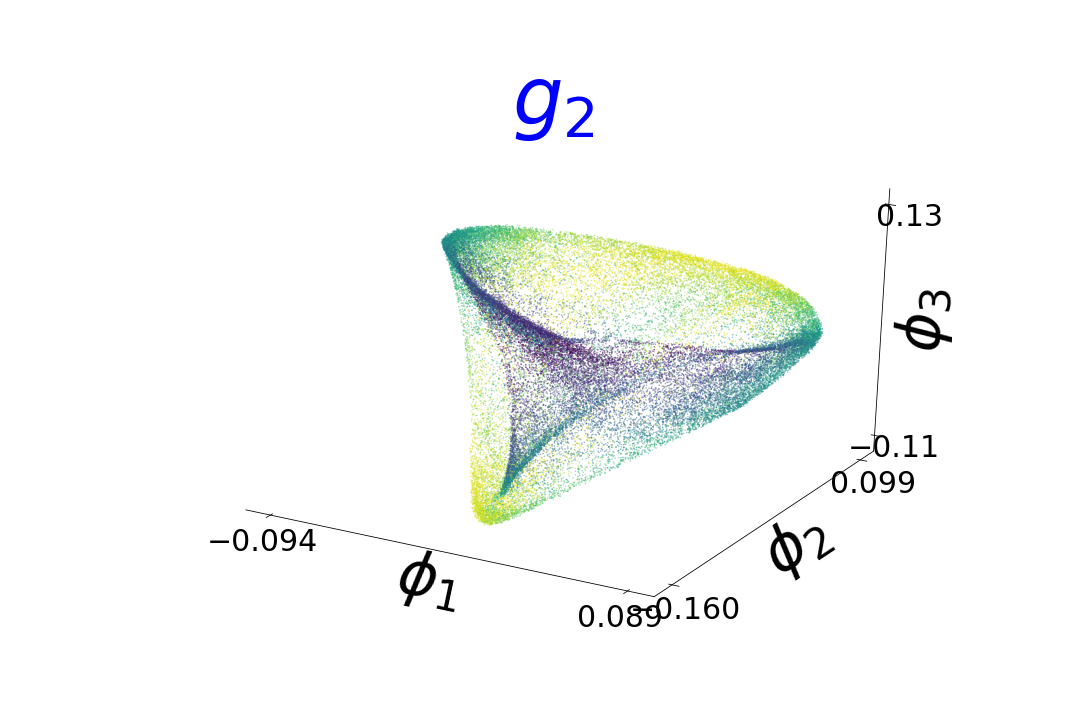}\label{fig:mal_tor2}}\hfill 
\caption{Manifold coordinates with physical meaning in Molecular Dynamics (MD) simulations. \ref{fig:toluene-bonds}-\ref{fig:malonaldehyde-bonds} Diagrams of the toluene ($C_7 H_8$), ethanol ($C_2H_5OH$), and malonaldehyde ($C_3H_4O_2$) molecules, with the carbon (C) atoms in grey, the oxygen (O) atoms in red, and the hydrogen (H) atoms in white. Bonds defining important torsions $g_j$ are marked in orange and blue  (see Section \ref{sec:exps} for more details). The bond torsion is the angle of the planes inscribing the first three and last three atoms on the line (\ref{fig:tor_explain}).  \ref{fig:tol_circ} Embedding of the configurations of toluene into $m=2$ dimensions, showing a manifold of $d=1$. The color corresponds to the values of the orange torsion $g_1$. \ref{fig:eth_tor1}, \ref{fig:eth_tor2} Embedding of the configurations of the ethanol in $m=3$ dimensions, showing a manifold of dimension $d=2$, respectively colored by the blue and orange torsions in Figure \ref{fig:ethanol-bonds}. \ref{fig:mal_tor1}, \ref{fig:mal_tor2}. Embedding of the configurations of the malonaldehyde in $m=3$ dimensions, showing a manifold of dimension $d=2$, respectively colored by the blue and orange torsions in Figure \ref{fig:malonaldehyde-bonds}.}
    \label{fig:molecs}
\end{figure}

For example, in chemistry, a common problem is to discover so-called
{\em collective coordinates} describing the evolution of molecular
configurations at long time scales, which correspond to
macroscopically interesting transformations of the molecule, and can
explain some of its
properties \citep{clementiOnuchicNymeyer:00,Noe2017-up}.  The
molecular configuration is represented by the $3N_a$ vector of spatial
locations of the $N_a$ atoms comprising the molecule. A {\em Molecular
Dynamics (MD) simulation} produces a sample of molecular
configurations; the distribution of this sample describes the
molecule's behavior in the given experimental conditions. It has been
shown empirically that manifolds approximate these high-dimensional
distributions \citep{Dsilva13-Nonlinear}.
Figure \ref{fig:toluene-bonds} shows the toluene molecule, consisting
of $N_a=15$ atoms, and \ref{fig:tol_circ} shows the mapping of an MD
simulated trajectory into $m=2$ dimensions (the embedding coordinates)
by a manifold learning algorithm. Visual inspection shows that this
configuration space is well-approximated by a one-dimensional manifold
parametrized by a geometric quantity, the {\em torsion} $g_1$ of the
methyl bond, which is the angle formed by the planes inscribing the
first three and last three atoms of the colored lines joining four
atoms in Figure \ref{fig:tol_circ}. Thus, $g_1$ is a collective
coordinate which explains the large scale shape of the data manifold
by the rotation of the $CH_3$ methyl group relative to the plane of
the other carbon atoms, filtered out from the faster modes of
vibration by the manifold learning algorithm. Similarly, the large
scale geometry of the ethanol and malonaldehyde MD data is explained
by two rotation angles each.

In this example, while the embedding algorithm was able to uncover the manifold structure of the data, finding the physical meaning of the manifold coordinates was done by visual inspection. In general, a scientist scans through many torsions and other functions of the configuration, in order to find ones that can be identified with the abstract coordinates output by a PCA or ML algorithm.  Manual inspection of such denoised coordinates for correspondences with features of interest is pervasive in a variety of scientific fields \citep{Chen2016-eq, Herring2018-cq}. The goal of this paper is to put this process on a formal basis and to devise a method for automating this identification, thus removing the time consuming visual inspections from the shoulders of the scientist. We introduce a method to automate association of the meaningless abstract coordinates output by an embedding algorithm with functions of the data that are meaningful or interesting in the domain of the problem.

In our paradigm, the scientist has a {\em dictionary} $\G$ of
functions to be considered as possible manifold coordinates. For the
examples in Figure \ref{fig:molecs}, $\G$ could be a set of candidate
torsions.  In other applications like single-cell genomics or
astronomy, data are measurements in high-dimensional feature spaces
such as gene counts, or spectra of stars and galaxies.  The dictionary
$\G$ then consists of functions like cell-type specific signatures for
the former, or element-specific spectral signatures for the
latter \citep{Blanton2017-qd, mcqueenMVdpZ:megaman16, Zhang2020-ki}.

In each case, we assume that the data lies on a low dimensional, smooth manifold $\M$ and that some embedding algorithm maps the data to coordinates denoted by $\phi$. 
We propose an algorithm, \ouralg, that replaces the abstract data-driven coordinates $\phi$ with an ``equivalent'' set of coordinates consisting of functions $g_1,\ldots g_s$ from $\G$. Finding this set of new coordinates, which have physical meaning, can be considered as finding an explanation for the manifold structure $\M$ of the data.  

To keep the approach as general as possible, we do not rely on a
particular embedding algorithm, making only the minimal assumption
that it produces a smooth embedding. We also do not assume a
parametric relationship between the embedding and the functions in the
dictionary $\G$. We only assume that the mapping between the
data manifold and the functions is sufficiently smooth.


The next section defines the problem formally,
and Section \ref{sec:background} presents the necessary background in manifold
estimation. Section \ref{sec:ouralg} develops our method, \ouralg.  
The relationship to previous work is discussed in
Section \ref{sec:related}. Section \ref{sec:theory} presents theoretical recovery results,  Section \ref{sec:exps} presents experiments, and Section \ref{sec:conc} concludes the paper. The Appendices present additional information including details about the functional dictionaries used, adaptions necessary to make our method work in the rotation and translation invariant molecular configuration space, and a useful adaptation to our main algorithm for when support recovery conditions are violated.

%% file: gradients-jmlr-problem.tex
\section{Problem formulation, assumptions and challenges}
\label{sec:problem}
We make a number of standard manifold learning assumptions.
Observed data
$\dataset=\{\xi_i \in \rrr^D : i \in 1 \dotsc n\}$ are sampled
i.i.d. from a {\em smooth manifold} \footnote{The reader is referred
  to \citet{smoothmfd} for the definitions of the differential
  geometric terms used in this paper.} $\M$ of intrinsic dimension $d$
embedded in a feature space $\rrr^D$ by the inclusion map. In this
paper, we will call {\em smooth} any function or manifold of class at
least ${\cal C}^4$. The precise notion of {\em near} varies with
  the embedding approach, and is beyond the scope of this paper. We
assume that the intrinsic dimension $d$ of $\M$ is known; for example,
by having been estimated previously by one method in
\citet{Kleindessner2015DimensionalityEW}. The manifold $\M$ is a {\em
  Riemannian manifold} with {\em Riemannian metric} inherited from the
ambient space $\rrr^D$. Furthermore, we assume the existence of a
smooth {\em embedding map} $\phi:\M\rightarrow \phi(\M)\subset\rrr^m$,
where typically $m << D$.  That is, $\phi$ restricted to $\M$ is
  a diffeomorphism onto its image, and $\phi(M)$ is a submanifold of
  $\rrr^m$.  We call the coordinates $\phi(\xi_i)$ in this $m$
dimensional ambient space the {\em embedding coordinates} $\phi_{1:m}$.  In practice, the
mapping of the data $\dataset$ onto $\phi(\dataset)$ represents the
output of an embedding algorithm, and we only have access to $\M$ and
$\phi$ via $\dataset$ and its image $\phi(\dataset)$.

As mentioned in the previous section, we are given a dictionary of user-defined and domain-related smooth functions $\G=\{g_1,\ldots g_p,\,\text{with }g_j: U \subseteq \rrr^D \rightarrow \rrr\}$, where $U$ is an open set containing $\M$.
We assume that $\phi(x)=h(g_{j_1}(x),\ldots\,g_{j_s}(x))$, where $h:O\subseteq \rrr^s\rightarrow \rrr^m$ is a smooth function of $s$ variables, defined on a open subset of $\rrr^s$ containing the ranges of $g_{j_1},\ldots\,g_{j_s}$.
Let $S=\{{j_1},\ldots\,{j_s}\}$, and $g_S=[g_{j_1}(x),\ldots\,g_{j_s}(x)]^T$.
We call this set the {\em functional support} or {\em explanation}.
In differential geometric terms, $g_S$ is strongly related to finding 
{\em coordinate systems},  {\em charts} and {\em parameterizations} of $\M$. 
For example, in the toluene
example, the functions in $\G$ are all the torsions in the
molecule, $s=1$, and $g_S=g_1$ is a chart for the
1-dimensional manifold traced by the configurations.
Hence, it is natural to associate $s = d$.

\paragraph{Indeterminacies}

Since the map $\phi$ given by the embedding algorithm is determined only up to diffeomorphism, the map $h$ cannot be uniquely determined, and it can therefore be overly restrictive to assume a parametric form for $h$. Hence, this paper aims to find the support set $S$ while circumventing the estimation of $h$.
Indeterminacies w.r.t. the support $S$ itself are also possible. For instance, the support $S$ may not be unique whenever the relationship  $g_1=t(g_2)$, where $t$ is a smooth monotonic function, holds for two functions in $\G$.
 In Section \ref{sec:theory}
we give conditions under which $S$ can be
recovered uniquely; intuitively, they consist of functional
independencies between the functions in $\G$. For instance, it is
sufficient to assume that that the dictionary $\G$ is a {\em
  functionally independent} set, i.e. there is no $g\in \G$ that can be
obtained as a smooth function of other functions in $\G$. 

%% file: gradients-jmlr-background.tex
\section{Manifold learning and intrinsic geometry}
\label{sec:background}

Our method relies on statistical estimators of several geometric quantities.
One of the most essential is estimation of the embedding map $\phi$.
In addition to the embedding map itself, we also estimate the tangent spaces of $\M$ and $\phi(\M)$.
This will allow us to perform support recovery on the differential level.
These estimation tasks are accomplished as follows.


\paragraph{The neighborhood graph and kernel matrix} The {\em neighborhood graph} is a data structure that encodes topological information about the dataset.
It associates to each data point $\xi_i\in\dataset$ its set of {\em neighbors} 
$\neigh_i=\{i'\in [n], \text{with}\,||\xi_{i'}-\xi_{i}||\leq r_N\}$, 
where $r_N$ is a {\em neighborhood radius} parameter.
The neighborhood relation is symmetric, and determines an undirected graph with nodes
 represented by the data points $\xi_{1:n}$.  We denote $|\neigh_i|$ by $k_i$.

This graph is used in construction of the local position matrices $\Xi_i = [\xi_{i'} : i' \in \neigh_i] \in \rrr^{k_i \times D}$, local embedding coordinate matrices $\Phi_i = [\phi(\xi_{i'}) : i' \in \neigh_i] \in \rrr^{k_i \times m}$, and the {\em kernel matrix}
$K \in \rrr^{n \times n}$ whose elements are
\beq
K_{i,i'}= \begin{cases} \exp\left( -\tfrac{||\xi_i-\xi_{i'}||}{\epsilon_N^2}\right)  \; &\text{ if} \;i'\in\neigh_i  \\
0 \; \text{otherwise}.
\end{cases}
\label{eq:kernelmatrix}
\eeq
These matrices encode geometric information about the dataset, and so are of crucial importance; $K$ is sparse, with sparsity
structure induced by the neighborhood graph. 
Typically, the radius $r_N$ and the {\em bandwidth} parameter
$\epsilon_N$ are related by $r_N=c \epsilon_N$ with $c$ a small
constant greater than 1. This value ensures that the entries in $K$ that are zeroed out are small. Rows of the kernel matrix $K$ will be
denoted $K_{i, \neigh_i}$ to emphasize that when a particular row is
passed to an algorithm, only $k_i$ values need to be
passed. Next, we show how the neighborhood graph, local position matrices, and kernel
matrix are used in manifold estimation algorithms.

\paragraph{The renormalized graph Laplacian}

The neighborhood graph and kernel matrix play essential roles in estimation of the
 {\em renormalized graph Laplacian}, also known as the {\em sample
  Laplacian}, or {\em Diffusion Maps Laplacian} $L$.
  This estimator, constructed by the 
\lapalg~ algorithm, converges to the manifold Laplace operator
$\Delta_\M$;
 \citet{coifman:06} shows that this estimator is unbiased
w.r.t. the sampling density on $\M$ (see also \citet{HeinAL:05}, \citet{HeinAL:07,TingHJ:10}).
$L$ is a sparse matrix; its $i$-th row contains non-zero values
only for $i'\in\neigh_i$. Thus, as for $K$, elements and rows of this matrix will
be denoted by $L_{i,i'}$ and $L_{i, \neigh_i}$, respectively, and the sparsity pattern of $L$ is given by the
neighborhood graph.
Construction of this neighborhood graph is
 the computationally expensive component of this
algorithm. 

We use the $m$ principal eigenvectors of $L$ (or alternatively,
of the matrix $\tilde{L}$ of Algorithm \lapalg) corresponding to its
smallest non-zero eigenvalues as embedding coordinates.
This embedding is known as the {\em
  Diffusion Map} \citep{coifman:06} or the {\em Laplacian Eigenmap} \citep{belkin:01} of
$\dataset$.
Although any algorithm which asymptotically generates a smooth embedding is acceptable for our general support recovery method, use of the eigenfunctions of the Laplacian as manifold embedding coordinates has special relevance to quantum systems \citep{Heller1983-hn, Zelditch2006-mw, Landsman_undated-ld,Sogge2014-rs}.


%
\begin{algorithm}[H]
\floatname{algorithm}{\lapalg}
\renewcommand{\thealgorithm}{}
\caption{(neighborhoods $\neigh_{i:n}$, local data $\Xi_{1:n}$, bandwidth $\epsilon_N$)}
\begin{algorithmic}[1]
  \STATE Compute kernel matrix $K$ using (\ref{eq:kernelmatrix})
 \STATE Compute normalization weights  $w_{i}\,\gets\,  \sum_{i' \in \neigh_i} K_{i,i'}$, $i=1,\ldots n$,  $W\,\gets\,\diag(w_{i} \; {i=1:n})$
  \STATE Normalize $\tilde{L}\,\gets\, W^{-1}K W^{-1} $
  \STATE Compute renormalization weights $\tilde w_i\,\gets\,  \sum_{i' \in \neigh_i }\tilde{L}_{i, i'} $, $i=1,\ldots n$, $\tilde W = \diag ( \tilde w_i \;{i=1:n})$
  \STATE Renormalize $L\,\gets\,\frac{4}{ \epsilon_N^2}({\tilde W }^{-1} \tilde L - I_n)$
  \STATE {\bf Output} Kernel matrix $K$, Laplacian $L$, [optionally $\tilde{w}_{1:n}$]
\end{algorithmic}
\end{algorithm}

\paragraph{Estimating tangent spaces in the ambient space $\rrr^D$}

The differential quantities associated with $\phi$ and $\G$ that are used in our support estimation approach are computed w.r.t. the tangent bundle of $\M$.
Since $\M$ is a submanifold of $\rrr^D$, the tangent space at data point $\xi_i$, denoted $\T_{\xi_i}\M$ is representable by an orthogonal basis matrix
$T_i\in\rrr^{D\times d}$.
The estimation of this matrix by
{\em Weighted Local Principal Component Analysis} \citep{Chen2013} is described in the
\tppalg~algorithm. For
this algorithm and others we define the SVD algorithm
$\svd(X,d)$ of a symmetrical matrix $X$ as outputting
$V,\Lambda$, where $\Lambda$ and $V$ are the largest $d$ eigenvalues
and their eigenvectors, respectively.
Denote a column vector of ones of length $k$ by $ \bm{1}_k$. 
\begin{algorithm}[H]
\floatname{algorithm}{\tppalg} \renewcommand{\thealgorithm}{}
\caption{(local data $\Xi_i$,  kernel row $K_{i,\neigh_i}$, intrinsic dimension $d$)}
\begin{algorithmic}[1]
\STATE Compute normalization weights $w_i \,\gets\, \sum_{i' \in \neigh_i} K_{i,i'}$
\STATE Compute weighted mean $\bar \xi_i \gets \frac{1}{w_i} K_{i,\neigh_i} \Xi_i $
 \STATE Compute weighted local differences \\ $Z_{i} \gets \diag ( K_{i , \neigh_i}^{1/2})  (\Xi_{i} -  \bm{1}_{k_i} \bar \xi_i)$
  \STATE Compute $T_i, \Lambda \leftarrow \text{SVD} (Z_i^T Z_i, d)$
  \STATE {\bf Output} $T_i$ 
\end{algorithmic}
\end{algorithm}

\paragraph{The pushforward Riemannian metric}

Geometric quantities such as angles and lengths of vectors in the tangent bundle $\T\M$ and distances along curves in $\M$ are captured by Riemannian geometry.
Recall our assumption that $(\M,\id)$ is a Riemannian manifold, with the metric $\id$ induced from $\rrr^D$. With this we associate to $\phi(\M)$ a Riemannian metric $\g$ which preserves the geometry of $(\M,\id)$. This metric - the pushforward metric - is defined by
\beq \label{eq:rmetric0}
\langle u,v\rangle_{\g}\;=\;\langle D\phi^{-1}(\xi)u,D\phi^{-1}(\xi)v\rangle
\quad\text{ for all }u,v\in\T_{\phi(\xi)}\phi(\M).
\eeq
In the above, $D$ denotes the differential operator, $D\phi^{-1}(\xi)$ the {\em pull-back} operator that maps vectors from $\T_{\phi(\xi)}\phi(\M)$ to $\T_\xi\M$, and $\langle , \rangle$  the Euclidean scalar product. 

For each $\phi(\xi_i)$, the pushforward Riemannian metric
expressed in the coordinates of $\rrr^m$ is a symmetric,
semi-positive definite $m\times m$ matrix $G_i$ of rank $d$.
The scalar product $\langle u,v \rangle_{\g}$ takes the form $u^TG_iv$.
The matrices $G_i$ can be estimated by the algorithm \rmalg~of
\citet{2013arXiv1305.7255P}. The algorithm uses only local information, and thus can be run
efficiently using the Laplacian, the neighborhood graph, and local embedding coordinate matrices. In the next section, we will use the output of this algorithm to estimate the differential $D\phi$.
\begin{algorithm}[H]
\floatname{algorithm}{\rmalg}
\renewcommand{\thealgorithm}{}
\caption{(Laplacian row $L_{i, \neigh_i}$, local embedding coordinates $\Phi_{i}$, intrinsic dimension $d$)}
\begin{algorithmic}[1]
\STATE Compute centered local embedding coordinates \\
$\tilde \Phi_i = \Phi_i - \bm{1}_{k_i} \phi(\xi_i)^T$
\STATE 
Form matrix $H_i$ by \\
$H_i \gets [H_{i,k,k'}]_{k,k' \in 1:m}$ with 
 $H_{i,k,k'}=\sum_{i'\in \neigh_i}L_{i,i'}\tilde \Phi_{i,i',k} \tilde \Phi_{i,i',k'}$ for $k,k'=1:m$. 
  \STATE  Compute $V_i, \Lambda_i \gets \text{SVD} (H_i, d)$
    \STATE $G_i\,\gets\,V_i \Lambda_i^{-1} V_i^T$.
  \STATE {\bf Output} $G_i$, optionally $V_i,\Lambda_i$
\end{algorithmic}
\end{algorithm}
%

%% file: gradients-jmlr-pullback.tex
\section{The \ouralg~algorithm}
\label{sec:ouralg}
\mmp{todo: 1) main FLASSO equation, normalize by 1/mn and explain 2) pass over
  appendix A (Hanyu?) 3) normalization of gradients in algorithm 4)
  change all $\lambda/\sqrt{mn}$ preferably to just $\lambda$ wherever
  possible, e.g. experiments 5) pass with a fine comb over 4 and 6.2
  to make sure all refs and notations are correct. 6) experiments
  section polish up (Sam) }
  The main idea of our approach is to
exploit the well-known mathematical fact that, for any differentiable
functions $f,g,h$, when $f=h\circ g$, the differentials $Df,Dh,Dg$ at
any point are in the {\em linear} relationship $Df= Dh Dg$.
 Since, given  coordinate functions
$\phi_{1:m}$ and dictionary
functions $g_{1:p}$ on a smooth manifold $\M$, our goal is to recover a
subset $g_{S}$ of $g_{1:p} $ such that $\phi_{1:m} =h \circ g_S$ without knowing $h$,
we propose to recover the subset $g_{S}$ 
by solving a set of dependent linear sparse recovery problems, one for
each data point. 
This linear relationship $D\phi = Dh Dg_S$ will be written in terms of gradients $\grad_\M \phi_{1:m}$ and $\grad_\M g_{1:p}$.
This section describes how to obtain the relevant gradients and solve the resulting set of {\em dependent linear sparse recovery} problems.

\subsection{Algorithm Overview}

The \ouralg~algorithm, the main algorithm of this paper, implements this idea. It takes as input data $\dataset$ sampled from
an unknown manifold $\M$, a dictionary $\G$ of functions defined on
$\M$ (or alternatively on an open subset of the ambient space $\rrr^D$
that contains $\M$), and an embedding $\phi(\dataset)$ in
$\rrr^m$. The output of \ouralg~is a set $S$ of indices in $\G$,
representing the functions in $\G$ that explain $\M$. 

The first part of the algorithm contains preparatory steps for geometric analysis covered in Section \ref{sec:background}. Steps \ref{alg:neb} and \ref{alg:lap} construct the
neighborhood graph and the Laplacian matrix used for manifold learning and tangent space estimation.

The second part of \ouralg~calculates the necessary gradients; this comprises Steps \ref{alg:tan}--\ref{alg:prob}. 
In Step \ref{alg:tan}, we estimate orthogonal bases of tangent subspaces by the \tppalg~algorithm described in Section \ref{sec:background}. 
The gradients of the dictionary
w.r.t. the manifold are then obtained as columns of the $d\times p$ matrix
$X_i$ in Steps \ref{alg:dict}, \ref{alg:dict-norm},  and \ref{alg:dict_proj}. These
operations are described in detail in Section \ref{sec:grad-g}. In
Step \ref{alg:pb}, the gradients at $\xi_i$ of the coordinates
$\phi_{1:m}$, also w.r.t. $\M$, are calculated as columns of
the $d \times m $ matrix $Y_i$ by the \dpullalg~algorithm described in
Section \ref{sec:pull-back}.

In the last part of \ouralg, Step \ref{alg:glasso} finds the support $S$ by solving the
sparse regression. A \flassoalg~algorithm \mmp{arises naturally} is called to perform the
sparse regression of the manifold coordinates' gradients $Y_{1:n}$ on
the gradients of the dictionary functions, represented by
$X_{1:n}$. The indices of those dictionary functions whose $\beta$
coefficients are not identically null represent the support set $\supp
\beta$. This is described in Section \ref{sec:flasso-manifold}.
Scaling of functions is addressed through normalization in Steps \ref{alg:dict-norm} and \ref{alg:y-norm}; this procedure is described in more detail in Section \ref{sec:ouralg-normalization}.

\begin{algorithm}[H]
\floatname{algorithm}{\ouralg}
\renewcommand{\thealgorithm}{}
\caption{(Dataset $\dataset$, dictionary $\G$, embedding coordinates $\phi(\dataset)$,  intrinsic dimension $d$, kernel bandwidth $\epsilon_N$, neighborhood cutoff size $r_N$, regularization parameter $\lambda$)}
\begin{algorithmic}[1]
\STATE Construct $\neigh_i$ for $i=1:n$; $i'\in \neigh_i$ iff $||\xi_{i'}-\xi_{i}||\leq r_N$, and local data matrices $\Xi_{1:n}$ \label{alg:neb}
\STATE Construct kernel matrix and Laplacian $K, L\,\gets\,$\lapalg($\neigh_{1:n},\Xi_{1:n},\epsilon_N$) \label{alg:lap}
\STATE [Optionally compute embedding: $\phi(\xi_{1:n})\gets$\embedalg$(\dataset,\neigh_{1:n},m,\ldots)$] \label{alg:embed}
  \FOR {$j=1,2,\ldots p$}
  \STATE Compute $\nabla_\xi g_j (\xi_i)$ for $i=1,\ldots n$ \label{alg:dict}
  \STATE Compute $\zeta_j^2$ by \eqref{eq:gammaj-discrete-xi} and normalize $\nabla_\xi g_j (\xi_i)\gets (1/\zeta_j ) \nabla_\xi g_j (\xi_i)$ for $i=1,\ldots n$ \label{alg:dict-norm}
  \ENDFOR
  \FOR {$i=1,2,\ldots n$} 
  \STATE Compute basis  ${T_i^\M} \gets $\tppalg$(\Xi_i, K_{i, \neigh_i},d)$ \label{alg:tan}
  \STATE Project $X_i\gets ({T_i^\M})^T\nabla_\xi g_{1:p}$ \label{alg:dict_proj}
  \STATE Compute $Y_i \gets$\dpullalg$(\Xi_i, \Phi_{i},T_i^\M,  L_{i, \neigh_i},  d)$ \label{alg:pb} \label{alg:prob}
  \ENDFOR
\STATE Compute $\zeta_k^2\gets\frac{1}{n}\sum_{i=1}^n\|y_{ik}\|^2$ (i.e. \eqref{eq:gammaj-discrete-M}), for $k=1,\ldots m$ and \\
normalize $Y_i\gets Y_i\diag\{1/\zeta_{1:m}\}$, for $i=1,\ldots n$. \label{alg:y-norm}
\STATE $\beta\,\gets$ \flassoalg$(X_{1:n}, Y_{1:n}, \lambda\sqrt{mn})$ \label{alg:glasso}
\STATE {\bf Output} $S=\supp \beta$ 
\end{algorithmic}
\end{algorithm}

There are several optional steps and substitutions in our algorithm. An embedding can be
computed in Step \ref{alg:embed}, or input separately by the user - we
denote this step generically as \embedalg.  Finally, although we explicitly describe tangent space estimation methods of both $\T_{\xi_i} \M$ and $\T_{\phi(\xi_i)} \phi (\M)$ in our algorithms, other approaches to estimate them may be used.

\comment{\mmp{removed (or subset) because it has to be also added to the $\nabla g_j$ gradients}
An embedding can be computed in Step \ref{alg:embed}, or input separately by the user.
We denote this step generically as \embedalg.}

\subsection{Gradients and coordinate systems}
\label{sec:explain-dphi}

Our algorithm regresses the gradients of the embedding coordinate
functions against the gradients of the dictionary functions.
Both sets of gradients are with respect to the manifold $\M$, and so this requires calculating or estimating various gradients in the same $d$-dimensional coordinate system.
This and the following two sections explain these procedures. 

First, note that by assumption we have two Euclidean spaces $\rrr^D$ and $\rrr^m$, in which manifolds $\M$ and $\phi(\M)$ of dimension $d$ are embedded.
Denote gradients w.r.t. the Euclidean coordinate systems in $\rrr^D$ and $\rrr^m$ by $\nabla_\xi$ and $\nabla_\phi$, respectively.
Since our interest is in functions on manifolds, we
also define the gradient of a function on a manifold $\M$.
The {\em gradient} of $f$ at $\xi$, on a Riemannian manifold $(\M,\g$), denoted $\grad_{\M} f(\xi)\in\T_\xi\M$, is defined by the identity  
\beq \label{eq:gradM}
\langle \grad_{\M} f(\xi),v\rangle_{\g}=Df(\xi)u\quad\quad\text{ for any $u\in \T_\xi \M$}.
\eeq
At each data point $\xi_i$, we fix bases $T_i^\M$ in $\T_{\xi_i}\M$
and $T_i^\phi$ in $\T_{\phi (\xi_i)} \phi (\M)$. Gradients expressed
in these coordinate systems are denoted by $\grad_{T_i^\M,\g}$ and
$\grad_{T_i^\phi,\g}$ respectively.  For a manifold $\M$ which is a
submanifold of $\rrr^D$, we denote by $\grad_{T_i^\M}(\xi)$ the value of
$\grad_{T_i^\M, \id}(\xi)$ w.r.t. the identity metric $\id$ inherited from
$\rrr^D$, and by $\langle,\rangle$ the Euclidean scalar product.

Note that in coordinates, $\grad_\M f$ depends on the metric $\g$, but at the same time, this definition shows that $\grad_\M f$ as a linear operator on $\T_{\xi_i}\M$ is invariant to the metric; it is just the first order derivative reflecting how $f$ changes along the manifold.  Hence, the left hand side must also be invariant to the metric.  It follows that $Df(\xi)u=u^T\grad_{T_i^\M}(\xi_i) $ for any $u\in \T_\xi\M$, and, furthermore,  that $\grad_{T_i^\M,\g}=G_i^{-1}\grad_{T_i^\M}$ for any other Riemannian metric $\g$. 

\subsection{Calculating the gradients of the dictionary functions}
\label{sec:grad-g}
Our goal is to obtain $X_i$, the matrix defined by
\beq \label{eq:X-manifold}
X_i\,=\,[\grad_{T_i^\M} g_j(\xi_i)]_{j=1:p}\,\in \rrr^{d\times p}.
\eeq
Let $g_j$ be a function in the dictionary $\G$.
By definition, for any basis $T_i^\M \in \rrr^{D\times d}$ of $\T_{\xi_i}\M$,
\beq
\label{eq:gradprojg} \grad_{T_i^\M} g_j(\xi_i)\;=\;(T_i^\M)^T\nabla_{\xi}g_j(\xi_i).
\eeq
In other words, $\grad_{T_i^\M}g_j$ is the projection of $\nabla_\xi g_j$ on the basis $T_i^\M$.
These bases of $\T_{\xi_i}\M$, for every $i$, are estimated by \tppalg~as described in Section \ref{sec:background}.
The gradients $\nabla_\xi g_j(\xi_i)$ are known analytically, by assumption.
We thus construct matrices $X_i$, for $i=1,\ldots n$, with $p$ columns representing the gradients of the $p$ dictionary functions as $X_i=({T_i^\M})^T\nabla_\xi g_{1:p}$, as in Step \ref{alg:dict_proj} of Algorithm \ouralg.
Now we turn to obtaining the manifold gradients of the coordinate functions $\phi_k$ in the same coordinate system.

\subsection{Estimating the coordinate gradients by pull-back}
\label{sec:pull-back}

\begin{figure}[H]
\setlength{\picwi}{0.85\llw}

\centerline{  \includegraphics[width=\picwi]{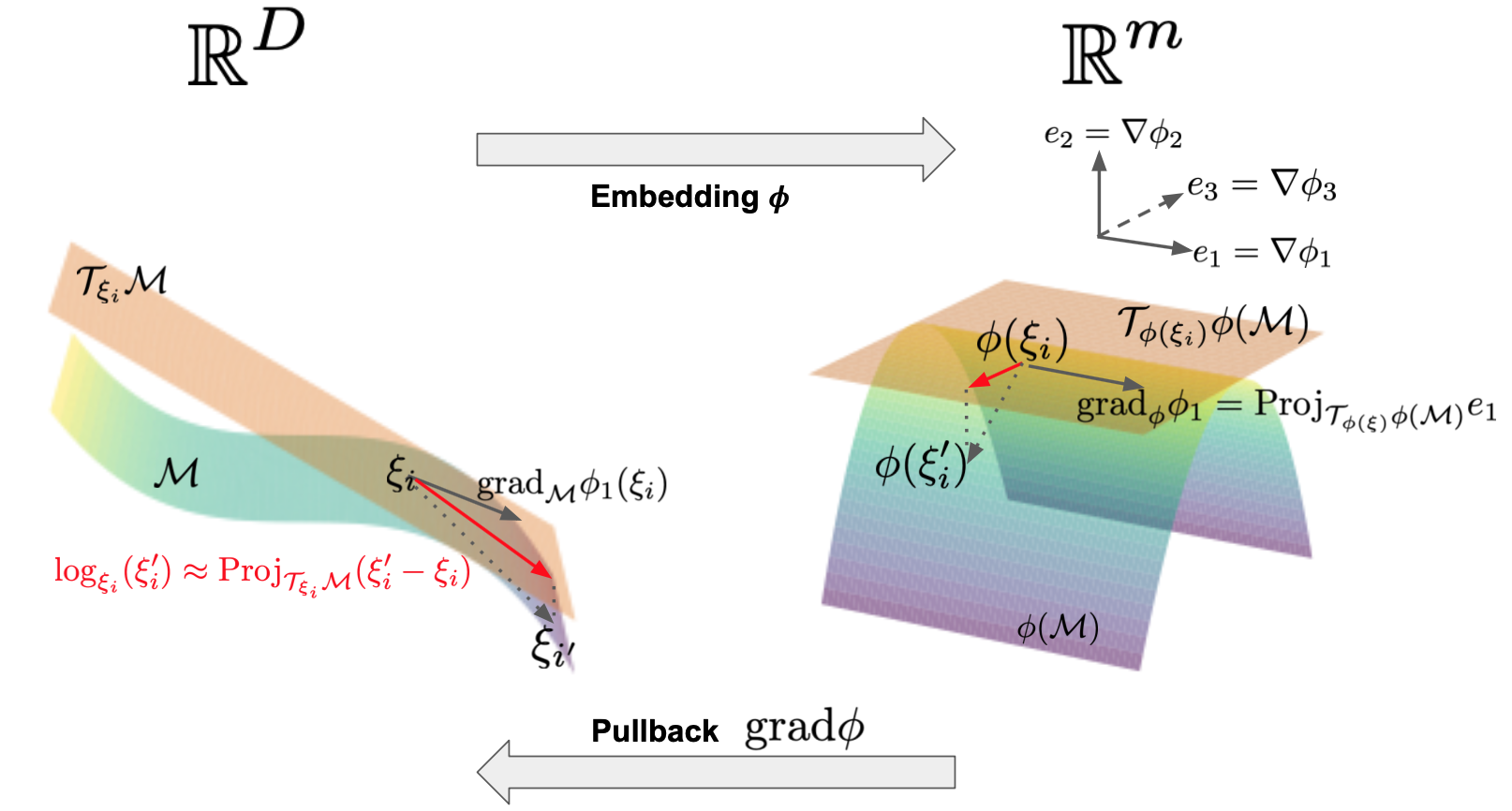}}
\comment{
\setlength{\picwi}{0.4\llw}
\begin{tabular}{cc}
  \includegraphics[width=\picwi,height=2\picwi]{../Figures/pullback_figure.png}
$\M$ in original ambient space  $\rrr^D$ &  $\phi(\M)$ in embedding space $\rrr^m$\\

\includegraphics[width=0.48\textwidth]{../Figures/ManifoldSpace3.png}
&
\includegraphics[width=0.48\textwidth]{../Figures/Embedded_space2.png}
\\
& 
\end{tabular}
}
\caption{\label{fig:pullback} Left: $\M$ with a tangent subspace at $\xi_i$, $\xi_{i'}-\xi_i$  (in black, dotted), the projection $\proj_{\txim}(\xi_{i'}-\xi_i)$  (in red), and the manifold gradient $\grad_\M\phi_1(\xi_i)$ of the first embedding coordinate $\phi_1$ (in black). Right: $\phi(\M)$ and tangent subspace at $\phi(\xi_i)$, with $\phi(\xi_{i'})-\phi(\xi_i)$ (in black, dotted),  $\proj_{\tphim}(\phi(\xi_{i'})-\phi(\xi_i))$ (in red) and the manifold gradient $\grad_{\phi(\M)} \phi_1$ (in black).  $A_{i,i'}$ is the (approximate) mapping of $B_{i,i'}$ through $D\phi^{-1}(\xi_i)$, as in \eqref{eq:rmetric0}. The gradient $\grad_{\phi(\M)}\phi_1$ is the the projection of the first unit vector onto $\tphim$.
\mmp{Embedding $D\phi$ or $\phi$?}}
\end{figure}

Since $\phi$ is implicitly determined by a manifold embedding algorithm, the gradients of $\phi_k$ are often not analytically available, and $\phi_k$ is known only through its values at the data points.
We therefore introduce an estimator of these gradients based on the notion of vector {\em pull-back} between tangent spaces.
Instead of estimating these gradients naively from differences
$\phi_k(\xi_i)-\phi_k(\xi_{i'})$ between neighboring points, we first
estimate their values in $\T_{\phi(\xi_i)}\phi(\M)$, where they have a
 simple expression, then pull them back in the coordinate system
$T_i^\M$. This estimation method is novel, and of some independent interest.
A schematic of this approach is given in Figure \ref{fig:pullback}.

The \dpullalg~Algorithm  takes as inputs the local neighborhoods $\Xi_i,\,\Phi_i$ of point $\xi_i$ in the original and embedding spaces, respectively, the basis $T_i^\M$ of $\T_{\xi_i}\M$, and the row of the Laplacian matrix corresponding to $i$, $L_{i,\neigh_i}$. From this local information, the algorithm first computes the tangent space  $\tphim$, then obtains the gradients of the coordinate functions $\phi$ in this space by projection, and finally pulls back these gradients in the coordinate system given by $T_i^\M$ by solving a least squares regression.

\paragraph{The tangent space \tphim} When $m=d$, this space is trivially equal to $\rrr^d$, so the problem is interesting in the case $m>d$. If the embedding induced by $\phi$ were an isometry, the estimation of
$\tphim$ could be performed by \tppalg, and the subsequent pull-back
could be done as described in \citet{Luo2009-mp}. Here we do {\em not}
assume that $\phi$ is isometric.

The method we introduce uses the push-forward Riemannian metric $\g$,
expressed as $G_i$ in the coordinates $\phi$ at $\xi_i$, to estimate
the $\tphim$.  By definition, the theoretical rank of $G_i$ equals $d$ and the $d$
principal eigenvectors of $G_i$ represent an orthonormal basis of
$\T_{\phi(\xi_i)}\phi(\M)$. $G_i$ and its decomposition are estimated by the
\rmalg~algorithm described in Section \ref{sec:background}. We denote
this basis $T_i^\phi\in\rrr^{m\times d}$. 

\paragraph{The gradient $\grad_{\phi(M)} \phi_k$} Trivially, the gradients of $\phi_{1:m}$ in the embedding space $\rrr^m$, are equal to the $m$ basis vectors of $\rrr^m$, i.e. $\nabla_\phi \phi_{1:m}=I_m$. Therefore $\grad_{\phi(M)} \phi_k$, expressed in the basis $T_i^\phi$, given by the top $d$ the eigenvectors of $G_i$, is equal to the projection of the corresponding basis vector onto the tangent subspace $\T_{\phi(\xi_i)} \phi(\M)$. In matrix form we have 
\beq\label{eq:tildeY}
[\grad_{T_i^\phi}\phi_k(\xi_i)]_{k=1}^m\;=\;(T_i^\phi)^TI_m.
\eeq

\paragraph{Pulling back $\grad_{\phi(\M)} \phi$ into $\T_{\xi_i}\M$} 

In order to bring these gradients into the same coordinate system as our dictionary functions,
we define the following matrices, with $\proj_T v$ denoting the Euclidean projection of vector $v$ onto subspace $T$.
\beq \label{eq:yYbetaijk-manifold}
Y_i\,=\,[y_{ik}]_{k=1}^m\,=\,[\grad_{T_i^\M}\phi_k(\xi_i)]_{k=1}^m\in \rrr^{d\times m},
\eeq
\beq
A_i\,=\,\left[\proj_{\txim}(\xi_{i'}-\xi_i)\right]_{i'\in
  \neigh_i} \in \rrr^{d\times k_i}\,,
\eeq
and
\beq 
B_i\,=\,\left[\phi(\xi_{i'})-\phi(\xi_i)\right]_{i'\in \neigh_i}
\;\in \rrr^{m\times k_i},
\quad
\tilde{B}_i\,=\,\left[\proj_{\tphim}   \left[ \phi (\xi_{i'}) - \phi(\xi_i) \right]  \right]_{i'\in \neigh_i},
\;\in \rrr^{d\times k_i}.
\eeq
The columns of $A_i$ and $Y_i$ are vectors in $\T_{\xi_i}\M$, the columns of $B_i$ are in $\rrr^m$ and the columns of $\tilde{B}_i$ are in $\T_{\phi(\xi_i)}\phi(\M)$.
These vectors are shown schematically in Figure \ref{fig:pullback}.  Note that when $m=d$, $B_i=\tilde{B}_i$. 

The key property that enables our estimator of $Y_i$ is that the columns of $A_i$ and $\tilde{B}_i$ are in
correspondence, because they represent (approximately)
the same vectors in two different coordinate systems, namely the {\em
  logarithmic maps} of point $i'$ in $\M$ and $\phi(\M)$ with respect to point $i$.
The accuracy of this approximation is shown in Appendix \ref{sec:log-map}\mmp{make it a lemma?}. The idea of the algorithm is then to use this correspondence in order to pull back the gradient of the coordinate function $\phi_k$ into the coordinates $T_i^\M$.

Specifically, since $D\phi_k$, the differential of
$\phi_k:\M\rightarrow \rrr$, as a linear functional on the tangent
bundle $\T\M$ is invariant to coordinate system, we calculate its
value on the columns of $\tilde{B}_i$ in the coordinate system given by $\phi$
itself, and equate these values with $\grad_{T_i^\M}\phi_k$ applied to
the columns of $A_i$, an expression in the coordinates $T_i^\M$. By
\eqref{eq:gradM} and Appendix A, we have that 
\beq 
(\grad_{T_i^\M}\phi_k(\xi_i))^TA_i\;=\;(\grad_{T_i^\phi}\phi_k(\xi_i))^T\tilde{B}_i+o(r_N).
\eeq
\mmp{hanyu says $r_N$}\hanyuz{In section 2, $r_N$ is the neighborhood parameter and $\epsilon_N$ is the bandwidth parameter in the gaussian kernel. Here it should be the neighborhood parameters $r_N$. In the appendix I switch them to $r_N$.}
In coordinates, $A_i=(T_i^\M)^T(\Xi_i^T-\xi_i\bm{1}_{k_i}^T)$ and
$\tilde{B}_i=(T_i^\phi)^T(\Phi_i^T-\phi(\xi_i)\bm{1}_{k_i}^T)$.
 These matrices are computed by Steps  \ref{alg:dpull-B} and \ref{alg:dpull-A} of Algorithm \dpullalg, while $Y_i$ contains the gradients we want to estimate.
The error term comes from approximating the
logarithmic map applied to points $\xi_{i'}$ and
$\phi(\xi_{i'})$ for $i'\in \neigh_i$ with the columns of $A_i$ and $\tilde{B}_i$.  Recalling that
$Y_i=[\grad_{T_i^\M}\phi_k(\xi_i)]_{k=1:m}$ we obtain
\beq
\label{eq:lin_sys}
Y_i^TA_i\;=\;[(T_i^\phi)^TI_m]^T(T_i^\phi)^TB_i+o(r_N).
\eeq
We solve this linear system in the least squares sense
\beq \label{eq:yi-manifold2}
Y_i\;=\;\arg\min_{Y\in\rrr^{d\times m}}\|A_i^TY-B_i^TT_i^\phi (T_i^\phi)^T\|^2
\eeq
to obtain
\beq \label{eq:yi-manifold3}
Y_i\;=\;A_i^\dagger B_i^TT_i^\phi(T_i^\phi)^T. 
\eeq
This solution is effectively the regression of the columns of $B_iT_i^\phi (T_i^\phi)^T$ on the columns of $A_i$ at each data point $\xi_i$. We call estimator \eqref{eq:yi-manifold3} the {\textit{pullback gradient estimator}} because of its implicit invocation of the notion of vector pullback.

To see this a different way, note that by equation \eqref{eq:rmetric0}, for any function $f:\phi(\M)\rightarrow \rrr$,
\beq\label{eq:v-pullback}
\langle D \phi^{-1}  u,   D \phi^{-1}   \grad_{\phi(\M)} f  \rangle \,=\, \langle u,  \grad_{\phi(\M)} f  \rangle_{\g},\quad \text{for all }u\in \T_{\phi(\xi_i)} \phi(\M)
\eeq
where $g$ is the push-forward metric associated with $\phi$.
Using this fact, and the invariance of gradient to metric, we have that, for any $w \in \T_{\xi_i}\M$, $D \phi^{-1}  \grad_{\phi(\M)} f = \grad_{\M} (f\circ \phi)$ for any smooth function $f:\phi(\M)\rightarrow \rrr$.
\comment{
\begin{align*}
&\langle D \phi^{-1}  \grad_{\phi(\M)} f - \grad_{\M} (f\circ \phi), w \rangle \\
=&\langle D \phi^{-1}  \grad_{\phi(\M)} f ,w\rangle - \langle \grad_{\M} (f\circ \phi), w \rangle \\
=&\langle \grad_{\phi(\M)}f, D\phi(w)\rangle_{\g}- \langle \grad_{\M} (f\circ \phi), w \rangle \\
=& Df(D\phi)(w) - D(f\circ\phi)(w)\\
=&0.
\end{align*} 
}
The above claims give us
$\langle D \phi^{-1} u,    \grad_{\M} (f \circ \phi) \rangle = \langle u,  \grad_{\phi(\M)} f \rangle$
where $u \in \tphim$ is an arbitrary tangent vector.  In coordinates $T_i^\phi$ and $T_i^\M$, we can write this equivalence as 
\beq\label{eq:v-pullback1}
\langle D \phi^{-1} u,    \grad_{T_i^\M} (f \circ \phi) \rangle = \langle u,  \grad_{T_i^\phi} f \rangle.
\eeq
If we then replace values of $(T_i^\phi )^Te^k$, $(T_i^\M)^T (\xi_{i'}
- \xi_i)$ and $(T_i^\phi)^T (\phi(\xi_{i'}) - \phi(\xi_i))$ for
$\grad_{T_i^\phi} \phi_k$, $D \phi^{-1} u$ and $D u$, respectively, we obtain \eqref{eq:lin_sys}.
\begin{algorithm}[H]
\floatname{algorithm}{\dpullalg}
\renewcommand{\thealgorithm}{}
\label{alg:pullback} 
\caption{local data $\Xi_i$, local embedding coordinates $\Phi_{i}$, basis $T_i^\M$ (Optional: $T_i^\phi$ or Laplacian row $L_{i,\neigh_i}$, intrinsic dimension $d$)}
\begin{algorithmic}[1]
  \STATE Compute pushforward metric eigendecomposition $T_i^\phi, G_i \gets$ \rmalg$( L_{i,\neigh_i}, \Phi_{i}, d)$.\label{alg:dpull-rmetric}
 \STATE Compute $B_{i} \gets  (\Phi_i^T-\phi(\xi_i)\bm{1}_{k_i}^T) $ \label{alg:dpull-B}
\STATE Compute $A_{i} \gets (T_i^\M)^T(\Xi_i^T -\xi_i\bm{1}_{k_i}^T) $\label{alg:dpull-A}
   \STATE Calculate $Y_i \gets A_i^\dagger B_i^TT_i^\phi(T_i^\phi)^T$ by solving linear system \eqref{eq:yi-manifold2} \label{alg:dpull-ls} 
  \STATE {\bf Output} $Y_i$ 
\end{algorithmic}
\end{algorithm}

\subsection{The \flassoalg~formulation}
\label{sec:flasso-manifold}

With the estimated gradients, we are now ready  to resolve the functional support problem.
  Recall that $X_i$ defined in
\eqref{eq:X-manifold} contains the gradients of the dictionary
functions $g_j$, and that  $y_{ik}\in \rrr^d$, the $k$-th column of $Y_i$, represents the coordinates of $\grad_{\M}\phi_k(\xi_i)$ in the chosen basis of \txim.
Further, given our assumption that $\phi=h\circ g_S$, let
$h_k$ be the $k$-th component of the vector valued function $h$, and denote
\beq \label{eq:beta-ijk} \beta_{ijk}=\fracpartial{h_k}{g_j}( g_j( \xi_i)),
\quad \beta=[\beta_{ijk}]_{i,k,j=1}^{n,m,p}, \eeq
\beq \label{eq:betas-manifold}
\beta_j=\operatorname{vec}(\beta_{ijk},\;i=1:n,k=1:m)\in\rrr^{mn},
\quad \beta_{ik}=\operatorname{vec}(\beta_{ijk},\;j=1:p)\in\rrr^{p}.
\eeq
Then, from the identity $\grad_{M} \phi_k = \grad_{M} (h_k \circ g_S)$ and the chain rule, one obtains the following linear model.
\beq
y_{ik}\;=\;\sum_{j=1}^p\beta_{ijk}x_{ij}+\epsilon_{ik}\;=\;X_i\beta_{ik}+\epsilon_{ik}
\quad \text{for all $i=1:n$, and $k=1:m$.}
\eeq
In the above regression of $Y_{1:n}$ on $X_{1:n}$, $\beta_{ik}$ is the set of regression
coefficients of $y_{ik}$ onto $X_i$. If there is some $h$ such that $\phi=h\circ g_S$, then the non-zero $\beta_{ijk}$ coefficients are estimates of $\fracpartial{h}{g_j}$ for $j\in S$. Further, $\beta_j$ represents
the vector of regression coefficients corresponding to the effect of
function $g_j$; therefore, the zero $\beta_j$ vectors indicate that $j\not\in S$. Hence, in each $\beta_{ik}$, only $|S|$ elements are non-zero. The term $\epsilon_{ik}$ is added to account for noise or model misspecification.

The key characteristic of the functional support that we leverage is that the same set $S$ of coefficients will be non-zero for all $i$ and $k$. Since finding this set $S\subset [p]$ is underdetermined, we use a sparsity inducing regularization that simultaneously zeros out entire $\beta_j$ vectors. Thus, our problem can be naturally expressed as a  {\em Group Lasso} \citep{Yuan2006-af}, with $p$ groups of  size  $mn$, consisting of the $\beta_{1:p}$ groups of coefficients of $\grad_\M g_{1:p}$. To solve it we minimize the following objective function w.r.t. $\beta$:
\beq\label{eq:flasso-manifold}
J_{\lambda}(\beta)\;=\;\frac{1}{2}\sum_{i=1}^n\sum_{k=1}^m||y_{ik}-X_i\beta_{ik}||^2+\frac{\lambda}{\sqrt{mn}}\sum_{j=1}^p||\beta_j||.
\eeq
The first term of the objective is the least squares loss of
regressing $Y_{1:n}$ onto $X_{1:n}$. The second is a regularization
term, which penalizes each group $\beta_j$ by its Euclidean
norm. This encourages most $\beta_j$ groups to be
identically 0. The normalization of the regularization coefficient
$\lambda$ by the group size $mn$ follows \citet{Yuan2006-af} takes into account that the least squares loss also grows proportionally to $mn$.
The use of Group Lasso for sparse functional regression was introduced in \citet{MKoelleZhang:arxiv1811-11891}.

Note that $J_\lambda(\beta)$ is convex in $\beta$ and invariant to
the change of basis $T_i$. Let $\tilde{T}_i=T_i\Gamma$ be a different
basis, with $\Gamma\in \rrr^{d\times d}$ a unitary matrix. Then,
$\tilde{y}_{ik}=\Gamma^Ty_{ik}$, $\tilde{X}_i=\Gamma^TX_i$, and
$||\tilde{y}_{ik}-\tilde{X}_i\beta_{ik}||^2=||y_{ik}-X_i\beta_{ik}||^2$ for any $\beta_{ik}\in\rrr^p$. 

\subsection{Computation, normalization, and tuning}
\paragraph{Computation}
\label{sec:ouralg-computation}

The first two steps of \ouralg~are construction of the neighborhood
graph and estimation of the Laplacian $L$. As shown in Section
\ref{sec:background}, $L$ is a sparse matrix, hence \rmalg~can be run
efficiently by only passing values corresponding to one neighborhood
at a time. Note that in our examples and experiments, Diffusion Maps
is our chosen embedding algorithm, so the neighborhoods and Laplacian are already available, though in general this is not the case.
The second part of the algorithm estimates the gradients and constructs matrices $Y_{1:n},X_{1:n}$.
The gradient estimation runtime, with  Cholesky decomposition-based solvers, is $O(qd^2 + nd^3)$ where $q=\sum_{i=1}^nk_i$ is the number of edges in the neighborhood graph.
The last major step is a call to the \glassoalg~solver, which estimates the support $S$ of $\phi$.
The computation time of each iteration in \glassoalg~is $O(n m pd)$.  
Note that when using a standard group lasso solver, the computation time is $O(n^2 m^2 pd)$ due to the block-diagonal structure of the problem implicit in flattening the $n$ by $p$ by $d$ covariate tensor.
We therefore use our own implementation of proximal FISTA to solve this problem \citep{Boyd2004-co, Jas2020-se}.
Finally, for large data sets, we perform the 'for' loop over a subset $\I\subset[n]$ of the original data while retaining the geometric information from the full data set.
This replaces the $n$ in the computation time with the smaller factor $|\I|$.

\paragraph{Normalization}
\label{sec:ouralg-normalization}
As with many sparse regression methods, normalization is necessary to balance the relative influence of  dictionary elements and embeddings coordinates.
Multiplying $g_j$ by a non-zero constant and dividing
its corresponding $\beta_j$ by the same constant leaves the
reconstruction error of all $y$'s invariant, but affects the norm
$||\beta_j||$.
Therefore, the relative scaling of the dictionary
functions $g_j$ can influence the recovered support $S$, by favoring
the dictionary functions whose columns have larger norm.
A similar effect is present if a particular embedding coordinate $\phi_k$ is rescaled by a constant.
For example, multiplying a certain $\phi_k$ by a number close to zero will cause the penalty accrued by learned coefficients for that coordinate to be smaller than for the other coefficients, and for that $\phi_k$ to dominate support recovery.

We therefore normalize all $\grad_{T^\M_i}\phi_{1:m}$ and $\grad_{T^\M_i}g_{1:p}$ as follows. Denote $f$ a function on $\M$, which can be either a coordinate function or a dictionary function.
When $f$ is defined on $\M$, but not
outside $\M$, we calculate the {\em normalizing constant}
\beq \label{eq:gammaj-discrete-M}
\gammanorm^2\;=\;\frac{1}{n}\sum_{i=1}^n\|\grad_{T_i^\M} f(\xi_i)||^2;
\eeq
then we set $f \leftarrow f/\gammanorm$.  
The above $\gammanorm$ is the
finite sample version of $\|\grad_\T f\|_{L_2(\M)}$, integrated
w.r.t. the data density on $\M$.
We apply this normalization to coordinate functions $\phi_k$, but it could also be applied to functions $g_j$ when they are defined only on $\M$.
A similar approach was used in \citet{Haufe2009-yt}.

When function $f$ is defined on a neighborhood around $\M$ in $\rrr^D$, we compute the normalizing constant with respect to  $\nabla_\xi f$. That is,
\beq \label{eq:gammaj-discrete-xi}
\gammanorm^2\;=\;\frac{1}{n}\sum_{i=1}^n\| \nabla_\xi f(\xi_i)\|^2.
\eeq
Then, once again, we set $f\leftarrow f /\gammanorm$.
We apply this normalization to our dictionary functions $g_j$.
This favors dictionary functions whose gradients are nearly tangent to the manifold $\M$, and penalizes the $g_j$'s which have large gradient components perpendicular to $\M$.

\paragraph{Tuning}
\label{sec:tuning}
Tuning parameters are often selected by cross-validation in Lasso-type problems.
However, in our setting, the recovered support generally span the tangent space, and as discussed in Section \ref{sec:theory}, we are theoretically motivated to identify a size $d$ support.
Since the cardinality of the support decreases as the tuning parameter $\lambda$ is increased, we thus base our choice of $\lambda$ on matching the cardinality of the support to $d$.
Sufficient conditions for this estimation strategy are given in Section \ref{sec:theory}.
To identify this $\lambda$, which we call $\lambda_0$, we perform a simple binary search over $\lambda$ in the range $[0, \lambda_{\text{max}}]$ where $\lambda_{\text{max}}$, the theoretical maximum $\lambda$ value, is $\lambda_{\text{max}} = \max_{j} ( \sum_{i=1}^n \sum_{k = 1}^m (\grad_{T_i^\M} g_j (\xi_i))^T (\grad_{T_i^\M} \phi_m (\xi_i)))^{1/2}$.

\paragraph{Variants and extensions}
\label{sec:variants}
The \ouralg~algorithm presented here can be extended in several interesting
ways. 

First, our current approach explains the embedding coordinates
$\phi$ produced by a particular embedding algorithm. However, the same
approach can be used to directly explain the tangent subspace of $\M$,
independently of any embedding.
Second, one could set up \flassoalg~problems that explain a single
coordinate function.  In general, manifold coordinates may not have
individual meaning, so it will not always be possible to find a good
explanation for a single $\phi_k$.  However, Figure \ref{fig:molecs} shows
that for the ethanol molecule, whose manifold is a torus, there exists
a canonical association of certain coordinates to particular torsions.

It is well-established in the support recovery and sparse coding literature \citep{Chen2001-hh, Hesterberg2008-hm, Breheny2011-ci, Lederer2015-pr, Hastie2015-sl} that at large $\lambda$, shrinkage can cause problems including variable selection inconsistency; and furthermore that intermediate $\lambda$ values can have desirable properties as a variable pruning rather than selection step.
Therefore, as a third possible extension of this work, in future research we plan to pursue a combination of the Group Lasso formulation \eqref{eq:flasso-manifold} with Group Sparse Basis Pursuit \citep{Qu2018-ty}. 
The so-called basis pursuit problems \citep{Chen1998-bm}, intimately related to regularized regression, are discussed in more detail in Appendix \ref{app:gsbp}.
 In the case of \ouralg, the corresponding basis pursuit problem is
\beq
\label{eq:bp}
\arg \min_{\beta :s = d } \sum_{j = 1}^p \|\beta_j\| s.t. \grad_{T_i^\M} \phi_k (\xi_i) = \sum_{j = 1}^p \beta_{ijk}  \grad_{T_i^\M} g_j (\xi_i) \quad \text{for all $i=1:n$, and $k=1:m$.}
\eeq
This problem is evidently not tractable, as it involves searching over all $d$-sets of dictionary functions. We suggest, following \citet{Hesterberg2008-hm}, to initially use an intermediate $\lambda$ values in \ouralg, in order to prune the dictionary to a smaller size. 
Subsequently, we can solve problem \eqref{eq:bp} with the pruned dictionary. 

\comment{Third, in future work we plan to extend the approach to features whose gradients cannot be computed analytically, or which do not depend explicitly on $\xi$, such as measurements taken separately from $\xi$. \mmp{confusing...}
\mmp{
 we plan to examine more closely cases such as the above. It is known \citep{AbrahamMarsden} that Hamiltonian systems are associated with tori; hence, it is natural to consider dictionaries of functions $g:\M\rightarrow S^1$.}}

\comment{\mmp{what is this?}Fourth, the pushforward metric associated with each of
the individual functional covariates can be estimated with respect to
the data. This enables estimation of sparse functional support that is
more robust to curvature than our current approach, , where the
codomain of $g_j$ is assumed to have identity metric, and is analogous
to finding a flat parameterization of $\M$. Normalization is
equivalent to saying that the metric on the codomain of the dictionary
functions is unknown by a constant factor, while the definition of
norm we use assumes that the metric on the codomain cancels out the
metric on $\phi(\M)$.}

%% file: gradients-jmlr-related.tex
\section{Related work}
\label{sec:related}
\comment{To our knowledge, ours is the first solution to estimating a function $f$ as a {\em non-linear} sparse combination of functions in a dictionary. Below we cite some of the closest related work.... check Rudy2019-tk... yes, they did it}


We draw a firm distinction between our approach and purely non-parametric methods that attempt to learn a parameterization of $\M$.
For example, the early works of
\citet{saulRoweis:03jmlr} and \citet{tehRoweis:nips} (and references therein)
propose parametrizing the manifold by finite mixtures of local linear
models, aligned so as to provides global coordinates, in a way
reminiscent of Local Tangent Space Alignment \citep{ZhangZ:04}.
Another idea is to use $d$ eigenfunctions of
the Laplace-Beltrami operator $\Delta_{\M}$ as a parametrization of
$\M$. 
Hence, the Diffusion Maps coordinates could be considered such a
parametrization \citep{coifman:06,CoiLafLeeMag05,Gear2012-rj}.
However, these are not in and of themselves interpretable, and it is not clear how many such coordinates are needed \citep{Chen2019-rm}.
 In \citet{mohammedNarayanan:localpcs17}, it was shown that principal curves
and surfaces can provide an approximate manifold parametrization.
These methods can often be used as embedding algorithms in our approach, but make no attempts at synergizing with an interpretable dictionary.
 \citet{Dsilva2018-dz} tackle the
related problem of choosing among the infinitely many Laplacian
eigenfunctions $d$ which provide a $d$-dimensional parametrization of
the manifold. Their approach is to solve a set of Local Linear Embedding
\citep{roweis:00} problems, each aiming to represent an eigenfunction
as a combination of the preceding ones. Similarly, \citet{Chen2019-rm} is another method for reducing the number of "covarying" eigenfunctions.
However, these methods fail to provide physical meaning for the selected functions.

Our work differs from the above entirely non-parametric methods in two key ways: (1) the explanations we obtain are endowed with the meaning of the
domain specific dictionaries, (2) less obviously, descriptors like
principal curves or Laplacian eigenfunctions are generally still
non-parametric (i.e exist in infinite dimensional function spaces),
while the parameterizations by dictionaries we obtain (e.g. the
torsions) are in finite dimensional spaces.
This distinction is mirrored in comparison with the many so-called {\em dictionary learning} methods in which a low-dimensional transformation is learned simultaneously with its inverse.
We note that our method is not dictionary learning per se, but rather sparse coding, in which the dictionary is given \citep{Szabo2011-og}.

The {\em symbolic regression} methods of \citet{Brunton-2016dt}, \citet{Rudy2019-tk}, and \citet{Champion2019-lu} for estimating governing laws of dynamical systems are perhaps most similar to this work.
These methods use sparse regression with respect to a dictionary and the idea of differential composition.
Their goal is to identify the functional equations of non-linear dynamical systems by regressing the time derivatives of the state variables on a subset of functions in the dictionary selected using a sparsity inducing penalty.
This provides a natural interpretability.
However, although these methods can loosely be considered univariate analogs, they do not consider the multidimensional data-manifold, and their synergies with dimension-reduction algorithms are developed in separate directions.

With respect to sparse regression, the seminal group lasso paper of \citet{Yuan2006-af} and support recovery analyses of \citet{Elyaderani2017-ce,Wainwright:2009sharp} are central to our approach.
However, our use of replicates in experiments is reminiscent of the Stability Selection method of \citet{Meinshausen2010-nr}.
Such methods address instabilities of the variable selection, in particular, when restrictive theoretical conditions are violated \citep{Zhao2006-od,Huan_Xu2012-iu}.  The empirically-based two-stage OLS-hybrid approach we elucidate in Appendix \ref{app:gsbp} for resolving this issue is based on ideas in \citet{Efron2004-eb,Meinshausen2007-ey, Hesterberg2008-hm}. Some attractive alternate approaches to this problem that we do not pursue are the use of non-convex penalties such as SCAD \citep{Fan2001-nv, Breheny2011-ci} and weighted data points in the Adaptive Lasso \citep{Zou2006-bj}.  We specifically note the method of \citet{Haufe2009-yt}, which applies group lasso to analyze sparse decomposition of vectors fields, albeit in a different setting.

As for our method, gradient estimation on manifolds is typically derived from the perspective of local linear regression and tangent space estimation \citep{Mukherjee2006-kt, Aswani2011-kd}.
However, as in \citet{Luo2009-mp}, we make explicit the logarithmic map by estimating and projecting upon the tangent space of $\phi(\M)$, and our estimates of this tangent space are made using the pushforward metric of \citet{Perraul-Joncas2013-uq}.

The role of our work with in the molecular dynamics literature is particularly relevant to enhanced sampling methods \citep{rohrdanzZhengMaggioniClementi:11, rohrdanzZhengClementi:mountain-passes13, Fiorin2013-fl,Fleming2016-nu}.
In these methods, exploration of the molecular state space is accelerated through biasing of simulation towards directions of large scale variation, which are typically identified through visual inspection.
Note that this method is practically useful despite the need to perform initial simulations in order to identify collective coordinates. 
More recently, reinforcement-learning type syntheses of these ideas have been applied \citep{Wang2019-mr, Pant2020-cz, Sidky2020-sn}.

Although in our application our dictionary consists of functions with physical meaning, our general principal of finding parametric geometrically-motivated approximations of learned representations is relevant to a range of machine learning contexts.
Examining functions in embedding coordinates is quite typical in genomics \citep{Amir2013-px}, and much deep learning work also makes use of explicit traversal of a latent space \citep{Lin2019-zu,Shukla2019-mn}.
It is also known in a range of settings that learned gradients provide interpretable \citep{Adebayo2018-rc} or otherwise statistically-useful information \citep{Wu2010-jj, Constantine2014,Yang2020-de}.
In \cite{Yang2020-de}, the gradient of the loss actually forms a type of tangent space estimator. 
However, our approach relies on the classical weighted local PCA method for tangent space estimation \citep{Joncas2017-kn, Aamari2019-dg}.
Improvement of this estimator in the presence of noise is an active area of research \citep{Puchkin2019-cq}.

%% file: gradients-jmlr-theory.tex
\section{Theoretical results}
\label{sec:theory}

Here we first study the conditions under which $f=h\circ g_S$ can be represented over a dictionary $\G$ that contains $g_S$. Not
surprisingly, we will show that these are \emph{functional independency}
conditions on the dictionary. Subsequently, we prove recovery conditions in the finite sample case. 

\subsection{Functional dependency}
\label{sec:existence}

We first study when a set of functions on an open subset
$U\subset\mathbb{R}^d$ can be \emph{almost smoothly} represented with a subset of
functionally independent functions. The following lemma implies that
if a set of non-full-rank smooth functions has a constant rank in a
neighborhood, then locally we can choose a
subset of these functions such that the other functions can be smoothly
represented by them. This is a direct result from the constant rank
theorem.

\begin{lemma}[Remark 2 after \cite{Zorich} Theorem 2 in Section 8.6.2]
Let $f:U\rightarrow \rrr^m$ be a mapping defined in an open neighborhood
$U\subset \rrr^d$ of a point $x^\star\in \rrr^d$. Suppose $f\in C^\ell$, the rank
of the mapping $f$ is $k$ at every point in $U$, and
$k<m$. Moreover, assume that the principal minor of order $k$ of the
matrix $Df$ is not zero at $x^\star$. Then in some neighborhood $U_{x^\star}\subset U$ there
exist $m-k \ C^\ell$ functions $g_i,i=k+1,\cdots,m$ such that for any $x=(x_1,\cdots,x_d)\in U(x^\star)$,
\begin{equation}
 	f_i(x_1,x_2,\cdots,x_d) = g_i(f_1(x_1,x_2,\cdots,x_d),f_2(x_1,x_2,\cdots,x_d),\cdots,f_k(x_1,x_2,\cdots,x_d)).
 \end{equation} 
 \label{lem:representation}
\end{lemma}

Applying this lemma we can construct a local representation of a
subset in $g_S$. Partitions of unity enable us to expand the above lemma from local to global. Mathematically, a {\em smooth partition of unity subordinate to $\{U_\alpha\}$} is an indexed family $(\psi_\alpha)_{\alpha\in A}$ of smooth functions $\psi_\alpha:\mathcal{M}\rightarrow \rrr$ with the following properties:
\begin{enumerate}[(i)]
	\item $0\leq \psi_\alpha(\xi)$ for all $\alpha \in A$ and all $\xi\in \mathcal{M}$;
	\item supp $\psi_\alpha \subset U_\alpha$ for each $\alpha\in A$;
	\item Every $\xi\in\M$ has a neighborhood that intersects supp $\psi_\alpha$ for only finitely many values of $\alpha$;
	\item $\sum_{\alpha \in A}\psi_{\alpha}(\xi)=1$ for all $\xi\in \M$.
\end{enumerate}

\begin{lemma}[\cite{smoothmfd} Theorem 2.23]
Suppose that $\M$ is a smooth manifold, and $\{U_\alpha\}_{\alpha\in A}$ is any indexed open cover of $\M$. Then there exists a {smooth} partition of unity subordinate to $\{U_\alpha\}$.
\label{lem:partition}
\end{lemma}

Now we state our main results.
\begin{theorem}\label{thm:uni} Assume $\mathcal{G}=\{g_i\}_{i=1}^p$ is the dictionary where $g_{1:p}$ are $C^\ell$ functions in an open set $U \subset \mathbb{R}^d$. For a subset $S\subset[p]$, let $g_S$ is defined as $\{g_i:i\in S\}$.  Consider $S'\subset [p],S'\neq S,|S'| < d$ such that $\rank Dg_{S'}=|S'|$ at a point.  Suppose that $\ell\geq d+1$. Then there exists a function $\tau:\rrr^{|S'|}\rightarrow \rrr^{|S|}$ that is almost everywhere $C^\ell$ on the range of $g_{S'}$, w.r.t. Lebesgue measure on $\rrr^{|S'|}$, such that $g_S = \tau \circ g_{S'}$ if
\begin{equation} \label{eq:rank}
	\rank\begin{pmatrix} D{g_S} \\ D{g_{S'}} \end{pmatrix} = \rank D{g_{S'}} \ \text{on} \ U
\end{equation}
holds globally. If  $\tau$ is smooth everywhere on the range of $g_{S'}$, then $\eqref{eq:rank}$ holds globally.
\end{theorem}
{\bf Proof} 
First, we show the existence of $\tau$. We claim that it suffices to prove the existence of function composition on the set where $\rank Dg_{S'} = |S'|$.
Consider $U=U_1\cup U_2$, where $U_1:=\{x: \rank Dg_{S'} = |S'|\}$, and $U_2 = U-U_1$.  $U_1$ is not empty by the assumption. Note that we can select an $|S'|\times |S'|$ submatrix $A_{S',\xi}$ in $Dg_{S'}$ and $\det A_{S',\xi}$ is a continuous function (and thus nonzero) in a neighborhood. This shows that $U_1$ is a nonempty open set. Locally, $g_{S'}$ is a diffeomorphism to its image; therefore $g_{S'}(U_1)$ contains an interior point, and thus has positive measure in $\rrr^{|S'|}$. From Sard's theorem \citep{smoothmfd}, we know that the range of $g_{S'}(U_2)$ is of Lebesgue measure zero in $\rrr^{|S'|}$. Therefore it suffices to show that there exists a $\tau\in C^\ell$ on $g_{S'}(U_1)$. To simplify the notation we use $U$ to denote $U_1$ in the following proof. By definition of $U$, we know that $g_{S'}$ is a diffeomorphism between $U$ and $g_{S'}(U)$. So the inverse $g_{S'}^{-1}$ is well defined and $C^\ell$. Also denote $s=|S|$ and $s'=|S'|$.
Let
\begin{equation}
	g_{S'\sqcup S}(\xi) = \begin{pmatrix}g_{S'}(\xi)\\g_{S}(\xi)\end{pmatrix},
\end{equation}
and use $D{g_{S'\sqcup S}}$ to denote the l.h.s. matrix in \eqref{eq:rank}. Here $\sqcup$ means disjoint union. To be specific, we use $g_{j_i}$ to denote the $i-$th function in the collection $[g_{S'};g_S]$
When the rank of $Dg_{S\sqcup S'}$ equals the rank of $D_{g_{S'}}$,
Lemma \ref{lem:representation} implies that there exists some
neighborhood $U_x\in\rrr^d$ of $x$ and $C^\ell$ functions
$\tau_x^i:\rrr^{|S'|}\rightarrow \rrr,i=s'+1,s'+2,\cdots,s'+s$ such that
\begin{equation}
	g_{j_i}(\xi) = \tau_x^i(g_{j_1}(\xi),\cdots,g_{j_{s'}}(\xi))=\tau_x^i(g_{S'}(\xi)),\ \text{for} \ i=s'+1,s'+2,\cdots,s'+s, \ \xi\in U_x.
\end{equation}
Here we should notice that $\tau_x^i$ is defined only on $g_{S'}(U_x)$. Since this holds for every $x\in U$, we can find an open cover $\{U_x\}$ of the original open set $U$. Since each open set in $\rrr^d$ is a manifold, the result of partition of unity in Lemma \ref{lem:partition} holds, namely that $U$ admits a smooth partition of unity subordinate to the cover $\{U_x\}$. We denote this partition of unity by $\psi_x(\cdot)$.
\par
Hence we can define 
\begin{equation}
\tau^i(y) = \sum_{x\in U} \psi_x(g_{S'}^{-1}(y))\tau_{x}^i(y),\quad y\in g_{S'}(U).
\end{equation}
where $\tau^i$ is a function mapping from $g_{S'}(U)\rightarrow \rrr$.
For each fixed $x \in U$, the function
$y\rightarrow \psi_x(g_{S'}^{-1}(y))\tau_x^i(y)$ for $y\in g_{S'}(U)$ is $C^\ell$.  According to the
properties of partition of unity, in a local neighborhood of each point, this is a summation of finitely many smooth functions.  Then this $\tau^i$ will be a $C^\ell$ function on $g_{S'}(U)$. Also, by $1=\sum_x \psi_x(\xi)$, it holds that $\tau^i(g_{S'}(\xi))=g_{j_i}(\xi)$ for any $i=s'+1,\cdots,s'+s$.
\par Therefore,
globally in $U$ we have
\begin{equation}
	 g_{S\sqcup S'}^i(\xi) = \tau^i(g_{1}(\xi),\cdots,g_{s'}(\xi)),\ \text{for} \ i=s'+1,s'+2,\cdots,s'+s,\ \xi \in U.
\end{equation}

\par Now we prove the reverse implication. If $\rank Dg_{S\sqcup S'}>\rank Dg_{S'}$, then
there is $j\in S$, so that $Dg_j\not\in \rowrange Dg_{S'}$.  Pick
$\xi^0\in U$ such that $Dg_j(\xi^0)\neq 0$; such an $\xi^0$ must exist
because otherwise it will be in $\rowrange Dg_{S'}$. By the theorem's
assumption, $Dg_{S}=D\tau Dg_{S'}$. This implies that $(Dg_S)^T$ is in
$\rowrange (Dg_{S'})^T$ for any $\xi$. But this is impossible at
$\xi^0$. \hfill $\BlackBox$
\par

This theorem essentially gives a condition for the existence of the explanation. Further, if $S$ is the set found by \ouralg, then checking that there is no subset satisfying the rank condition implies that the explanation is unique in the dictionary. We say that a set of functions $g_S$ on a metric space $X$ is \emph{$C^\ell$ (smooth) functionally dependent} at $\xi$ if there is a subset $S'\subset S,S'\neq S$, a function $\tau:\rrr^{|S'|}\rightarrow \rrr^{|S|}$ and a neighborhood $U$ around $\xi$  such that
\begin{enumerate}[(i)]
\item $g_S = \tau \circ g_{S'}$ on $U$;
\item $\tau$ is $C^{\ell}$ (smooth) globally on $g_{S'}(U)\subset\rrr^{|S'|}$;
\item $y-\tau (y_{S'})\not\equiv 0$ on any neighborhood $O(g_S(\xi))\subset \rrr^{|S|}$. Here $y=(y_1,\cdots,y_{|S|})\in \rrr^{|S|},y_{S'}=(y_i)_{i\in S'}\in \rrr^{|S'|}$.
\end{enumerate}
The condition (iii) here eliminates the possibility of a trivial $\tau$.
$S$ is {\em functionally independent} if it is nowhere functionally dependent. Based on Theorem \ref{thm:uni}, we formulate the rank condition below as a necessary and sufficient condition of functional independence.
\begin{corollary}[Functional Independence]
\label{cor:funind}
Suppose $\mathcal{M}$ is a $d-$dimensional smooth manifold and $g_{S}: \mathcal{M}\rightarrow \rrr^d$ are $d \ C^\ell$ functions. Suppose $g_S(\cal M)$ has a positive measure in $\rrr^d$. Then they are functionally independent on $\mathcal{M}$ iff  $\rank Dg_S(\xi)$ is $d$ everywhere on $\mathcal{M}$ except for a closed subset $W\subset\mathcal{M}$ with no interior point.
\end{corollary}
{\bf Proof}
First we show that the rank condition implies functional independence.
Suppose $g_S$ is functionally dependent. Then by definition we have that $g_S = \tau \circ g_{S'}$ on a neighborhood for some $S'$ with $|S'| < |S|=d$. Then on this neighborhood $\rank Dg_S \leq \rank Dg_{S'} \leq d-1$. This contradicts the assumption. On the other hand, suppose $Dg_S$ is functionally independent. We claim that for any $\xi' \in V_0$, $\rank Dg_S(\xi') \geq \rank Dg_S(\xi)$: For any $\xi$ there is a $\rank Dg_S \times \rank Dg_S$ non-degenerate submatrix of $Dg_S(\xi)$ whose determinant is non-zero. Therefore, by the smoothness of $g_S$, there exists a neighborhood $V_0$ such that this submatrix of the Jacobian is invertible in this neighborhood and the claim holds.

We therefore start from a point $\xi$ where $\rank Dg_S = d-1$. Select functions that are full rank at this point, and denote them by $g_{S'}$. There is a neighborhood $V_1$ of $\xi$ with $\rank g_{S'}(\xi)=d-1$. If $\rank Dg_S  = d-1$ holds on some neighborhood $V_2$ of $\xi$, then after selecting a chart $(U,\varphi)$ containing $V_1\cap V_0$,  we have that
\begin{equation}
	\rank\begin{pmatrix} D{g_S\circ \varphi^{-1}} \\ D{g_{S'}\circ \varphi^{-1}} \end{pmatrix} = \rank D{g_{S'}\circ\varphi^{-1}}
\end{equation}
holds on $V_2\cap V_1\cap V_0$. Thus, Theorem \ref{thm:uni} implies that $g_S$ cannot be functionally independent (consider a composition with $\varphi$). Therefore, the set $\{x: \rank Dg_S = d-1\}$ has empty interior in $\mathcal{M}$.
Similarly, in every neighborhood $V_{k}$ of any point where $0\leq \rank Dg_S=k < d-1$, there must be a point such that $\rank Dg_S = k+1$. Then in every neighborhood $V_{k+1}\cap V_{k}$ of this new point there must be a point such that $\rank Dg_S = k+2$.  By induction, there must be a point in $V_k$ such that $\rank Dg_S=d$. Therefore we conclude that the set $W=\{\xi:\rank Dg_S \leq d-1\}$ contains no interior point. Also, it is closed because $\{\xi: \rank Dg_S = d\}$ is open.
\hfill $\BlackBox$

\begin{theorem}
\label{thm:mfd}
Let $\mathcal{G}$ and $g_S$ be defined as before. $\mathcal{M}$ is a smooth manifold with dimension $d$ embedded in $\rrr^D$. Suppose that $\psi: \mathcal{M}\subset \rrr^D \rightarrow \rrr^m$ is also an embedding of $\mathcal{M}$ and has a decomposition $\psi(\xi) = h\circ g_S(\xi)$ for every $\xi \in \mathcal{M}$ where $h$ is smooth.
If the dictionary $g_S$ contains $d$ functions denoted by $g_{S'}$,  that are smooth functionally independent on $\cal M$, then there exists a $\widetilde{h}$ such that $\psi = \widetilde{h}\circ g_{S'}$ on every $\xi \in \mathcal{M}$. Here, the
function $\widetilde{h}$ is smooth almost everywhere in the range of $g_{S'}$.
\end{theorem}

{\bf Proof}
Consider the set $U=\{x: \rank Dg_{S'}=d\}$. It is an open subset of the manifold $\cal M$ and therefore a smooth manifold. For each point $x\in U$, select a local chart $(V,\varphi)$ such that $V\subset U$. With the same argument in the proof of Corollary \ref{cor:funind}, we know that there exists a smooth functions $\tau_x$ on $V$ such that $g_S = \tau \circ g_{S'}$ holds on V. Also, since $V$ is an open neighborhood, we conclude that the measure of $g_{S'}(U)\geq g_{S'}(V)$ should be strictly positive. Therefore the partition of unity technique used in the proof of Theorem \ref{thm:uni} can show that there exists a function $\tau$ on $U$ that is smooth over $g_{S'}(U)$ such that $g_S = \tau \circ g_{S'}$ holds globally on $U$. We can define $\tau$ on $\mathcal{M}\setminus U$ to be anything, and Sard's theorem implies that $g_{S'}(U)$ would be a measure zero set in $\rrr^d$.  Finally, we just write $\widetilde{h}=h\circ \tau $
\hfill $\BlackBox$

The assumptions of these theorems are reasonable. 
Even though any smooth map $f:\mathcal{M}\rightarrow \rrr^d$ on a compact manifold $\mathcal{M}$ must have at least one singular point, Theorem \ref{thm:uni} will still hold almost everywhere as long as $g_S$ is smooth almost everywhere.
The existence of such functions $g_S$ with rank $d$ almost everywhere is guaranteed by the fact that a single coordinate chart can cover any compact manifold except for a set of measure zero, known as the cut-locus of the chart \citep{ShengW:09, Bishop2013-sr}.
One can, for example, find one function explaining the whole circle $S^1$ embedded in $\rrr^2$ except one point.
Thus, these theoretical results should be considered conditions on the dictionary, rather than the manifold itself.



\subsection{Discussion of practical recovery from samples}
\label{sec:discrete}
In a finite sample setting, Theorem \ref{thm:uni} states that $S$ and
$S'$ are equivalent explanations for $f$ whenever \eqref{eq:rank}
holds on open sets around the sample points. In this situation, it is
very likely to find many subsets $S'\subset [p]$ of cardinality $d$
that are full rank in neighborhoods of all data points. Still assuming that all
gradients are exact, for all such $S'$ the first term of $J_\lambda(\beta)$ in \eqref{eq:flasso-manifold} will be zero; in other words there will be many equivalent explanations of $\phi$ in
$\G$.
\comment{Not true: "The solution to (\ref{eq:flasso-manifold}) will be the
subset $S$ that minimizes the loss term
$\sum_{j=1}^p\|\beta_j\|$ on the data set."}
However, one can define a subset $S$ as the expected minimizer in \eqref{eq:bp}, or in a purely oracle sense.
For our theoretical analyses, we thus fix a subset $S$ and regard it as the true support throughout this section.

The tendency of \ouralg~ to select a support with a low value of $\sum_{j=1}^p\|\beta_j\|$ is reasonable, and even desirable because, according to \cite{GO2011}, a particular group $S$ will be recovered by Group Lasso methods, if (i) it is close to perpendicular to the linear subspace generated by all other groups, and (ii) group features in $S$ are close to orthogonal matrix. The first condition will be discussed later in this Section. As for condition (ii), we note that if a set $S'$ is not full rank on $\M$, the Jacobian $Dg_{S'}$ will be ill-conditioned at the data near the critical points, which will result in very large $\beta_{ji}$ values. Hence, such a subset will be heavily penalized. Moreover, features $g_j$ which vary much in a direction normal to $\M$ will have, due to the gradient normalization, smaller values for $\grad_{T_i^\M} g_j$; therefore their $\beta_j$ coefficients will be large relatively to the coefficients of functions that vary within $\M$. 

We now analyze the situation when $\grad_{T_i^\M} g_j$ and $\grad_{T_i^\M} \phi_k$
are estimated with noise, showing that it is qualitatively similar to
noiseless case. Specifically, we provide recovery guarantees for the
(\flasso) problem that highlight the influence of the aforementioned
factors, as well as of condition (i). The guarantees are
deteriministic, but they depend on the noise sample variance, hence
they can lead to statistical guarantees holding w.h.p. in the usual
way. For simplicity, we analyze support recovery for $m=1$, hence for
a single dependent variable $y$. Namely we assume that that the data
$y_{1:n}\in \rrr^d$ satisfy
\beq \label{eq:what}
y_i\;=\;\sum_{j=1}^p \beta_{ij}^*x_{ij}+\epsilon_i\quad \text{for $i=1:n$}
\eeq
and we rewrite the \flasso~problem as
\beq 
\min_\beta\tfrac{1}{2}\sum_{i=1}^n||y_i-X_i
\beta_i||_2^2+\lambda\sqrt{n} \sum_{j=1}^p ||\beta_j||,
\label{eq:J0}
\eeq
according to the notations of equation \eqref{eq:flasso-manifold} with the index $k$ dropped. 
A first theorem deals with support recovery, proving that all
coefficients outside the support are zeroed out, under conditions that
depend only on the dictionary, the fixed support $S$ and the noise.
The second result completes the previous with error bounds on the
estimates $\bhat_j$, assuming an additional condition on the magnitude
of the true $\beta_j$ coefficients.  Since these coefficients are
partial derivatives w.r.t. the dictionary functions, the condition
implies that the dependence on each function must be strong enough to
allow the accurate estimation of the partial derivatives.

We introduce the following quantitites. The {\em incoherence} of $\G$ is defined as
\beq \label{eq:mu}
\mu\;=\;\max_{i=1:n,j\in[p],j'\in S,j\neq j'}\frac{|x_{ij}^Tx_{ij'}|}{\|x_{ij}\|\|x_{ij'}\|}.
\eeq
This definition differs in two ways from the standard definition of
incoherence as $\max_i\max_{j,j'\in [p]}|x_{ij}^Tx_{ij'}|$. First, here we
do not make the common assumption that the columns of the \flasso~design
matrix are norm 1 (details to be found in the proof of Theorem \ref{thm:S}). Rather, the recovery result
we pursue assumes the normalizations in Section
\ref{sec:ouralg-normalization}; hence to preserve a measure of
incoherence independent of the column norms, we must rescale by
$\|x_{ij}\|\|x_{ij'}\|$. Second, because we condition on the set $S$,
it is not necessary to require that the gradients outside the support
$S$ be incoherent. 
\comment{However, the incoherence, in other words near
orthogonality, between gradients in the support and outside, as
well as within $S$, still matter, and are measured by $\mu$.}

We further consider the internal collinearity of the support $S$ as follows. Let
\beq\label{eq:Sigma}
\Sigma_i\;=\;\left[x_{ij}^Tx_{ij'}\right]_{j,j'\in S}
\quad 
\text{and}
\quad\Sigma\;=\;\diag\{\Sigma_{1:n}\}.
\eeq
\begin{lemma}\label{lem:sigma--1}
\[
\|\Sigma_i^{-1}\|\,\leq\,\frac{1}{(\min_{j\in S}\|x_{ij}\|^2)[1-(s-1)\mu]}
\quad\text{for all }i=1:n.
\]
\end{lemma}
{\bf Proof} 
It is easy to see that 
\beq
\Sigma_i\;=\;\diag\{\|x_{ij}\|_{j\in S}\}\tilde{\Sigma}_i\diag\{\|x_{ij}\|_{j\in S}\}
\quad
\text{with}
\quad
\tilde{\Sigma}_i\;=\;\left[\frac{|x_{ij}^Tx_{ij'}|}{\|x_{ij}\|\|x_{ij'}\|}\right]_{i,j\in S}\,.
\eeq
By the Gershgorin Theorem, since all the off-diagonal elements of $\tilde{\Sigma}_i$ are bounded in absolute value by $\mu$,  the minumum eigenvalue of $\tilde{\Sigma}_i$ is bounded below by $1-(s-1)\mu$. When this quantity is positive, then the maximum eigenvalue of $\tilde{\Sigma}_i^{-1}$ is 
\beq
\|\tilde{\Sigma}_i^{-1}\|\,\leq\,\frac{1}{1-(s-1)\mu}\,=\,\nu.
\eeq
A smaller $\nu$ means that the $x_{ij}$ gradients are closer to being orthogonal at each datapoint $i$. Furthermore, $\|\Sigma_i^{-1}\|\leq \|\tilde{\Sigma}_i^{-1}\|\|\diag\{\|x_{ij}\|^{-1}_{j\in S}\}\|^2\leq \frac{\nu}{\min_{j\in S}\|x_{ij}\|^2}$. \hfill $\Box$

Finally, the noise level $\sigma$ is defined by
\beq
\max_{i=1:n}||\epsilon_i||^2=d\sigma^2.
\eeq
\begin{theorem}[Support recovery]\label{thm:S} Assume that equation (\ref{eq:what}) holds, and that $\sum_{i=1}^n||x_{ij}||^2=\gamma_j^2$ for all $j=1:p$.  Let $\gammamax=\max_{j\not \in S}\gamma_j$, $\kappa_S=\max_{i=1:n}\frac{\max_{j\in S}\|x_{ij}\|}{\min_{j\in S}\|x_{ij}\|}$. 
Denote by $\bbar$ the solution of (\ref{eq:J0}) for some $\lambda>0$.
If $1-(s-1)\mu >0$ and 
\beq \label{eq:flasso-mani-condition}
\gammamax\left(\frac{\mu}{1-(s-1)\mu}\frac{\kappa_S}{\min_{i=1}^n\min_{j'\in S}\|x_{ij'}\|}+\frac{\sigma \sqrt{d}}{\lambda\sqrt{n}}\right)
\;\leq\;1
\eeq
then $\bbar_{ij}=0$ for $j\not \in S$ and all $i=1,\ldots n$.
\comment{$y_i\;=\;\xb_i\beta_i+w_i\in \rrr^d$ with $E[w_i]=0,Var(w_i)=\sigma^2$.}
\end{theorem}

\noindent{\bf Proof} 
We structure equation \eqref{eq:what} in the form
\beq
y\;=\;\barbx \bbar^* +\barw
\quad\text{with }y=[y_i]_{i=1:n}\in\rrr^{nd},\;
\bbar=[\beta_i]_{i=1:n}\in\rrr^{np},\;
\eeq
$\tilde{X}_{ij}\in\rrr^{nd}$ is obtained from $x_{ij}$ by padding with
zeros for the entries not in the $i$-th segment,
$\barbx=[[\tilde{X}_{ij}]_{j=1:p}]_{i=1:n}\in \rrr^{nd\times np}$,
and $\barbx_j=[\tilde{X}_{ij}]_{i=1:n}\in \rrr^{nd\times n}$ collects
the colums of $\barbx$ that correspond to the $j$-th dictionary
entry. Note that
\beq
\tilde{X}_{ij}^T\tilde{X}_{ij'}\;=\;x_{ij}^Tx_{ij'}
\quad \text{and} \quad
\tilde{X}_{ij}^T\tilde{X}_{i'j'}\;=\;0\text{ whenever }i\neq i'.
\eeq
The proof is by the {\em Primal Dual Witness} method, following \cite{Elyaderani2017-ce,GO2011}. 
It can be shown \cite{Elyaderani2017-ce,Wainwright:2009sharp} that $\bbar$ is a solution to (\flasso) iff, 
\text{for all $j=1:p$},
\beq \label{eq:kkt0} 
\barbx_j^T\barbx(\bbar -\bbar^*)-\barbx_j^T\barw+\lambda z_j\;=\;0\;\in\rrr^n
\text{ with }z_j=\frac{\beta_j}{||\beta_j||}
\text{ if }\beta_j\neq 0 \text{ and $||z_j||<1$ otherwise}.
\eeq
The matrix $\barbx_j^T\barbx$ is a diagonal matrix with $n$ blocks of size $1\times p$, hence the first term in \eqref{eq:kkt0} becomes 
\beq
[ x_{ij}^TX_i(\bbar_{i:}-\bbar_{i:}^*)]_{i=1:n}\in \rrr^{n}.
\eeq
Similarly $\barbx_j^T\barw=[x_{ij}^T\epsilon_i]_{i=1:n}\in \rrr^n$.

We now consider the solution $\bhat$ to problem \eqref{eq:J0} under
the additional constraint that $\beta_{ij'}=0$ for $j'\not \in S$. In
other words, $\bhat$ is the solution we would obtain if $S$ was
known. Let $\zhat$ be the optimal dual variable for this problem, and
let $\zhat_S=[\zhat_j]_{j\in S}$.

We will now complete $\zhat_S$ to a $z\in \rrr^{np}$ so that the pair $(\bhat, z)$ satisfies \eqref{eq:kkt0}. If we succeed, then we will have proved that $\bhat$ is the solution to the original \flasso~problem, and in particular that the support of $\bhat$ is included in $S$. For simplicity we denote $\lambda'=\lambda\sqrt{n}$. 

From \eqref{eq:kkt0} we obtain values for $z_j$ when $j\not \in S$.
\beq\label{eq:zjcomp}
z_j\;=\;\frac{-1}{\lambda'}\barbx_j^T\left[\barbx^T(\bhat-\bbar^*)-\barw\right].
\eeq
In the same time, if we consider all $j\in S$, we obtain from \eqref{eq:kkt0} that $\barbx_S=[\barbx_j]_{j\in S}$ (here the vectors $\beta_S,\beta_S*$ and all other vectors are size $ns$, with entries sorted by $j$, then by $i$). 
\beq \label{eq:beta-err}
\barbx_S^T\barbx_S(\bhat_S-\beta_S^*)-\barbx_S^T\barw+\lambda' \zhat_S\;=\;0.
\eeq
Solving for $\bhat_S-\beta_S^*$ in  \eqref{eq:beta-err}, we obtain
\beq
\bhat_S-\beta_S^*\;=\;(\barbx_S^T\barbx_S)^{-1}\left(\barbx_S^T\barw-\lambda' \zhat_S\right)\;=\;\Sigma^{-1}\left(\barbx_S^T\barw-\lambda' \zhat_S\right).
\eeq 
After replacing the above in \eqref{eq:zjcomp} we have
\beq \label{eq:zj2}
z_j\;=\;\frac{-1}{\lambda'}\barbx_j^T\left[\barbx_S\Sigma^{-1}\barbx^T_Sw-\barbx_S\Sigma^{-1}\lambda' \zhat_S-\barw\right]
\;=\;\barbx_j^T\barbx_S\Sigma^{-1}\zhat_S+\frac{1}{\lambda'}\barbx_j^T(I-\barbx_S\Sigma^{-1}\barbx_S^T)\barw.
\eeq
Finally, by noting that $\Pi=I-\barbx_S\Sigma^{-1}\barbx_S^T$ is the projection operator on the subspace $\range(\barbx_S)^\perp$, we obtain that 
\beq \label{eq:zj}
z_j\;=\;(\barbx_j^T\barbx_S)\Sigma^{-1}\zhat_S+\frac{1}{\lambda'}\barbx_j^T\Pi\barw,
\quad\text{for $j\not\in S$}.
\eeq
We must show that $||z_j||<1$ for $j\not\in S$. 
To bound the first term, we note that $\barbx_j^T\barbx_S$ is $n\times ns$, block diagonal, with blocks of size $1\times s$, and with all non-zero entries bounded in absolute value by $\mu$. Hence, for any vector $v=[v_i]_{i=1:n}\in \rrr^{ns}$, 
\beq \label{eq:intermed-1}
||\barbx_j^T\barbx_Sv||^2=||[(x^T_{ij}x_{iS})v_i]_{i=1:n}||^2\leq\sum_{i=1}^n||(x^T_{ij}x_{iS})v_i||^2
\,\leq\,\sum_{i=1}^n\|(x^T_{ij}x_{iS})\|^2\|v_i\|^2.
\eeq
In our case $v_i=\Sigma_i^{-1}\zhat_{iS}$, hence by  by Lemma \ref{lem:sigma--1} 
\beq \label{eq:intermed-1.1}
\|v_i\|\leq ||\Sigma_i^{-1}||||\zhat_{iS}||\leq\nu\frac{1}{\min_{j'\in S}\|x_{ij'}\|^2}.
\eeq
On the other hand, 
\beq \label{eq:intermed-1.2}
\|x_{ij}^Tx_{iS}\|^2\;=\;\sum_{j'\in S}(x_{ij}^Tx_{ij'})^2
\;\leq\;\sum_{j'\in S}\mu\|x_{ij}\|^2\|x_{ij'}\|^2
\;\leq\;\mu\max_{j'\in S}(\|x_{ij'}\|^2)\|x_{ij}\|^2.
\eeq
Bounds \eqref{eq:intermed-1.1} and \eqref{eq:intermed-1.2} together with equation \eqref{eq:intermed-1} yield
\beqa 
(\barbx_j^T\barbx_S)\Sigma^{-1}\zhat_S
&\leq&\nu^2\mu^2\sum_{i=1}^n\|x_{ij}\|^2\max_{j'\in S}(\|x_{ij'}\|^2)\frac{1}{(\min_{j'\in S}\|x_{ij'}\|^2)^2}
\;\leq\;\nu^2\mu^2\kappa_S^2\sum_{i=1}^n\|x_{ij}\|^2\frac{1}{\min_{j'\in S}\|x_{ij'}\|^2}\\
&\leq&\nu^2\mu^2\kappa_S^2\frac{1}{\min_{i=1}^n\min_{j'\in S}\|x_{ij'}\|^2}\sum_{i=1}^n\|x_{ij}\|^2
\;=\;\nu^2\mu^2\kappa_S^2\frac{1}{\min_{i=1}^n\min_{j'\in S}\|x_{ij'}\|^2}\gamma_j^2 \label{eq:intermed-2}
\eeqa
To bound the second term, we note that $\Pi$ is a block diagonal matrix, $\Pi=\diag\{\Pi_{1:n}\}$, with $\Pi_i=I_d-x_{iS}^T\Sigma_i^{-1}x_{iS}$.
Hence, the norm squared of this term is bounded above by
$\sum_{i=1}^n||\Pi_i\epsilon_i||^2||x_{ij}||^2/(\lambda')^2\leq\sum_{i=1}^n||\epsilon_i||^2||x_{ij}||^2/(\lambda')^2\leq d\sigma^2/(\lambda')^2\sum_{i=1}^n||x_{ij}||^2=(d\sigma^2/(\lambda')^2)\gamma_j^2$.

Replacing these bounds in \eqref{eq:zj} we obtain that 
\beq
||z_j||\leq ||\barbx_j^T\barbx_S\Sigma^{-1}\zhat_S||+||\frac{1}{\lambda'}\barbx_j^T\Pi\barw||
\,\leq\,\left( \frac{\mu\nu\kappa_S}{\min_{i=1}^n\min_{j'\in S}\|x_{ij'}\|}+\frac{\sigma \sqrt{d}}{\lambda'}\right)\gamma_j
\;\text{for any $j\not \in S$}.
\eeq
\hfill$\Box$

The first term inside the parenthesis relates to the properties of the support $S$. The factor $\frac{\mu}{1-(s-1)\mu}$ measures the near-orthogonality of the gradients in $S$, while the factors $(\min_{i=1}^n\min_{j'\in S}\|x_{ij'}\|)^{-1}$ and $\kappa_S$ measure the conditioning of $S$ with respect to the gradient norms. They are optimal when all gradients in $S$ are bounded away from $0$, and when their sizes are relatively equal. The second term depends on the noise amplitude, and can be made arbitrarily small by increasing the regularization coefficient $\lambda$. 

We now consider the recovery of all the non-zero coefficients, which will complete the exact support recovery proof. From the result below, we shall see that having non-zero $\bhat_j$ for a $j\in S$ requires that the original $\beta_j$ is large enough w.r.t. noise level and condition numbers of the problem. This conflicts with the requirement that $\beta_j$ is small, suggesting one possible way that the \flasso~recovery may fail, namely that the smallest of the non-zero $\beta_j$'s may be ``regularized out'' before all the nuisance $\bhat_j$ are.

\begin{corollary}\label{thm:beta} Assume that equation \eqref{eq:J0} and condition (\ref{eq:flasso-mani-condition}) hold. Let $\kappa=\frac{\mu}{1-(s-1)\mu}\frac{\kappa_S}{\min_{i=1}^n\min_{j'\in S}\|x_{ij'}\|}$ and $\gamma_S=\|\barbx_S\|$. Denote by $\bhat$ the solution to problem \eqref{eq:J0} for some $\lambda>0$. If (1) $\lambda=c\frac{\gammamax\sigma\sqrt{d}}{1-\kappa\gamma\max}$, $c>1$, and (2) $||\beta_j^*||>\sigma\sqrt{d}(\gammamax+\gamma_S)+\lambda(1+\sqrt{s})$ for all $j\in S$, then the support $S$ is recovered exactly and 
\beq\label{eq:betaerr}
||\bhat_j-\beta^*_j||
<\sigma\sqrt{d}(\gammamax+\gamma_S)+\lambda(1+\sqrt{s})
=\sigma\sqrt{d}\gammamax\left[1+\gamma_S/\gammamax+c\frac{1+\sqrt{s}}{1-\kappa\gammamax}\right]
\quad\text{ for all $j\in S$.}
\eeq
\end{corollary}

{\bf Proof of Corollary \ref{thm:beta}} According to Theorem \ref{thm:S}, $\bhat_j=0$ for $j\not \in S$. It remains to prove the error bound for $j\in S$. According to
Lemma V.2 of \cite{Elyaderani2017-ce}, for any $j\in S$, \mmp{inline this too}
\beqa
||\bhat_j-\beta^*_j|| &\leq
&||\barbx_j^T\barw||+||\barbx_S^T\barw||+\lambda(1+\sqrt{s})\\ &\leq
&(||\barbx_j||+||\barbx_S||)||\barw||+\lambda(1+\sqrt{s})\\ &\leq
&\sigma\sqrt{d}(\gammamax+\gamma_S)+\lambda(1+\sqrt{s})\\ &=
&\sigma\sqrt{d}\gammamax\left[1+\gamma_S/\gammamax+c\frac{1+\sqrt{s}}{1-\kappa\gammamax}\right].
\eeqa
Hence, if
$||\beta_j^*||$ is greater than the r.h.s. of the above, $\bhat_j\neq 0$ and the
support is recovered exactly.  \hfill $\Box$
In equation \eqref{eq:betaerr}, the factor $\sigma\sqrt{d}$ represents
the noise amplitude, while $\gammamax$ bounds the amplitude of the
nuisance covariates $\barbx_j$ outside of $S$. A smaller $\gammamax$
means that the contribution of these nuisance covariates will be
smaller. The term $\gamma_S$ bounds the collinearity of the noise with
the true support covariates. The last term measures the bias
introduced in $\bhat_j$ by the regularization; note that $\lambda$
itself depends on the noise amplitude $\sigma\sqrt{d}$.

Recall from Sections \ref{sec:flasso-manifold} and 
\ref{sec:ouralg-normalization} that $\gamma_j$ represents the finite
sample estimate of the $L_2$ norm of $\grad_{T_i^\M} g_j$. When the
dictionary functions $g_j$ are defined on $\M$, but not outside $\M$,
then $\grad_{T_i^\M} g_j$ is normalized by equation
\eqref{eq:gammaj-discrete-M} and consequently $\gamma_j=\sqrt{n}$ for all
$j$.  If first the gradients $\nabla_\xi g_j$ are computed, then
normaled in ambient space $\rrr^D$ by
\eqref{eq:gammaj-discrete-xi}, after projection on the tangent bundle $\T\M$, $\gamma_j\leq \sqrt{n}$. Thus, by explicitly considering the variability in
the norms of $\|x_{ij}\|$ for $j\not\in S$, we see that features $g_j$
whose gradient $\nabla_\xi g_j$ is not tangent to the manifold are
easier to rule out. Regarding scaling of the l.h.s. of equation \eqref{eq:flasso-mani-condition}, it is easy to see that the term $\frac{\sigma\sqrt{d}\gammamax}{\lambda\sqrt{n}}$ is $\bigOO(1)$ w.r.t. $n$; the first term $\kappa\gammamax$ is invariant to any rescaling of the $\barbx$ by a scalar. 


%% file: gradients-jmlr-exp.tex
\section{Experiments}
\label{sec:exps}
We demonstrate the ability of \ouralg~ to identify explanations of manifolds and their embedding coordinates in both toy and scientific manifold learning problems.
Section \ref{sec:exp_setup} describes the general experimental procedure, while Section \ref{sec:md} describes some specific adjustments to this protocol necessary for analyzing molecular dynamics (MD) data.
Sections \ref{sec:swiss}--\ref{sec:mds_data} describe our experimental results. \footnote{Code to run experiments is available at \url{https://github.com/sjkoelle/manifoldflasso_jmlr}.}


\begin{table}[h]
\begin{center}
\begin{tabular}{ | l | c | c | c| c| c | c | c  |c| c | c |  }
\hline
Dataset & $n$ & $N_a$ & $D$  & $d$  & $\epsilon_N $&   $m$ & $n'$ & $p$ & $\omega $ \\ \hline
\srdata & 10000 & NA & 49 & 2 & .18  &  2 &  100 & 51 & 1 \\
\redata & 10000 & 9 & 50 & 2 & 3.5  &  3 & 100 & 12 & 25\\
\hline 
\ethdata & 50000 & 9 & 50 & 2 & 3.5 & 3 & 100 & 12  & 25\\
\maldata & 50000 & 9 & 50 & 2 & 3.5 & 3 & 100 & 12  & 25 \\
\toldata & 50000 & 16 & 50 & 1 &  1.9   & 2& 100 & 30 &  25\\
\hline
\ethdata & 50000 & 9 & 50 & 2 & 3.5 & 3 & 100 & 756  & 25\\
\maldata & 50000 & 9 & 50 & 2 & 3.5 & 3 & 100 & 756  & 25 \\
\hline
\end{tabular}
\end{center}
\caption{Summary of experiments. \srdata~ and \redata~ are toy data, while \toldata, \ethdata, and \maldata~ are from quantum molecular dynamics simulations by \cite{chmielaTkaSauceSchuPMull:force-fields17}. The columns list the following experimental parameters: $n$ is the sample size for manifold embedding, $N_a$ is the number of atoms in the molecule, $D$ is the dimension of $\xi$,  $d$ is the intrinsic dimension, $\epsilon_N$ is the kernel bandwidth, $m$ is the embedding dimension, $n'$ is the size of the subsample used for \ouralg, $p$ is the dictionary size, and $\omega$ is the number of independent repetitions of \ouralg. More details are in Section \ref{sec:exp_setup}}
\label{tab:exps}
\end{table}

\subsection{Experimental setup}\label{sec:exp_setup}

For all of the following experiments, the data consist of $n$ data points in $D$ dimensions, as well an embedding $\phi_{1:m} (\dataset)$.
We assume access to the manifold dimension $d$, a kernel bandwidth $\epsilon_N$ used in the estimation of the tangent spaces, and $p$ dictionary functions.
Except where otherwise specified, $m$ and $\epsilon_M$ are used in the preliminary step of generating embeddings $\phi_{1:m}$ using the Diffusion Maps algorithm as \embedalg. \ouralg~ is applied to a uniformly random subset of size $n' = |\I|$ and this process is repeated $\omega$ number of times.
These parameters are passed to the \lapalg, \tppalg, \rmalg, and \dpullalg~algorithms, and are summarized in Table \ref{tab:exps}.
The regularization parameter $\lambda$ ranges over $[0, \lambda_{\text{max}}]$ as described in Section \ref{sec:tuning}.
We note that $d<3$ in all cases, and so Theorem \ref{thm:S} applies.

\subsection{Molecular dynamics data}{\label{sec:md}

The method of MD simulations is one of the
principal tools in the study of molecular systems. Such simulations provide detailed information on the fluctuations and
conformational changes of the simulated system, and are now routinely
used to investigate the structure, dynamics and thermodynamics of
biological macromolecules and their complexes. In such simulations, the positions of atoms within a
molecule are sampled as they proceed through time from some initial
conditions according to interatomic effects.
The distribution of this sample describes the molecule's behavior in the given experimental conditions.
It has been shown empirically that manifolds approximate these high-dimensional distributions \citep{Dsilva13-Nonlinear}. 
Furthermore, substantial theoretical work is dedicated to demonstrating this property, i.e. that the states arising from dynamical systems concentrate around a {\em slow manifold} describing the evolution of configurations at long time scales.
Accordingly, application of manifold learning to find what are in this setting called the {\em collective coordinates} has achieved great success.
Even though the vector
of atomic coordinates can take any value, due to interatomic
interactions, the relative positions of atoms within the molecule lie
near a low-dimensional manifold. Performing manifold learning on
these data separates the conformational changes, modeled by
the manifold, from the fluctuations represented by the ``noise''
around the manifold.

\paragraph{Representing molecular configurations} 
Our MD data are quantum-simulations from \cite{chmielaTkaSauceSchuPMull:force-fields17}.
The raw data consists of $X,Y,Z$ coordinates for each of the $N_a$ atoms of the chosen molecule.
For a single observation, we denote these by $r_i \in\rrr^{3N_a}$.
The first step in our data analysis pipeline is to featurize the configuration in a way that is invariant to rotation and translation.
In the present experiments, we follow \cite{ChenMcqueenKoelleMChmielaTkatchenko:mlcules-dum19} and represent a molecular configuration as a vector $a_i \in\rrr^{3 {N_a \choose 3}}$ of the \textit{planar angles} formed by triplets of atoms.
We then perform an SVD on this featurization, and project the data onto the top $D = 50$ singular vectors to remove linear redundancies; we denote the new data points by $\xi_{1:n}$.
The \embedalg~and \tppalg~algorithms work directly with $\xi$ in dimension $D$.
Other possible representations such as applying a Procrustes transform to each configuration to align it with the first one give similar results, and no matter which low level representation we choose, large-scale conformational changes are described by the relative rotations of groups of atoms - the bond torsions illustrated in Figure \ref{fig:molecs} \citep{ChenMcqueenKoelleMChmielaTkatchenko:mlcules-dum19}.
\comment{in chemistry it has been known for a long time that }

\paragraph{Dictionaries for MD data}
Therefore, in the \redata, \ethdata, \maldata, and \toldata~ MD
datasets, we construct dictionaries consisting of bond {\em torsions}.
We then apply \ouralg~to select combinations of these higher-level torsion
features that explain the manifold in the lower-level planar angle
feature space.  Given an ordered
4-tuple of atoms $ABCD$, the torsion $g_{ABCD}$ is the angle of the
planes defined by the locations of $ABC$ and $BCD$.  Note that
$g_{ABCD} \equiv g_{DBCA}\equiv g_{DCBA}\equiv g_{ACBD}$.  Any torsion
$g$ is expressible in closed form as functions of the planar angles
feature vector $a$.  In particular, a torsion $g_{ABCD}$ is a function
of the angles of the triangles $ABC$, $ABD$, $ACD$, and $BCD$, and we
 compute the gradients of the torsions by automatic differentiation
\citep{Paszke2019-us}.

One cannot use the obtained gradients directly in \ouralg, since the angular features overparameterize the molecular {\em shape space
$\Sigma_3^{N_a}$} \citep{addicoatcollins:2010,Kendall1989-hu} of
dimension $D'=3N_a-7$, and off-manifold gradients are therefore not well-defined.
For example, whether one chooses to use triangles $ABC$, $ABD$, and $ACD$, or $ABC$, $ABD$, and $BCD$ to compute $g_{ABCD}$ has no effect on the value of $g_{ABCD}$, but changes the value of the gradient in $\mathbb R^D$.
We therefore project the gradients on the tangent bundle of the shape space as it is embedded in $\mathbb R^D$. Details are given in Appendix \ref{app:shape-space}.
It is on these gradients that we perform the normalization as described in Section \ref{sec:ouralg-computation}.
Remaining specifics of our MD data analytics pipeline are in Appendix \ref{app:dictionary-details}.


\subsection{Synthetic data results}\label{sec:syndata}

\paragraph {\ouralg~ on \srdata}
\label{sec:swiss}
We use the well known swiss roll dataset to demonstrate that \ouralg~is invariant to the choice of embedding algorithm.
Our \srdata~dataset consists of points sampled from a two dimensional rectangle and rolled up along one of the two axes, then randomly rotated in $D=49$ dimensions. 
We learn the manifold using three techniques: Local Tangent Space Alignment, Diffusion Maps, and Isomap, shown in Figures \ref{fig:sr_ltsa}, \ref{fig:sr_isomap} and \ref{fig:swissroll_spectral}.
For comparison, we also analyze the ``trivial embedding'' consisting of coordinates given by projection onto the rectangle edges (Figure \ref{fig:swissroll_internalembed}). 
These rectilinear coordinates are colored in red and blue, and show clear associations with individual embedding coordinates.

The dictionary $\G$ consists of $g_{1,2}$, the two rectilinear coordinates, as well as $g_{j+2}=\xi_j$, for $j=1,\ldots 49$, the coordinates of the feature space. 
Applying \ouralg~to the embeddings identifies the set $S=\{g_1,g_2\}$ as the manifold explanation, and identifies the association of the recovered support with individual embedding coordinates $\phi_{1,2}$.
By visual inspection of Figures \ref{fig:swissroll_internalembed}, \ref{fig:sr_ltsa}, \ref{fig:sr_isomap}, and \ref{fig:swissroll_spectral}, we see that all embedding algorithms recover the original manifold, although the embeddings $\phi^{\rm Iso},\phi^{DM},\ldots$ are not isometric (this is more noticeable with Diffusion Maps), and sign changes are possible. However, Figures \ref{fig:sr_internal_mflasso}, \ref{fig:sr_ltsa_mflasso}, \ref{fig:sr_isomap_mflasso} and \ref{fig:sr_spectral_mflasso} demonstrate that \ouralg~ recovers the two manifold-specific coordinate functions in each case, while the coefficients $\beta_{3:51}$ decay rapidly to 0 with $\lambda$. Furthermore, each of $g_{1,2}$ is always mapped to the correct embedding coordinate. The regularization paths are virtually identical for all embeddings, even though the embeddings are not isometric.
\comment{Note the differential orientation of the internal coordinates in Figures  \ref{fig:sr_internal_mflasso} and \ref{fig:swissroll_internalembed} with the other embeddings.}
\begin{figure}[H]
\subfloat[]{\includegraphics[width=4cm, trim={-.1cm, -.5cm,
-.5cm, -.5cm}, clip]{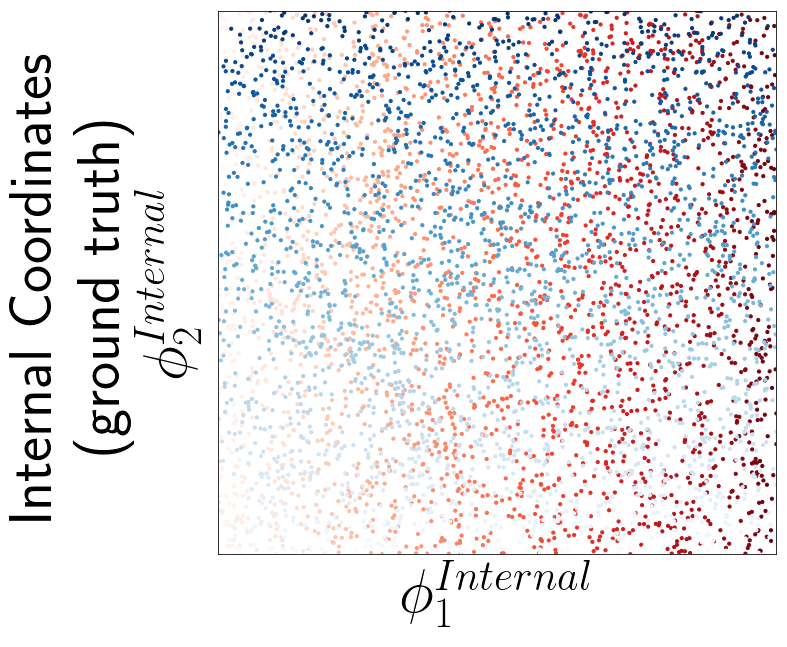}\label{fig:swissroll_internalembed}}\hfill
\subfloat[]{\includegraphics[width=10cm,  trim={-1cm, -1cm,
-1cm, -1cm}, clip]{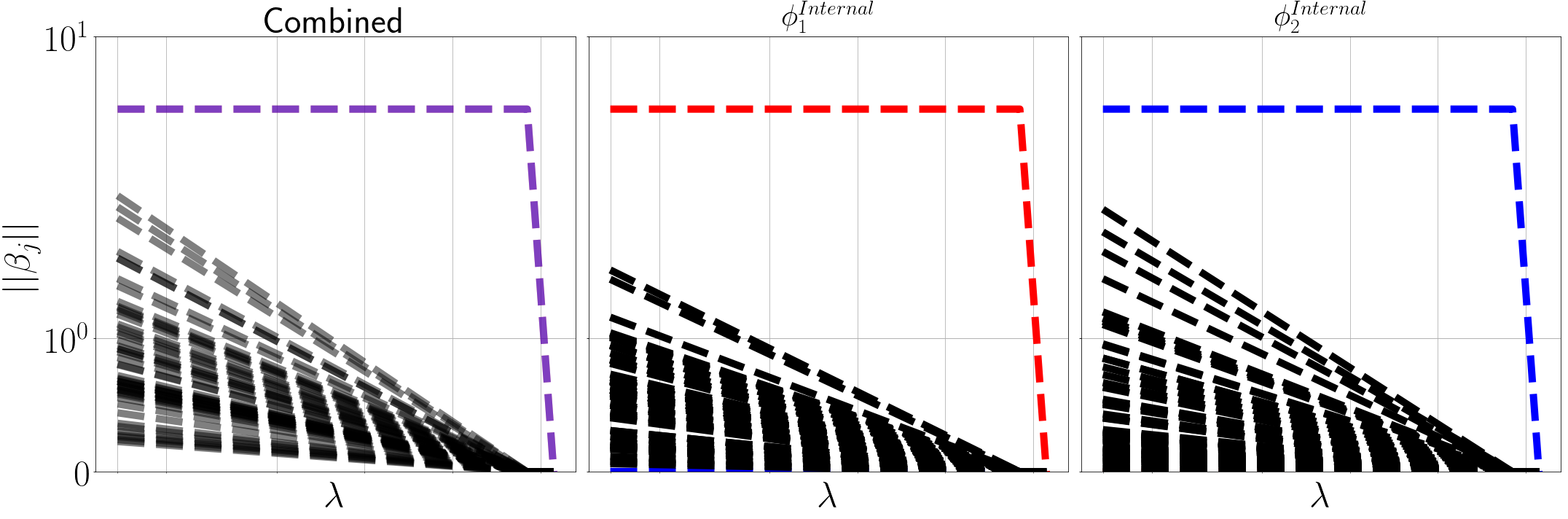}\label{fig:sr_internal_mflasso}}\hfill
\newline
\subfloat[]{\includegraphics[width=4cm, trim={-.1cm, -.5cm,
-.5cm, -.5cm}, clip]{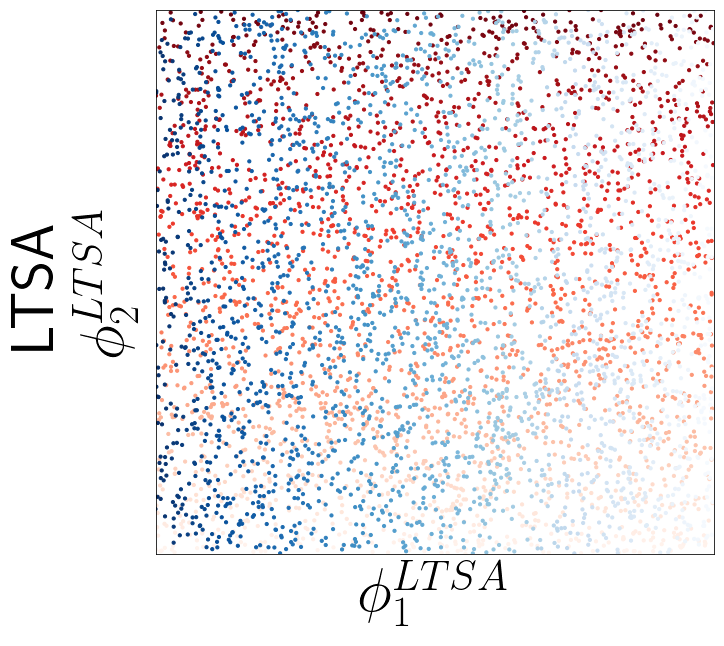}\label{fig:sr_ltsa}}\hfill
\subfloat[]{\includegraphics[width=10cm,  trim={-1cm, -1cm,
-1cm, -1cm}, clip]{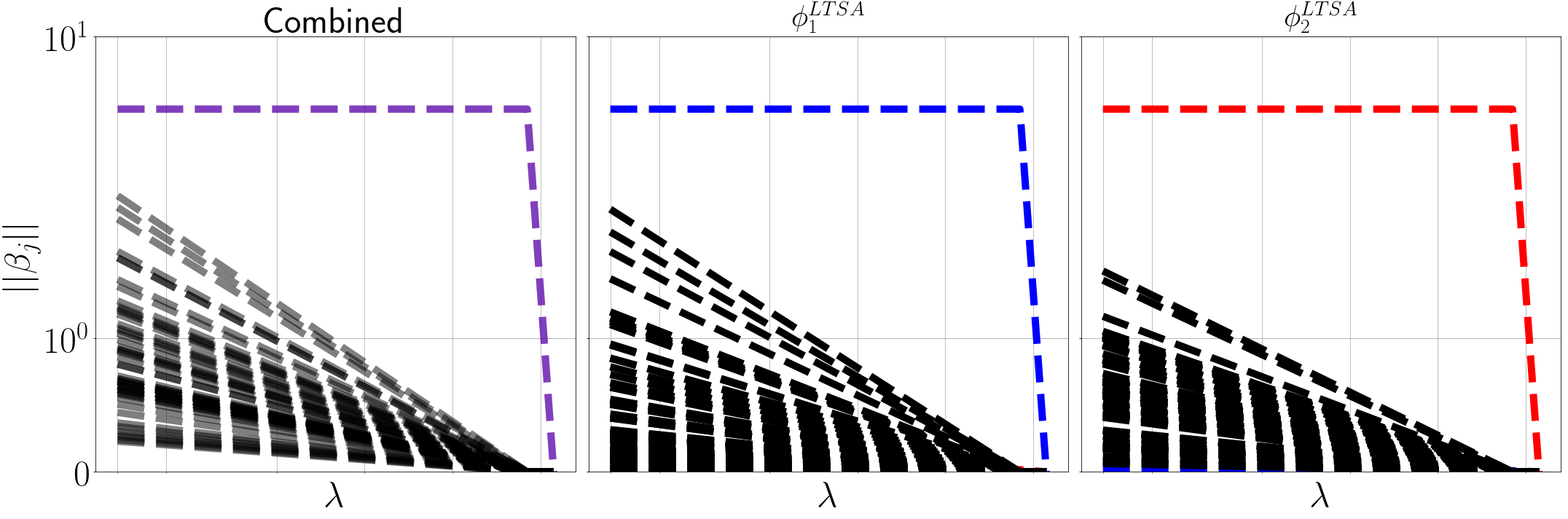}\label{fig:sr_ltsa_mflasso}}\hfill
\newline
\subfloat[]{\includegraphics[width=4cm, trim={-.1cm, -.5cm,
-.5cm, -.5cm}, clip]{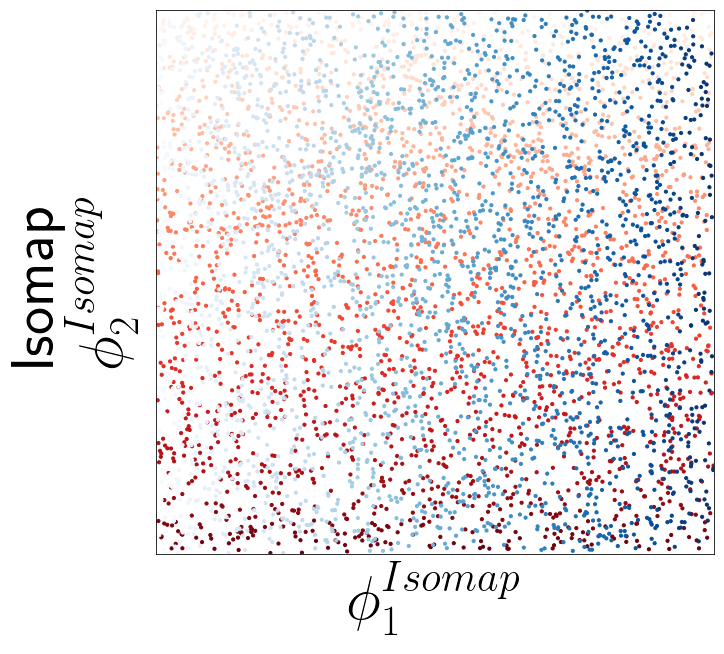}\label{fig:sr_isomap}}\hfill
\subfloat[]{\includegraphics[width=10cm,  trim={-1cm, -1cm,
-1cm, -1cm}, clip]{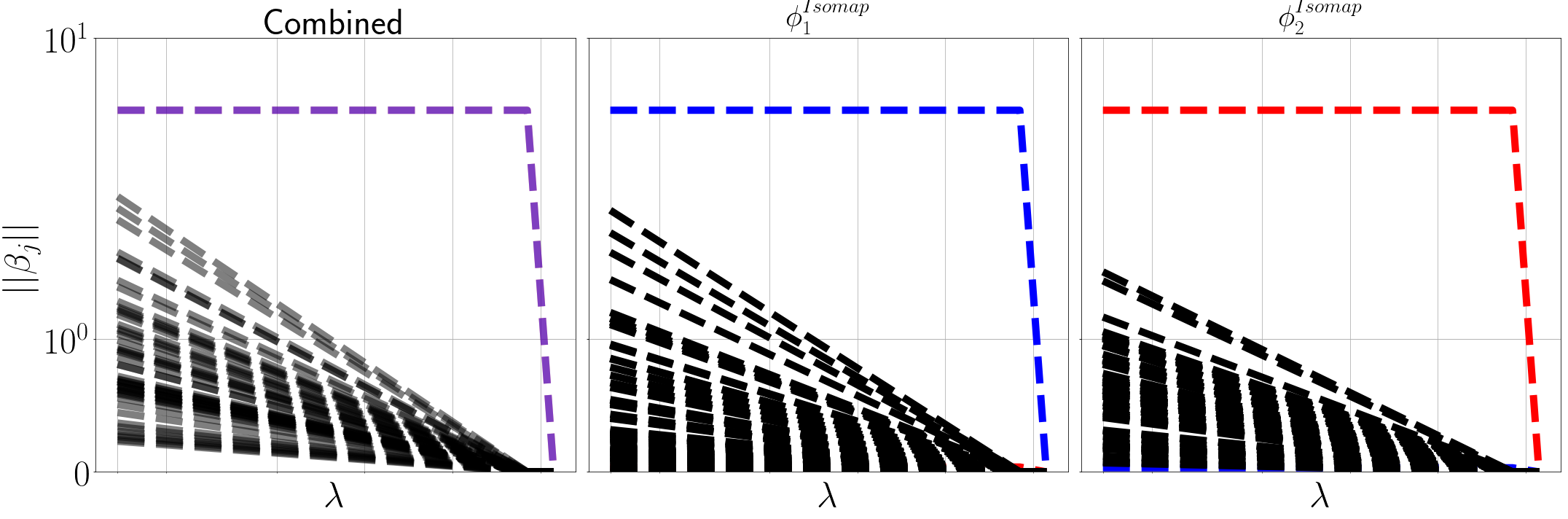}\label{fig:sr_isomap_mflasso}}\hfill
\newline
\subfloat[]{\includegraphics[width=4cm, trim={-.1cm, -.5cm,
-.5cm, -.5cm}, clip]{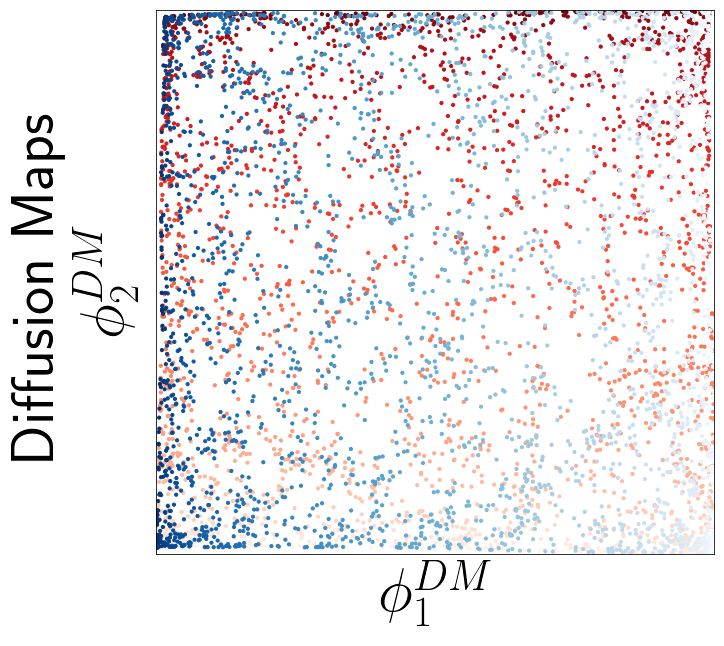}\label{fig:swissroll_spectral}}\hfill
\subfloat[]{\includegraphics[width=10cm,  trim={-1cm, -1cm,
-1cm, -1cm}, clip]{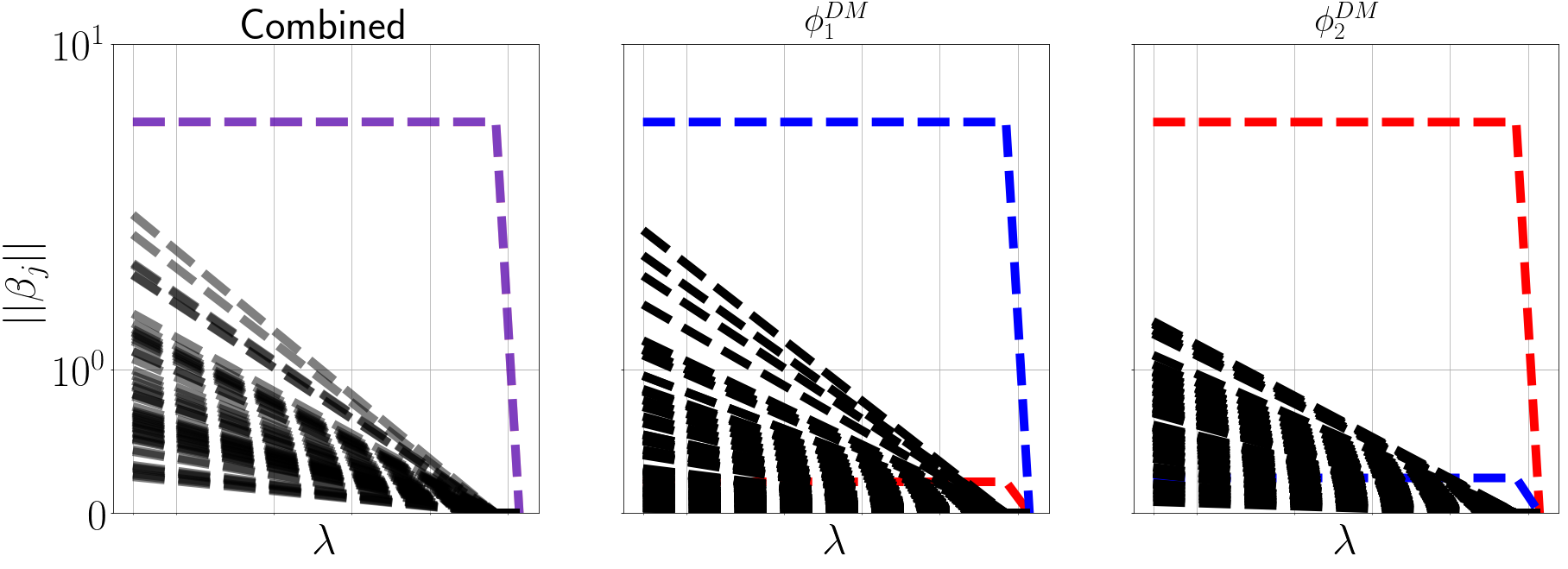}\label{fig:sr_spectral_mflasso}}\hfill
\caption{Results for \srdata~ embedded using a variety of manifold learning algorithms.
Figure \ref{fig:swissroll_internalembed} shows the data mapped w.r.t. the edges of the rectangle.
Figures \ref{fig:sr_ltsa}, \ref{fig:sr_isomap}, and \ref{fig:swissroll_spectral} display embeddings of \srdata~ generated by several different manifold learning methods, colored by the rectilinear coordinates in red and blue. 
Figures \ref{fig:sr_internal_mflasso}, \ref{fig:sr_ltsa_mflasso}, \ref{fig:sr_isomap_mflasso}, and \ref{fig:sr_spectral_mflasso} display the regularization paths of \ouralg~ for these embeddings.
The combined norms $ \|\beta_j\|$ used in \ouralg~ are given on the left, and the norms for the individual embedding coordinates $ \|\beta_{jk}\|$ on the right.
}
\label{fig:sw}
\end{figure}

\paragraph{\ouralg~on a Rigid Ethanol skeleton}\label{sec:rigid}

As a prelude to real MD data, we demonstrate the workings of
\ouralg~in a controlled setting by applying it to a simple
non-dynamical simulation of a rigidly-rotating ethanol molecule. We
first construct an ethanol skeleton composed of the atoms shown in
Figure \ref{fig:reth-diagram}. We then sample as we rotate the atoms
around the C-C and C-O bonds.
In contrast with the MD trajectories, which
are simulated according to quantum dynamics, these two angles are
distributed uniformly over a grid, and Gaussian noise is added to the
position of each atom.  We call the resultant dataset \redata. 
As expected given our two a priori known degrees of freedom, Figures \ref{fig:reth-g1} and
\ref{fig:reth-g2} show that the estimated manifold is a two-dimensional
surface with a torus topology parameterized by bond
torsions $g_1$ and $g_2$ similar to that observed for the MD
\ethdata~ in Figure \ref{fig:molecs}.

The dictionary consists of the twelve torsions implicitly defined by the bond diagram\footnote{These are all 4-tuples of atoms connected by a path in the figure, modulo the natural equivalence relation on torsions previously described.}
 in Figure \ref{fig:reth-diagram}. All of these torsions circumscribe one of the central C-C and C-O bonds. Counting permutations of peripheral hydrogens, we can see that there are $9$ of the former, and $3$ of the latter, which we denote $g_{0:8}$ and $g_{9:11}$ in Figure \ref{fig:re12_cosine}. Hence, any pair $\{g_j,g_{j'}\}$ with $j\in \{0:8\},j'\in\{9:11\}$ is an equally correct coordinate system for this manifold. 
This is shown in Figure \ref{fig:re12_cosine} by the incoherences $\mu_{jj'}$, i.e. mean pairwise cosines of the dictionary functions.
Comparing the row and column labels of Figure \ref{fig:re12_cosine} with Figure \ref{fig:reth-diagram} shows that the collinearities of these gradients clearly cluster by central bond.
Thus, we expect \ouralg~ to recover one torsion from each group.
Indeed, in the regularization path of an individual replicate of \ouralg~  shown in Figure \ref{fig:re12_replicate_noise}, collinear torsions are killed off, and a representative torsion is selected from each group. 
Finally, Figure \ref{fig:re12_support} shows that \ouralg~selects such orthogonal pairs in 18 out of $ 25$ random replicates of the $n'$ points.

\begin{figure}[H]
\subfloat[]{\includegraphics[width=5cm]{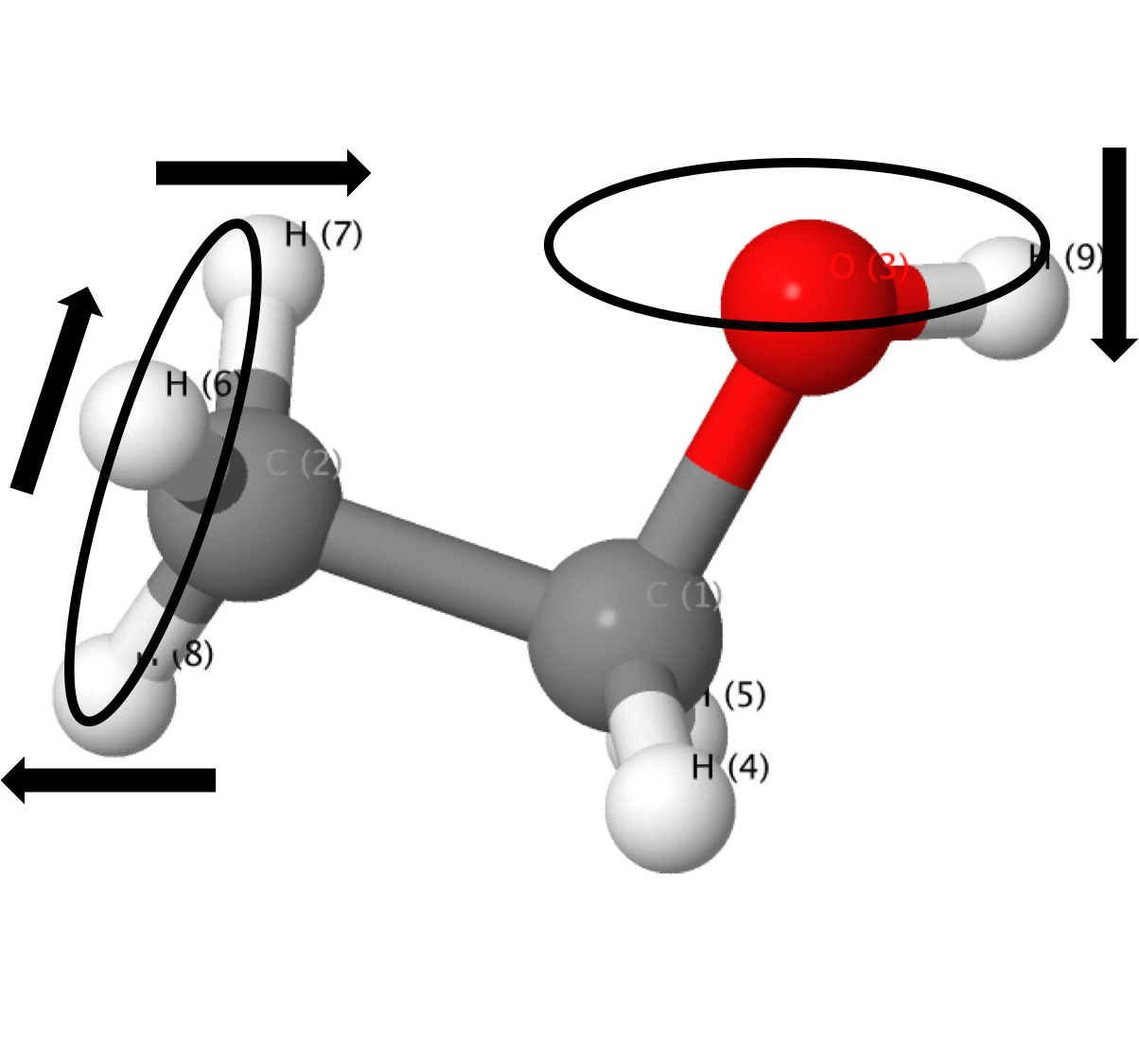}\label{fig:reth-diagram}}\hfill
\subfloat[]{\includegraphics[width=5cm]{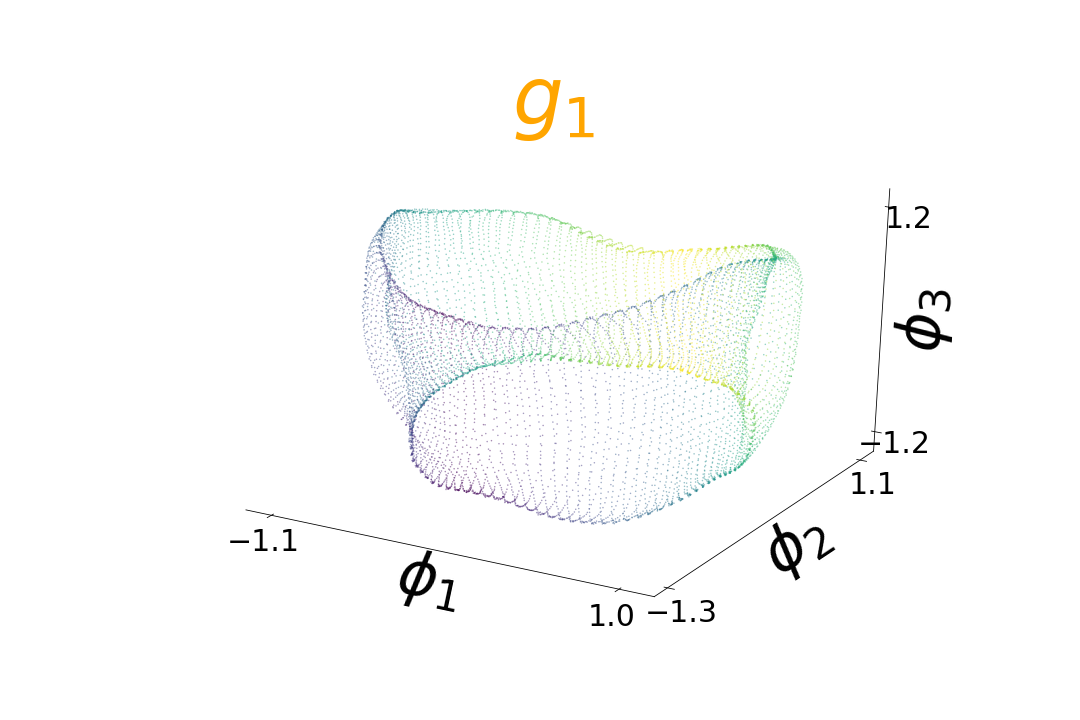}\label{fig:reth-g1}}\hfill
\subfloat[]{\includegraphics[width=5cm]{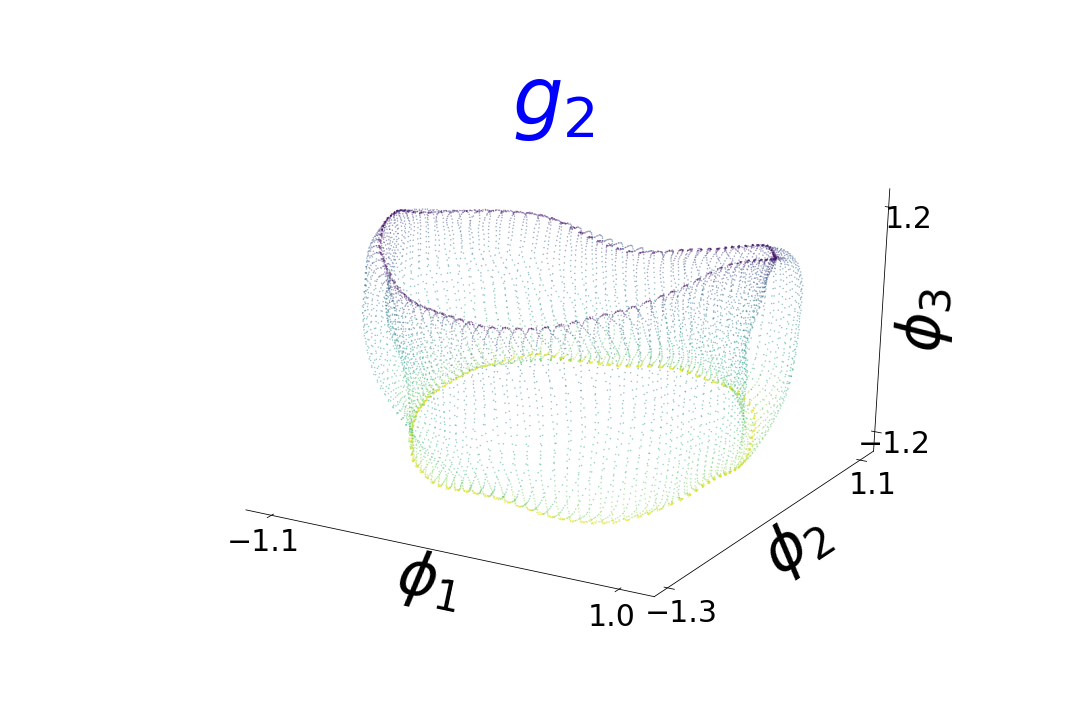}\label{fig:reth-g2}}\hfill 
\newline
\subfloat[]{\includegraphics[width=5cm]{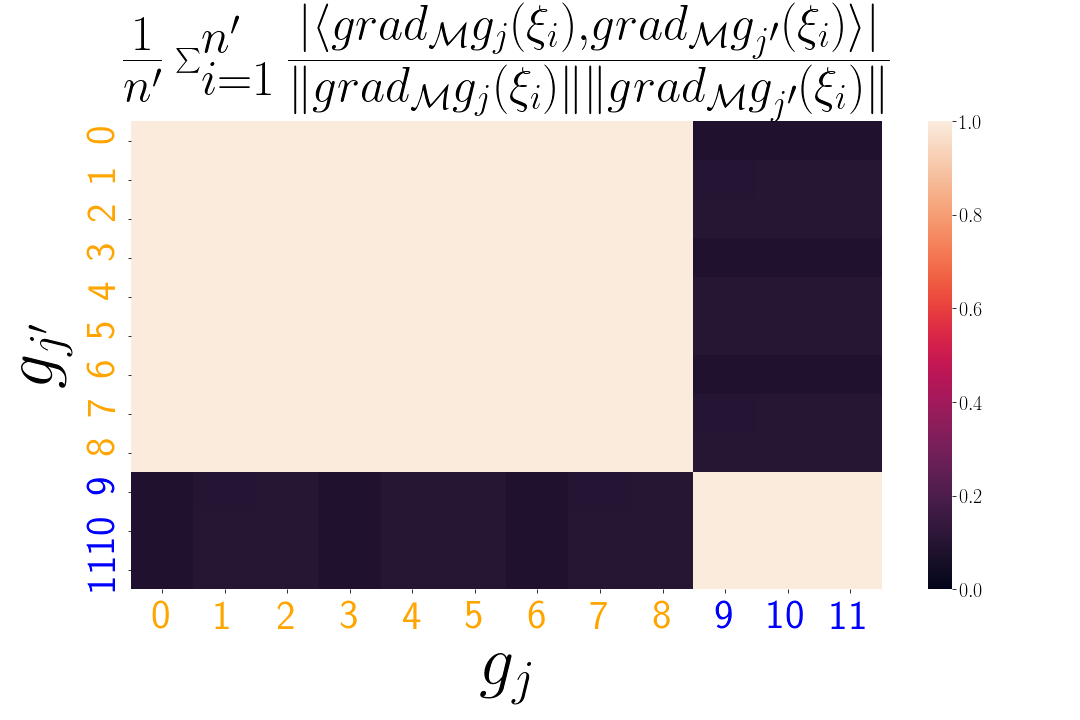}\label{fig:re12_cosine}}\hfill
\subfloat[]{\includegraphics[width=5cm]{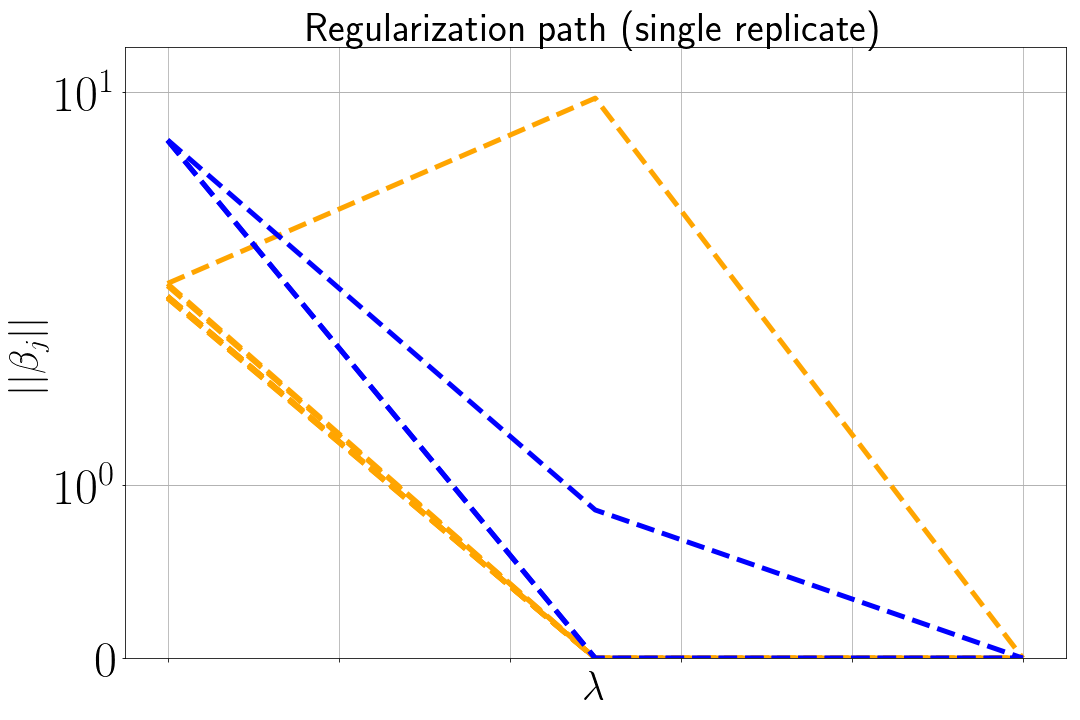}\label{fig:re12_replicate_noise}}\hfill
\subfloat[]{\includegraphics[width=5cm]{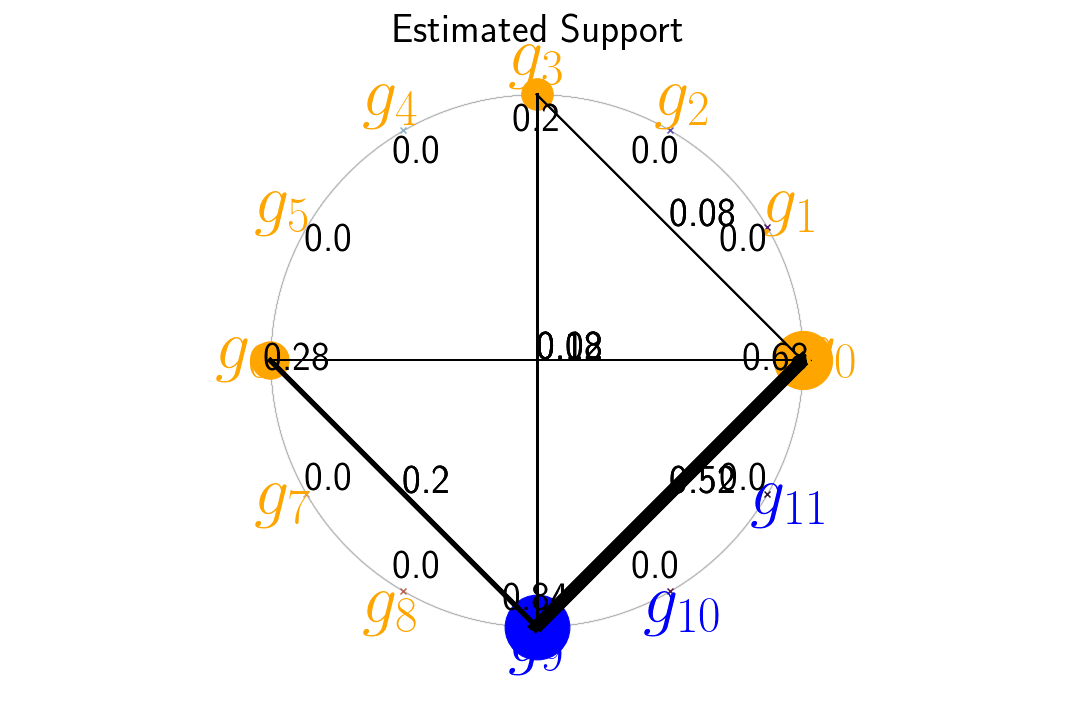}\label{fig:re12_support}}\hfill
\caption{Results of \ouralg~for \redata.
 Figure \ref{fig:reth-diagram} shows the simplified dynamics of our rigid molecular simulation.
 Atoms in the rigid ethanol skeleton are articulated around the C-O and C-C bonds by a torus of rotations.
Figure \ref{fig:reth-g1} shows the learned torus, colored by C-C torsion $g_1$ from Figure \ref{fig:molecs}.
Figure \ref{fig:reth-g2} shows the same torus, colored by the C-O torsion $g_2$ from Figure \ref{fig:molecs}.
Figure \ref{fig:re12_cosine} displays the incoherences, i.e. pairwise collinearities of dictionary gradients; C-C torsions are in orange, C-O torsions in blue.
Figure \ref{fig:re12_replicate_noise} shows regularization paths $\|\beta_j\|$ vs. $\lambda$ for a single replicate.
The chord diagram in Figure \ref{fig:re12_support} represents the frequency of selecting each pair of torsions in replicate experiments.
The listed frequencies with which individual torsions are selected are given by the sizes of the perimeter dots corresponding to each dictionary element, while the frequencies with which pairs of torsions are selected are given by the line widths connecting the dots.  Frequencies are also given by the numbers next to the respective graphical indicators.
}
\label{fig:rigid-ethanol}
\end{figure}

\subsection{Molecular Dynamics results}\label{sec:mds_data}

In the same manner, we use \ouralg~ to identify torsions that govern
the dynamics of the molecules in Figure \ref{fig:molecs}.  From the
machine learning point of view, MD data from well-studied molecules
are an excellent testbed: the manifold hypothesis is believed to hold
approximately, there is sufficient data to learn a manifold, and the
ground truth is available and can validate our algorithms. Moreover,
MD data are challenging problems for manifold learning.  Appendix
\ref{app:feature-space} displays \toldata, \maldata, and \ethdata~in
the $\xi$ representation, showing high amplitude noise outside the
manifold; indeed, MD data has multiscale structure and the ``noise''
is non-uniformly distributed and highly correlated in the $\rrr^D$
space.  Since the gradients of the dictionary functions are calculated
analytically from the $\xi$ coordinates, and the data does not lie
exactly on $\M$, the values of $\grad_{T_i^\M} g_j$ will necessarily be
noisy as well.  

From the scientific point of view, high quality MD data
are highly expensive to generate, taking weeks or months of
supercomputer time \citep{Bowers2006-sa,Fiorin2013-fl}.  Fast
automated analysis of these data by identification of so called {\em collective
variables} serves both in the scientific understanding of the data and in acceleration sampling methods \citep{rohrdanzZhengClementi:mountain-passes13}.
Moreover, every new simulation represents a new manifold, and a new
manifold explanation problem.

We first show that \ouralg~ can distinguish groups that correspond to the chemical bonds in Figure \ref{fig:molecs}, as would typically be done by a scientist using prior domain knowledge. 
Next, we repeat the analysis with no prior knowledge, including all distinct 4-tuples of atoms in the dictionary.

\paragraph{Dictionaries based on bond diagrams}

Bond diagrams such as the ones in Figure \ref{fig:molecs} are based on a priori information about molecular structure garnered from historical work.
Building a dictionary based on this structure is akin to many other methods in the field \citep{Krenn2020-aj, Xie2019-kw}. 
As in the case of \redata, our dictionaries consist of all equivalence classes of 4-tuples of atoms implicitly defined by bond diagrams, and the incoherence plots for \ethdata~and \maldata~in Figures \ref{fig:e12_cosines} and \ref{fig:mal12_cosines} show two groups of highly dependent torsions, corresponding to the two bonds between heavy atoms in the molecules.
Therefore, success means recovering a pair of incoherent torsions out of these dictionaries.
For \toldata, the manifold dimension is $d=1$ and success means recovering one of the 6 torsions associated with the peripheral methyl group bond.
For this molecule, there are also $p-6=24$ torsions that to not explain the data manifold.
We apply \ouralg~with these dictionaries to the embeddings shown in Figure \ref{fig:molecs}.

As Figure \ref{fig:apriori-results} shows, \ouralg~ is always able to identify  torsions corresponding to the expected labelled bonds.
Figures \ref{fig:e12_replicate},  \ref{fig:mal12_replicate}, and  \ref{fig:tol30_replicate}  show regularization paths for single replicates of \ouralg, and  Figures \ref{fig:e12_support},  \ref{fig:mal12_support} and  \ref{fig:tol30_support}  show frequencies of support recovery of sets of size $d$ over $w=25$ replicates.
 \ouralg~finds that the toroidal \ethdata~manifold is explained by pairs of torsions from the C-O and C-C bonds, while \maldata~is explained by one of each of the two central bonds. \toldata~is explained by the torsion of the peripheral methyl group.
These agree with our domain-expert validated parameterizations from Figure \ref{fig:molecs}.
Torsion association with individual embedding coordinates is examined in Appendix \ref{app:sup_exp}

For all quantum MD experiments, we examine the support recovery condition Theorem
\ref{thm:S}. We first note that Figures \ref{fig:e12_cosines} and
\ref{fig:mal12_cosines} show that even without foreknowledge of a
unique true support, the incoherence parameter $\mu$ must be quite close to $1$, since it is a maximum over set of cosines whose mean is plotted. The empirical distributions of the parameters of this
Theorem across replicates are listed in Appendix \ref{app:sup_exp}.
The high values of the incoherence parameter $\mu$ and otherwise
unfavorable empirical support recovery parameters listed in the table
indicate that we cannot expect a unique recovery.
However, \ouralg~is still successful in obtaining representative
torsions from the desired bonds in Figure \ref{fig:molecs}. The similarity between the results on this real data, with more challenging noise and variable sampling density, and the result on the synthetic \redata~are witness to the robustness of the \ouralg~method.

\comment{We posit
that the  ``success'' of \ouralg~ is due to
the small amount of between-cluster collinearity, and the fact that we
are satisfied with any member of each function cluster.}
\begin{figure}[htb]
\subfloat[]{\includegraphics[width=5cm, trim={-.5cm, -.5cm,
-.5cm, -.5cm}, clip]{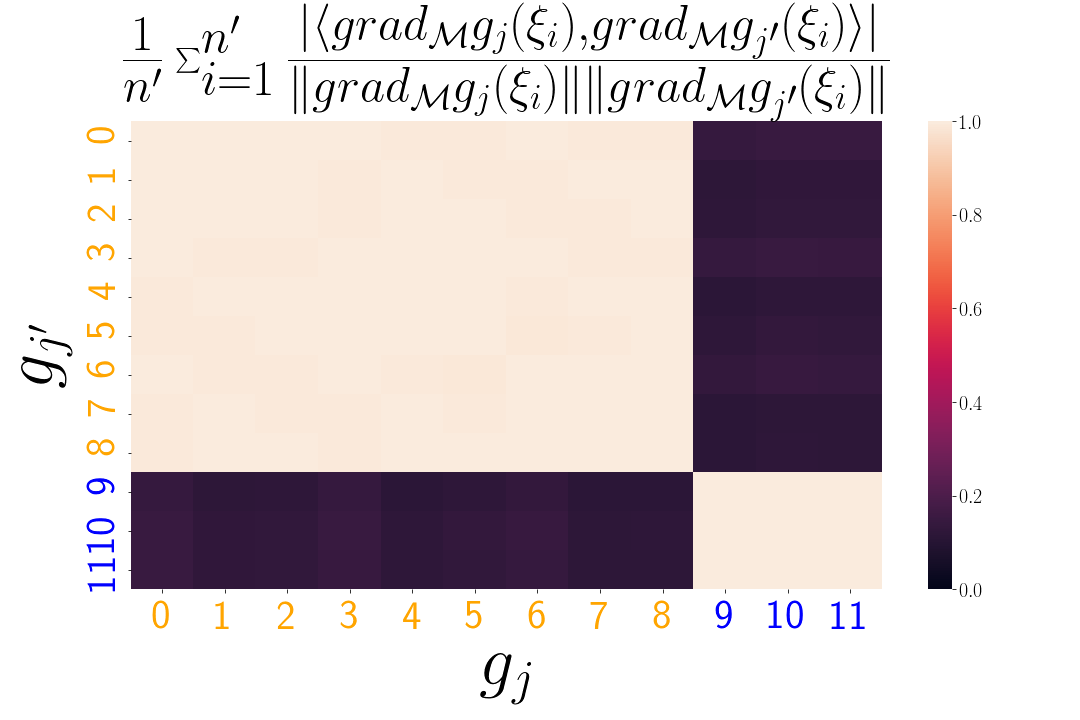}\label{fig:e12_cosines}}\hfill
\subfloat[]{\includegraphics[width=5cm,  trim={-1cm, -1cm,
-1cm, -1cm}, clip]{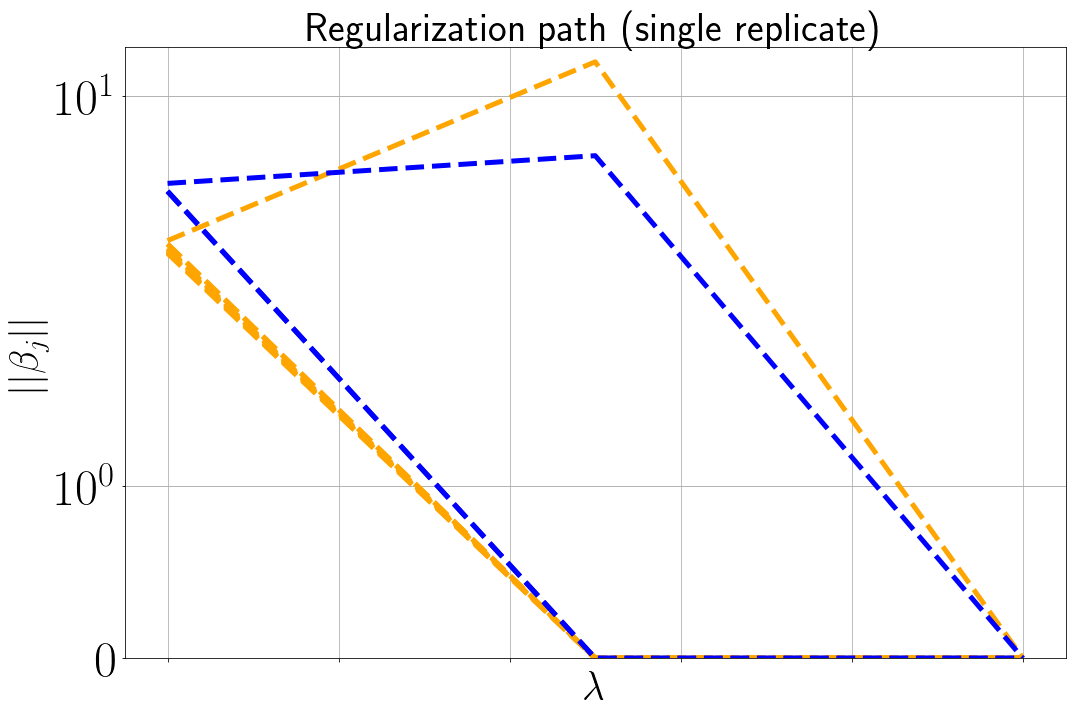}\label{fig:e12_replicate}}\hfill
\subfloat[]{\includegraphics[width=5cm, trim={0, -3cm, 0, 0},
clip]{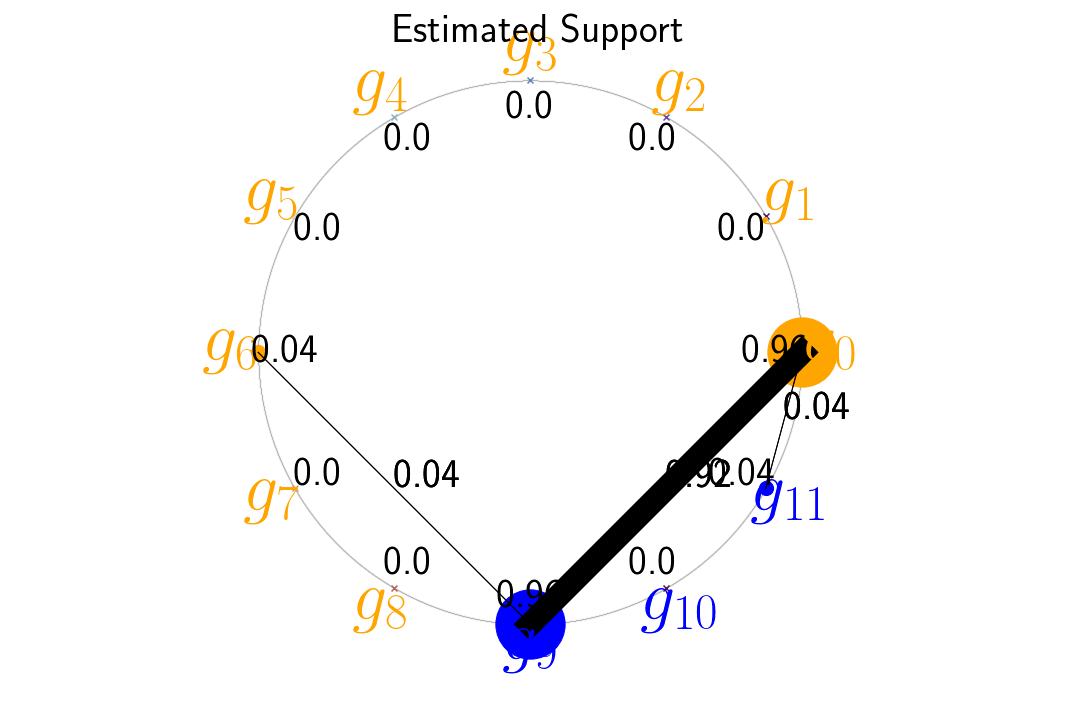}\label{fig:e12_support}}\hfill 
\newline
\subfloat[]{\includegraphics[width=5cm,  trim={-1cm, -1cm,
-1cm, -1cm}, clip]{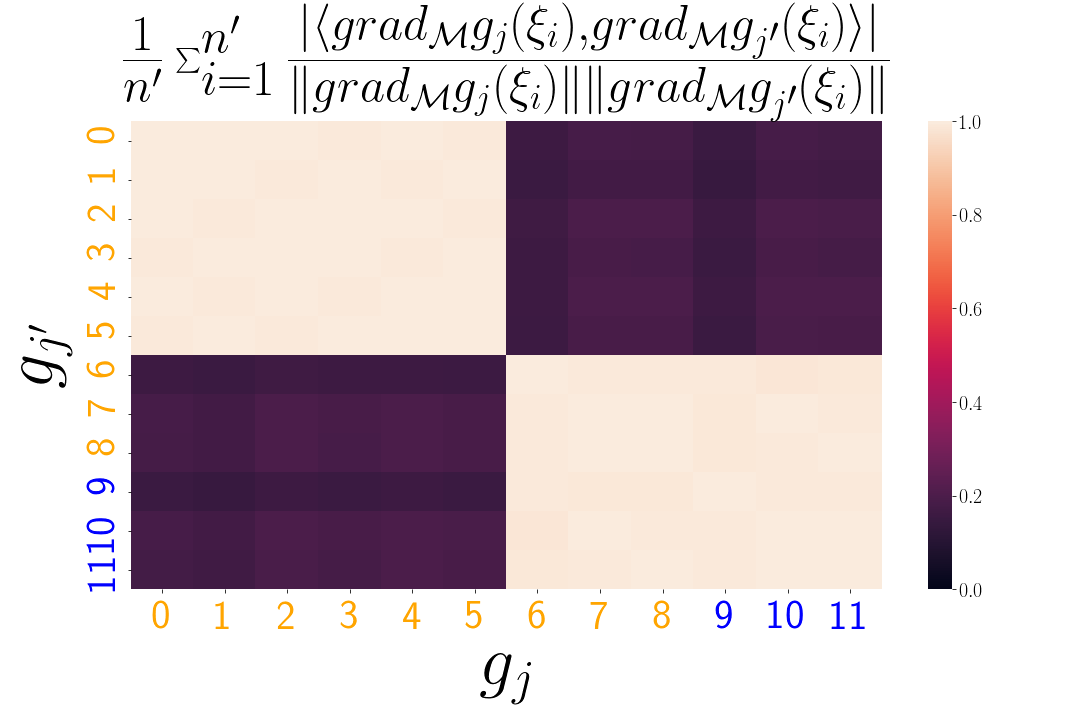}\label{fig:mal12_cosines}}\hfill
\subfloat[]{\includegraphics[width=5cm,  trim={-1cm, -1cm,
-1cm, -1cm}, clip]{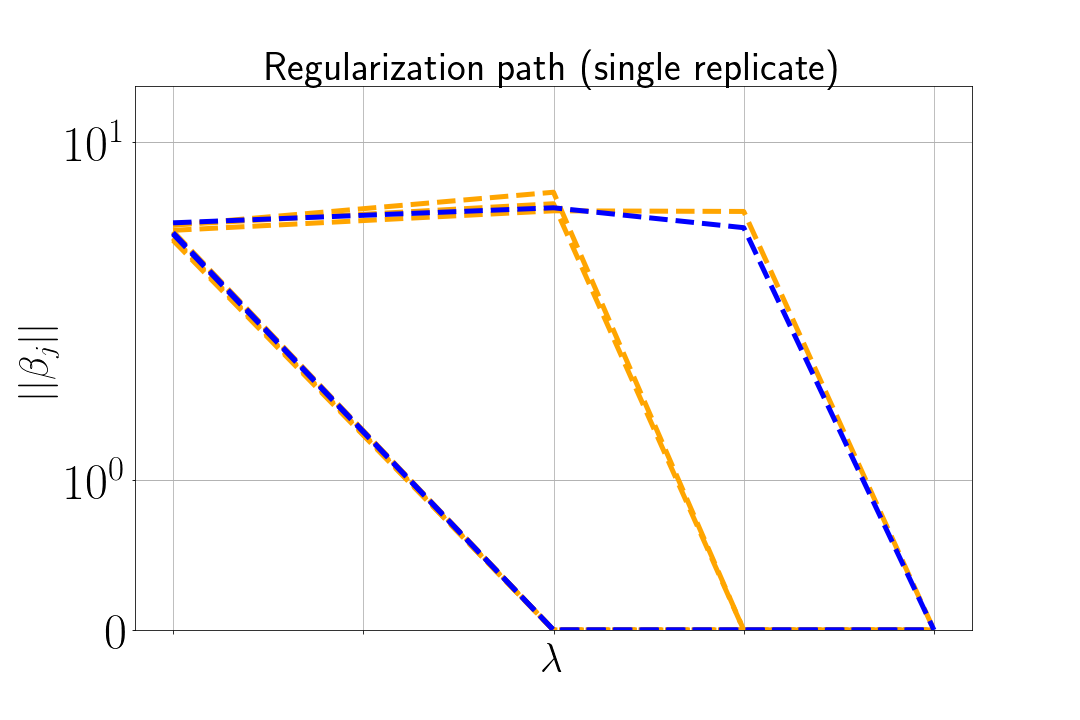}\label{fig:mal12_replicate}}\hfill
\subfloat[]{\includegraphics[width=5cm,  trim={-1cm, -1cm,
-1cm, -1cm}, clip]{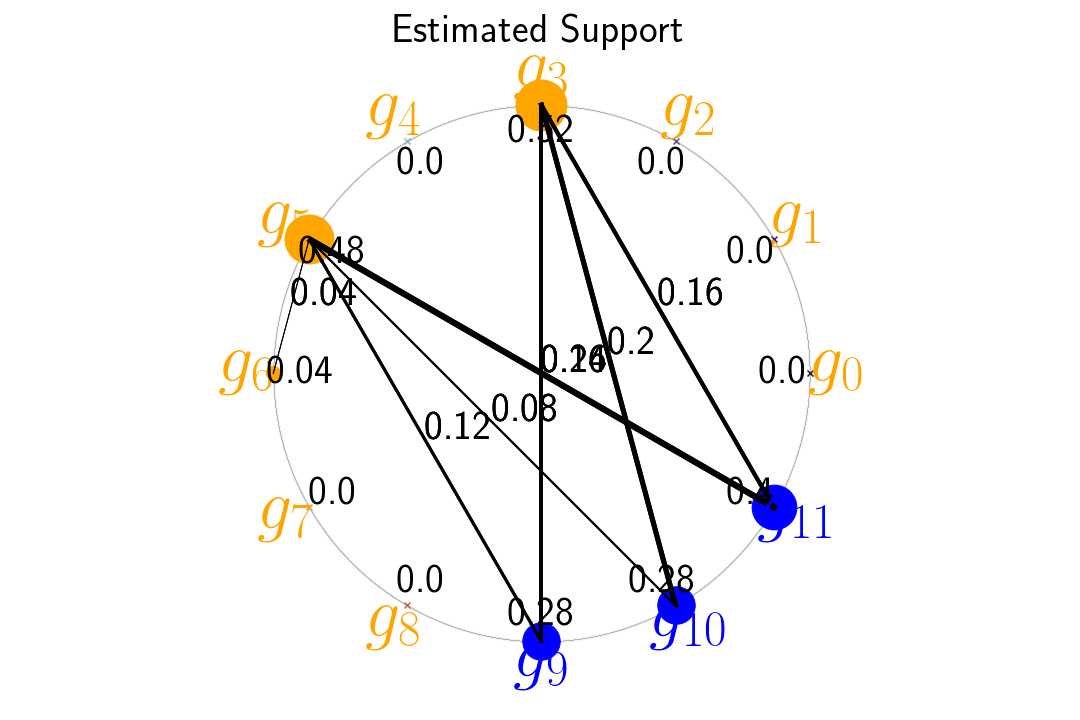}\label{fig:mal12_support}}\hfill
\newline
 \hspace*{.33\textwidth}\quad
\subfloat[]{\includegraphics[width=5cm,  trim={-1cm, -1cm,
-1cm, -1cm}, clip]{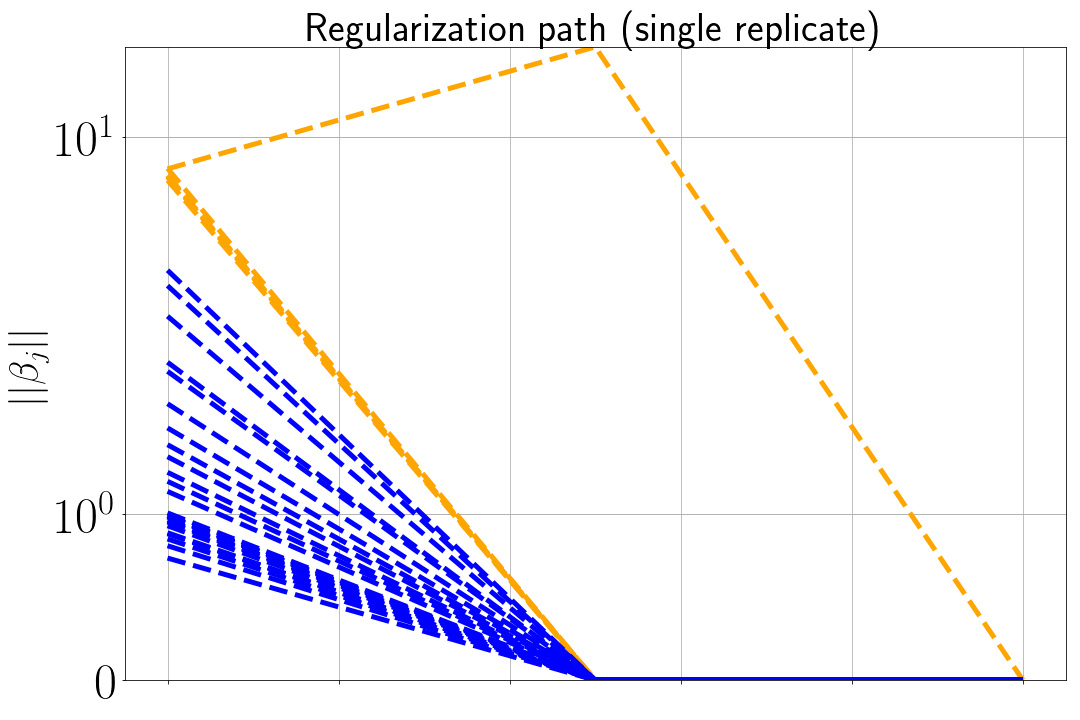}\label{fig:tol30_replicate}}\hfill
\subfloat[]{\includegraphics[width=5cm,  trim={-1cm, -1cm,
-1cm, -1cm}, clip]{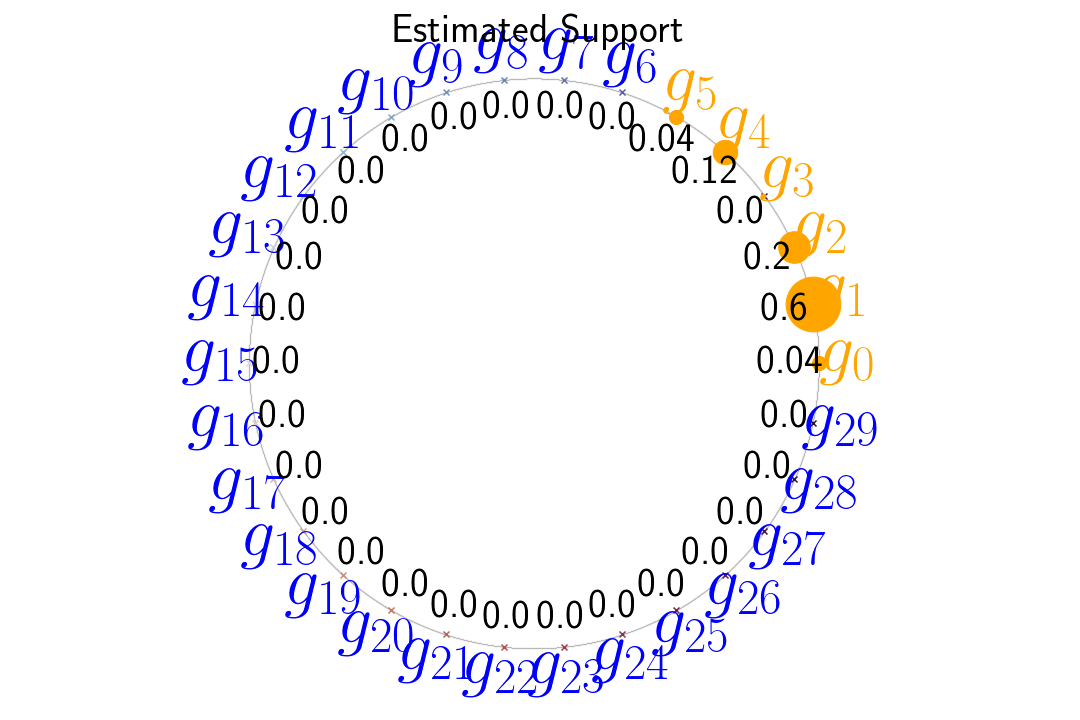}\label{fig:tol30_support}}\hfill
\caption{Results for MD data with a priori dictionaries given by the bond diagrams in Figure \ref{fig:molecs}. 
The three rows correspond to \ethdata, \maldata, and \toldata, respectively.
Figures \ref{fig:e12_cosines} and \ref{fig:mal12_cosines} display pairwise collinearities of dictionary gradients, colored by bond as in Figure \ref{fig:molecs}.
Toluene, a $1-d$ manifold, has trivial cosines, and so these are not shown. 
Figures \ref{fig:e12_replicate}, \ref{fig:mal12_replicate}, and \ref{fig:tol30_replicate} show overall regularization paths of $\|\beta_j\|$ for single replicates.
Figures \ref{fig:e12_support}, \ref{fig:mal12_support}, and \ref{fig:tol30_support} show chord diagrams displaying frequency of support recovery of sets of size $d$ for $25$ replicates.
As for \redata, two-dimensional support recovery frequency is denoted by chord width, and one-dimensional support recovery frequency is denoted by size of perimeter dot.
Note that 'blue' in toluene corresponds to torsions in the benzene ring.
}
\label{fig:apriori-results}
\end{figure}

\paragraph{Results from full dictionary}
\comment{Actual interatomic interactions are often more complex than exhibited in a bond diagram; thus, the analyzed molecules have more interactions than those represented in Figures and \ref{fig:toluene-bonds}-\ref{fig:mda-bonds}, and potentially more interesting bond torsions.
Indeed, a motivating application of quantum dynamics simulations is to uncover molecular behavior that is not encapsulated by the simplified bond diagrams.
This motivates us }
Now we test \ouralg~in the extreme case when the dictionary consists off all possible torsions, i.e. all $\binom{N_a}{4}$ 4-tuples modulo equivalence.
For \ethdata~and \maldata~we obtain $p=756$ torsions\footnote{We do not analyze \toldata, because for $d=1$ the solution is available analytically, making this example somewhat trivial.}. 
Such a large $p$ is challenging for $l_1$ regularized estimation, due to the bias mentioned Section \ref{sec:variants} for large $\lambda$. 
Moreover, examining Figures \ref{fig:eth-fullcos} and \ref{fig:mal-fullcos}, we see that, besides the two groups of collinear torsions in the previous dictionary, there are other torsions, about a fourth of the 756, that are coherent with both groups. 
While we do not necessarily expect \ouralg~to succeed, or to be used in such a way in practice, this experiment will inform us on the robustness of \ouralg~in a situation that is challenging for any type of sparsity inducing regularization.

\comment{The large number of functions and complex pattern of coherences in the dictionaries presents a difficult challenge for our method, and is confirmed by the empirical parameters in Appendix \ref{app:sup_exp}. 
besides the increase dictionary size, there are dictionary functions coherent with both ground truth torsions.}

The results of \ouralg~with the full dictionary for \ethdata~ and \maldata~ are displayed in Figure \ref{fig:agnostic-results}. For consistency between replications, we choose {\em a priori} the ground truth to be represented by torsions $g_{74,176}$ and $g_{0,8}$, which are representative torsions for \ethdata, respectively for \maldata, depicted in Figure \ref{fig:molecs}. We can evaluate the selected $d=2$ functions for coherence with this ground truth. 
In this most challenging setting, \ouralg~ identifies supports with mean incoherences with the true support of $.68 \pm .32$ and $.95 \pm .1$ for \ethdata~and for \maldata, respectively.
This is apparent from comparing selected torsions in Figures \ref{fig:eth-full-mflasso-support} and \ref{fig:mal-full-mflasso-support} with their collinearities in Figures \ref{fig:eth-full-mflasso-support-cos} and  \ref{fig:mal-full-mflasso-support-cos}. 
Thus, we can see that \ouralg~ performs preferably on \maldata.

In the latter figures, collinearities of the selected supports with example functions from the representative true support are also plotted.
We can see that the selected support functions are often strongly coherent with the ground truth functions, both when the selected support is almost orthogonal, and when the selected support functions are not. In the latter case, both selected support functions are strongly coherent with only one of the ground truth functions.
Note that when both selected functions are more coherent with a single element of the true support, we use the pairwise coherences with higher mean.
The results are visualized in  Appendix \ref{app:sup_exp}, which shows the embeddings colored by the selected torsions.
There is a clear visual correspondence between coherences between torsions and their colorings of the manifolds learned from \ethdata~ and \maldata; thus, when orthogonal pairs are selected, we capture information that would otherwise necessarily be obtained visually from the embeddings.
However, the \maldata~plots also demonstrate that even for this simple manifold, associating manifold coordinates to dictionary functions by visual inspection is delicate work. 
From a chemistry perspective, orthogonal recovered torsions generally flank pairs of hydrogens of which each is attached to one of the central atoms in the putatively true bonds.
Thus, it makes sense that these peripheral torsions could geometrically describe the same motion as the putative true support.


Examination of the selected regularization paths in Figures \ref{fig:eth_fulldict_replicate} and \ref{fig:mal_fulldict_replicate} shows that a small number of unselected functions persist quite far into the regularization path.
Thus, when \ouralg~fails to select orthogonal functions, for example due to the documented support recovery instability and bias at high values of $\lambda$ for Lasso methods in general \citep{Meinshausen2007-ey, Hesterberg2008-hm, Huan_Xu2012-iu}, it  it is natural to consider a two-stage variable selection procedure in which a secondary variable selection step is applied after initial pruning, as in \citet{Hesterberg2008-hm}.
To test this, we perform a two-stage sparse regression.
In the first stage we heuristically choose $\lambda=\lambda_{max}/2$, which eliminates most dictionary functions; in the second stage we perform an exhaustive search over the remaining dictionary to optimize \eqref{eq:bp}.
This approach is described in detail in Appendix \ref{app:gsbp}.
In all cases, this simple variation recovers an explanation with functions that are highly collinear with the ground truth.
In the plotted replicate, the number $p'$ of dictionary elements selected at $\lambda_{max}/2$ is about 10 for \ethdata~ and 4 for \maldata.
Here, collinearities with the true support are $.97 \pm .03$ for \ethdata~ and $.96 \pm .01$ for \maldata, and we avoid selecting pairs of functions that are collinear with the same element of the true support.
Visual inspection of the colored embeddings for \ethdata~ in Appendix \ref{app:sup_exp} also confirms that these conform to our visual intuition of orthogonally varying torsions, and as for \maldata, these hydrogen-hydrogen torsions tend to abut the true central bonds. 
Together, these experiments show that on noisy, large $p$ problems, and with massive violations of the incoherence conditions,  \ouralg, while sometimes not successful on its own, can robustly prune the dictionary.

\comment{ but the selected support has larger cardinality than $d$. In aa second stage, one selects the best $d$ functions from this smaller set by brute force.
We do not formally define $\lambda_{-1}$ - it is simply the previous coefficient in the lambda search path - but we note that in prediction, utilization of a somewhat arbitrary value of $\lambda$ that is lower than would be required for support recovery is standard, and already makes Equation \eqref{eq:flasso-manifold} well-defined.
Examination of the collinearities of the functions retained at $\lambda_{-1}$ in Figures \ref{fig:eth-full-lm1-cos}, and \ref{fig:mal-full-lm1-cos} shows that, among the surviving functions, orthogonal pairs persist.
This section also contains a more detailed description of the shrinkage problem.
}


\newpage
\begin{figure}[htb]
\subfloat[]{\includegraphics[width=3cm]{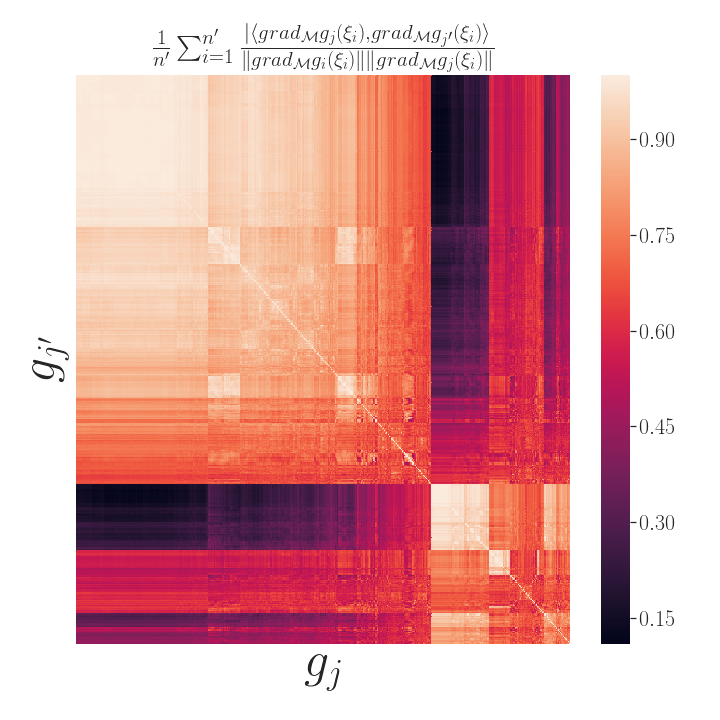}\label{fig:eth-fullcos}}\hfill
\subfloat[]{\includegraphics[width=3cm]{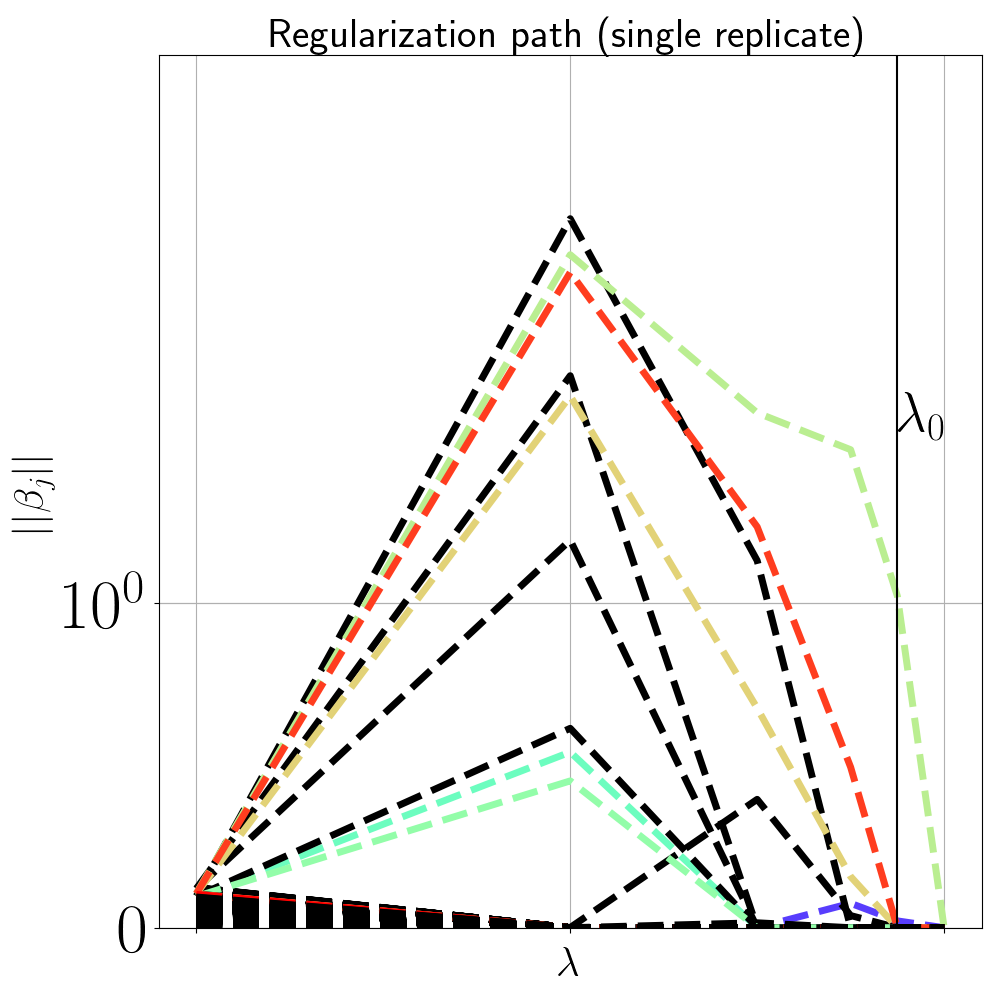}\label{fig:eth_fulldict_replicate}} \hfill
\subfloat[]{\includegraphics[width=3cm]{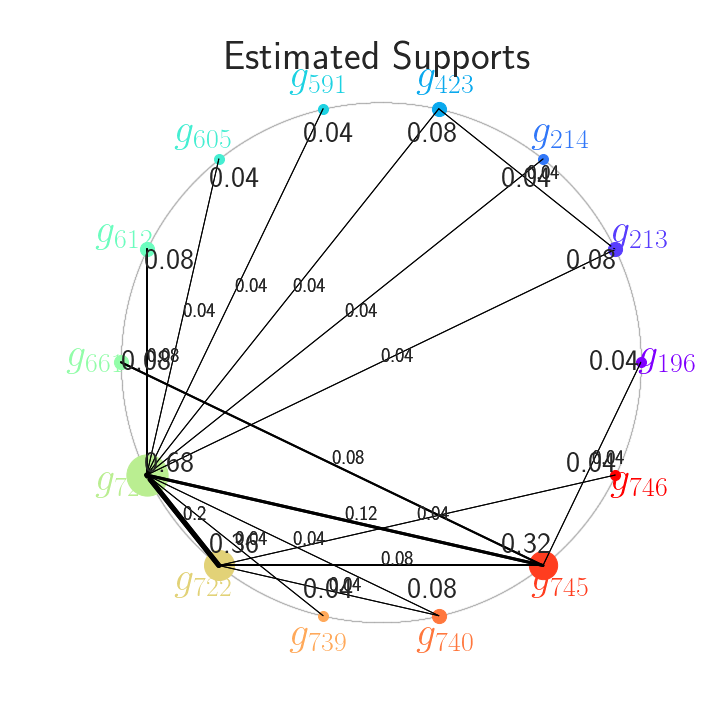}\label{fig:eth-full-mflasso-support}}\hfill
\subfloat[]{\includegraphics[width=3cm]{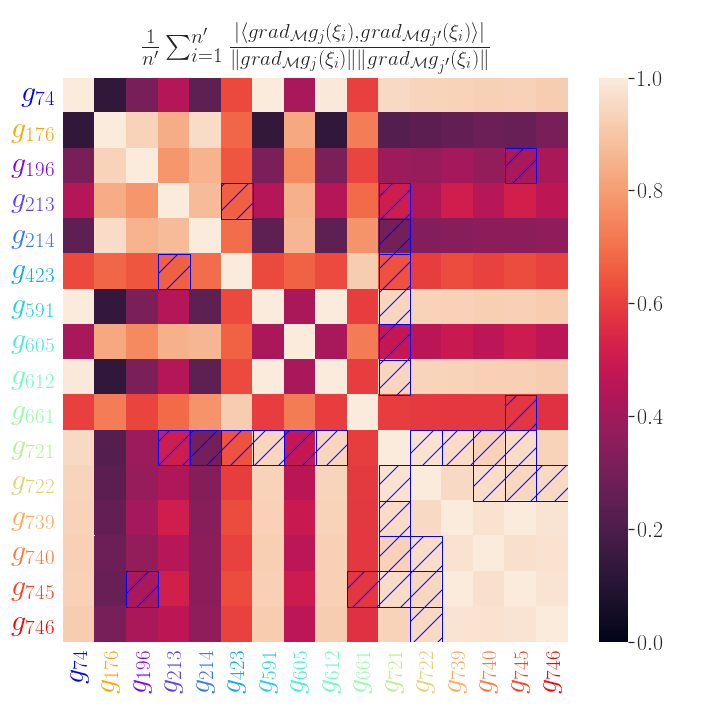}\label{fig:eth-full-mflasso-support-cos}}\hfill 
\newline
\subfloat[]{\includegraphics[width=3cm]{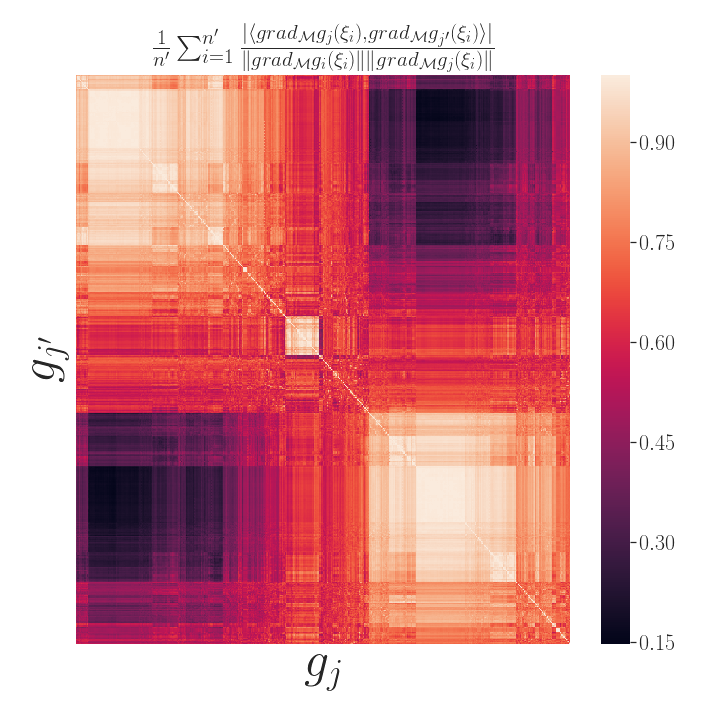}\label{fig:mal-fullcos}}\hfill
\subfloat[]{\includegraphics[width=3cm]{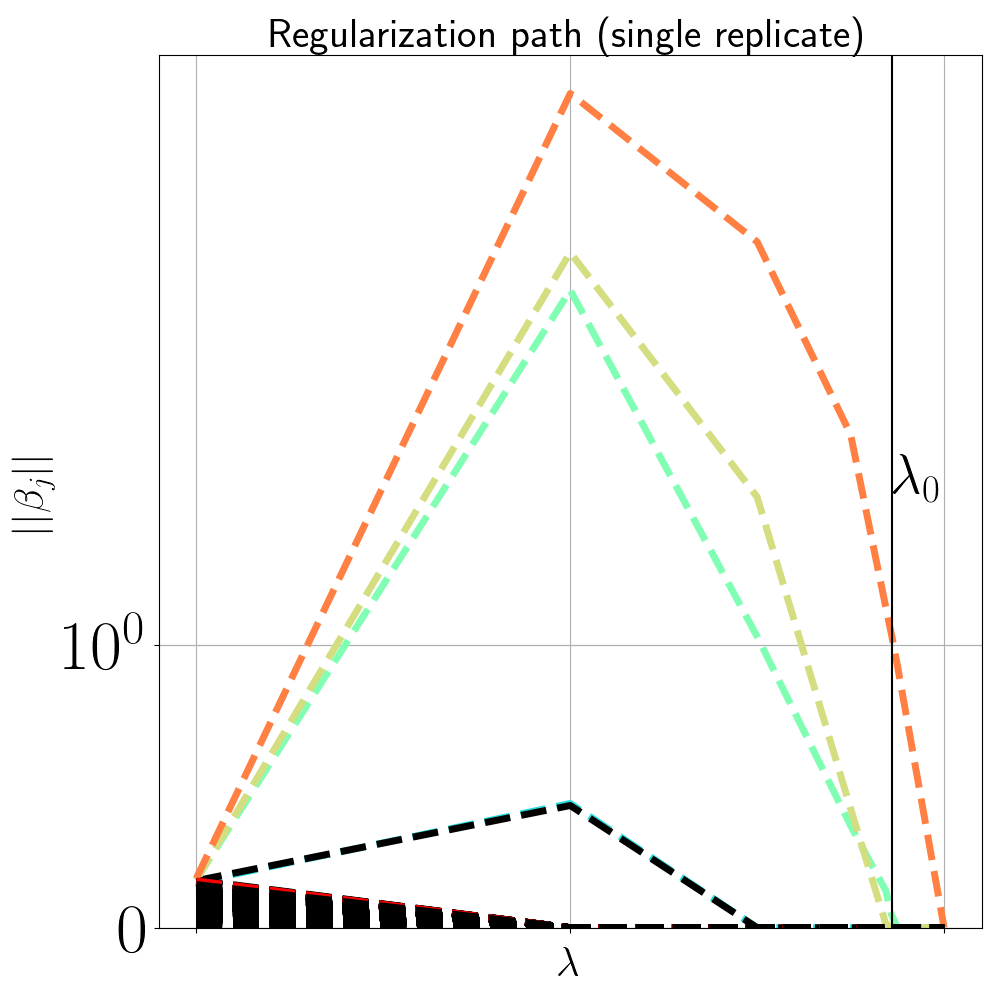}\label{fig:mal_fulldict_replicate}} \hfill
\subfloat[]{\includegraphics[width=3cm]{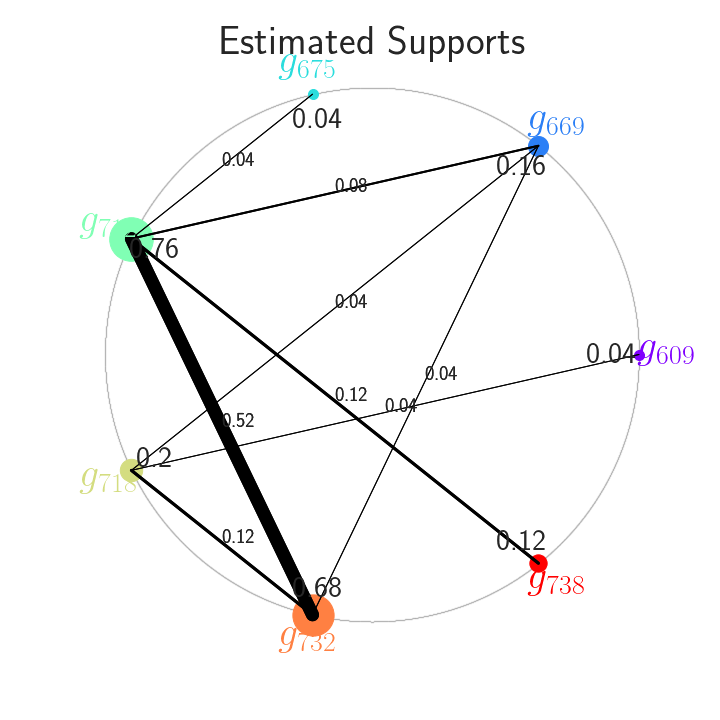}\label{fig:mal-full-mflasso-support}}\hfill
\subfloat[]{\includegraphics[width=3cm]{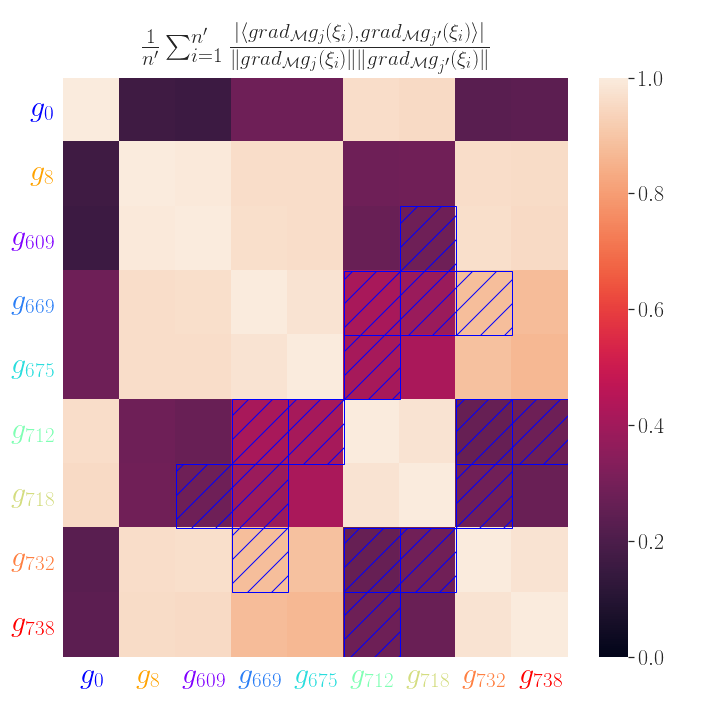}\label{fig:mal-full-mflasso-support-cos}}\hfill 
\caption{Results for MD data with full dictionaries consisting of all possible torsions.
The top and bottom rows show results for \ethdata~ and \maldata, respectively.
Figures \ref{fig:eth-fullcos} and \ref{fig:mal-fullcos} show mean cosine collinearity of dictionary gradients ordered by heirarchical clustering.
Figures \ref{fig:eth_fulldict_replicate} and \ref{fig:mal_fulldict_replicate} show examples of regularization paths for single replicates that select relatively orthogonal functions.
The tuning parameter at which $|S| = d$ is indicated as $\lambda_0$.
Functions are colored if they are selected in any replicate.
Figure \ref{fig:eth-full-mflasso-support} and  \ref{fig:mal-full-mflasso-support} shows support recoveries given by \ouralg~ over different replicates.
Figure \ref{fig:eth-full-mflasso-support-cos} and \ref{fig:mal-full-mflasso-support-cos} and shows mean cosine collinearity of selected supports.
$g_{74,176}$ and $g_{0,8}$ are representative torsions from the true support, while the others are selected in any replicate.
Pairs that are selected in any replicate are marked with a blue box.
}
\label{fig:agnostic-results}
\end{figure}

\newpage

%% file: gradients-jmlr-conclusion.tex
\section{Conclusion}
\label{sec:conc}
The approach of \ouralg~is to reconstruct the
differentials of the manifold coordinates from
differentials of functional covariates. It is robust to
non-linearity in both the algorithm and the covariates. It
requires functions that are smooth, as well as the assumption that the
data lie near a smooth manifold. We estimate the differentials of the
manifold embedding algorithm, but use differentials of functional
covariates that are available analytically. We demonstrate this
approach on molecular dynamics simulations that generate
high-dimensional point clouds sampled from the configuration space of
a given molecule. Our functional covariates are bond torsions, and the
embedding coordinates display a denoised version of the data.
Together, these examples demonstrate the efficacy of \ouralg~ for high-dimensional automated attribution of higher-level features such as bond torsions with data geometry in a feature space given by lower-level features such as planar angles.
It is able to associate high-level features with other functions such as the individual embedding coordinates, as well as the overall geometry determined by the tangent space estimation and entire embedding.
Its limitations are consistent with the general behavior of l1-regularized methods, thus circumventing them with existing tools appears promising. 

Both linear and non-linear dimension reduction methods map data to abstract coordinates, derived from agnostic, intrinsic data properties, such as the covariance matrix, in the case of PCA, or the Laplacian, in the case of the Laplacian Eigenmaps algorithm. By regressing the abstract coordinate functions on a dictionary $\G$ of functions of the data that have meaning in the domain of the problem, we automatically
establish relationships between the learned manifold and domain knowledge. The expert is freed from the tedious work of visually inspecting each possible function $g$ with the manifold coordinates; her expertise is used by specifying covariate functions of the data. The recovered results come with guarantees which can be partially checked in practice. With the obvious simplifications, \ouralg~could also be used to assign explanations to coordinates obtained by PCA. 

Variations of the methods and results presented here could solve a variety of related problems. For example, suppose that two different experiments produce data sets in the same ambient space $\rrr^D$ and that, from these, we learn manifolds $\M_1$ and $\M_2$ which are both 2-tori. Explaining  $\M_{1,2}$ with a dictionary $\G$ can tell us if the manifolds are ``the same'' from the physics point of view. Moreover, one can seek common or overlapping explanations from a single dictionary for different data sources. In other words, by explaining manifolds estimated purely from data with  domain-dependent dictionaries, we produce transferable knowledge, that does not depend on the particularities of the sample, or embedding algorithm, and that can be communicated between experts in the language of their domain.

%% file: gradients-jmlr-appendix.tex
\appendix
\section{Approximating the logarithmic map by orthogonal projection} 
\label{sec:log-map}
In this appendix, we illustrate the details of the approximation to logarithmic map by orthogonal projection in section \ref{sec:pull-back}.  We assume that $\mathcal{M}$ is a submanifold isometrically embedded in $\mathbb{R}^D$.  Also $\mathcal{M}$ is assumed to be at least $C^4$ and compact. The function $\phi$ is assumed to be at least $C^3$. 

Let $\gamma(s)$ be the geodesic pass through a point $\xi$ at $s = 0$ and a different point $\xi'$ in
$\M$ for some $s > 0$, where $s$ is the arc length parameter of the geodesic. Then, the {\em logarithmic map} \citep{DoCarmo} of $\xi'$
w.r.t. to $\xi$ is defined as the vector $l:=s\gamma'(0)\in \T_{\xi}\M$. We denote it by $\log_{\xi}\xi'$. Also, denote by $u$ the orthogonal projection
$\proj_{\T_\xi\M}(\xi'-\xi)$. Then,

\begin{proposition}
For all $\xi$ not on the boundary of $\M$ and all $\xi'$ such that $\lVert \xi' - \xi \lVert \leq r$ for some $r > 0$, it holds that 
  $$\lVert\proj_{\T_{\xi}\M}(\xi'-\xi) - \log_{\xi}\xi'\lVert=o(r), \quad
  \lVert\proj_{\T_{\phi(\xi)}\phi(\M)}(\phi(\xi')-\phi(\xi)) - \log_{\phi(\xi)}\phi(\xi')\lVert=o(r)$$.
\end{proposition}

\begin{proof}

This proposition follows the results in Appendix B. in \citet{coifman:06}.  First, from the
assumption it follows that the geodesic $\gamma\in C^{3}$
\mmp{check this; can we make it C3}\hanyuz{Yes.} The Christofel symbol will be $C^3$ if the manifold is $C^4$. And hence the solution of geodesic equation will be $C^3$ according to standard ODE theory. 

Therefore by Taylor expansion, $\gamma(s)=\gamma(0)+s\gamma'(0)+\frac{s^2}{2}\gamma^{(2)}(0)+\frac{\gamma^{(3)}(\tilde{s})}{6}s^3,$ where $\tilde{s}\in(0,s)$. 
Recall that $\gamma^{(2)}$ is a vector orthogonal to $\T_{\xi}\M$ for a geodesic; moreover, when the manifold $\M$ is $C^3$ and compact, the magnitudes of $\gamma^{(3)}$ is uniformly bounded on $\M$. Note that $\gamma(0)=\xi,\gamma(s)=\xi'$, so $\xi'-\xi=l+O(s^3)$, i.e. $l=\xi'-\xi+O(s^3)$.

Lemma 7 in \citet{coifman:06} implies $\|u\|^2=s^2+O(r^4)$. Therefore,
$s^2=\|u\|^2+O(r^4)\leq \|\xi-\xi'\|^2+O(r^4)\leq
r^2+O(r^4)=O(r^2)$. Hence,
$s^3=\bigOO(r^{3})$. Now consider $l,u$ and
$\xi'-\xi$  as points in $\rrr^D$. We have
that $\|\xi'-\xi-u\|\leq\|\xi'-\xi-l\|$, and by triangle inequality,
$\|l-u\|\leq \|\xi'-\xi-u\|+\|\xi'-\xi-l\|\leq
2\|\xi'-\xi-l\|=o(r)$. Hence, we have shown the first part of the desired result.

Now we turn to $\proj_{\T_{\phi(\xi)}}(\phi(\xi')-\phi(\xi))$. In the
pushforward Riemannian metric $G$, $\phi(\gamma)$ is the geodesic
between $\phi(\xi)$ and $\phi(\xi')$ in $\phi(\M)$.  When $\M$
is compact, and $\phi\in C^3(\M)$, then $\phi$ and the derivatives $\phi'_{1:m}$ are
uniformly continuous, hence the derivatives of $\phi(\gamma)$ remain
bounded by the derivatives of $\gamma$, and
$\|\phi(\xi')-\phi(\xi)\|=\bigOO(r)$. Therefore, we can
apply the previous argument to complete the proof. \hanyuz{Previous arguments assumes that $\phi(\gamma(s)) \in C^3$.}
\end{proof}

\mmp{still to do: collect the assumptions. Can we make M=C3 instead of C infinity? When is phi C1? I think it must be to have RMetric estimator. Less important: to make it a proposition in the appendix; to move the part about phi back to section 4.3. make the o(epsilon) precise. yes, make it a proposition, it will be easier to read.}

\section{The shape space}
\label{app:shape-space}
Here we define the shape space, and show how to obtain the gradient of a function $g_j$ of a molecular configuration, at a non-singular point, in the tangent bundle of this space.

We define the \textit{shape space}
\begin{align*}
 \Sigma_3^{N_a} = \rrr^{3N_a} \slash (E(3) \times \rrr^+ ),
\end{align*}
where $E(3)$ is the three dimensional Euclidean group composed of rigid rotations and translations in $\rrr^3$, and $\rrr^+$ is a dilation factor relative to the mean position of the $N_a$ atoms. That is, $\Sigma_3^{N_a}$ is the space of positions of $N_a$ atoms in $\rrr^3$ with equivalences given by translation, rotation, and dilation. Away from singularities of measure zero, $\Sigma_3^{N_a}$ is a Riemannian manifold \citep{Le1993-mg, addicoatcollins:2010}.

Denote the Euclidean coordinates of $ \rrr^{3N_a}$ by $r$, and the Euclidean position of each data point by $r_i$. Recall that we compute $a_i = a(r_i)$ for $i \in 1, \dotsc n$, where $a$ is the vector-valued function
\begin{align*}
a: \rrr^{3N_a} \to \rrr^{3{N_a \choose 3}}
\end{align*}
that computes the angles formed by all triples of atoms in the molecule. This angular featurization of the data respects the symmetries of the shape space, and embeds the shape space.
We compute bases for the tangent spaces of $\Sigma_3^{N_a}$ as follows. For every analyzed point $i$, we compute the matrix of partial derivatives, also known as the Wilson B-matrix,
\begin{align*}
W_i = \frac{\partial a}{\partial r} (r_i) \in \mathbb R^{3N_a \times \mathbb R^{3 {N_a \choose 3}}}.
\end{align*}
Note that $W_i$ is the transpose Jacobian of $a$. 
This computation is done using automatic differentiation. We then calculate the reduced Singular Value Decomposition
\begin{align*}
W_i = U_i \Lambda_i V_i^T
\end{align*}
where  $\Lambda_i$ is a diagonal matrix of dimension $3N_a - 7$, containing the non-zero singular values of $W_i$. A deductive explanation for the rank of $W_i$ is that translation, rotation, and dilation correspond to a total of 7 degrees of freedom. The $3N_a - 7$ corresponding singular vectors in $V_i$ are a basis for the tangent space $\T_i  \Sigma_3^{N_a}$ in $\rrr^{3 {N_a \choose 3}}$ \citep{addicoatcollins:2010}. Let $a_i = a(r_i)$ for $i \in 1, \dotsc n$. We can then project
\begin{align*}
\grad_{\Sigma_3^{N_a}} g_j (a_i) &= V_i V_i^T \nabla_{a} g_j (a_i),
\end{align*}
where $\nabla_a g_j (a_i)$ is obtained with automatic differentiation using a close-form expression for the dictionary function in the angular coordinates $a$ of $\rrr^{3 {N_a \choose 3}}$, and $\grad_{\Sigma_3^{N_a}}$ is the gradient on the shape manifold in the angular coordinates.

Recall that we apply Principal Component Analysis to the angular features matrix  $a_{1:n} \in \rrr^{n \times D}$. To perform PCA, we use Singular Value Decomposition:
\begin{align*}
a_{1:n} = M \Pi N^T.
\end{align*}
Denote by $P$ the matrix formed with the first $D$ columns of $N$; $P$ projects the angular features into a lower dimension space that reduces redundancy while capturing the vast majority of the variability. \comment{For this paper, we use $D = 50$.} That is, \\
\begin{align*}
\xi_i &= a_i P,\;\text{for}\; i=1,\ldots n.
\end{align*}
The gradient of $g_j$ with respect to coordinates $\xi$ are given by
\begin{align*}
\grad_{\xi} g_j (\xi_i)= P^T \grad_{\Sigma_3^{N_a}} g_j (a_i).
\end{align*}
We use $\grad_{\xi} g_j (\xi_i)$ as $\nabla_{\xi} g_j (\xi_i)$ in \ouralg.

\begin{figure}[H]
\setlength{\picwi}{0.3\llw}
\begin{tabular}{cc} 
\includegraphics[width=4\picwi,height=2\picwi]{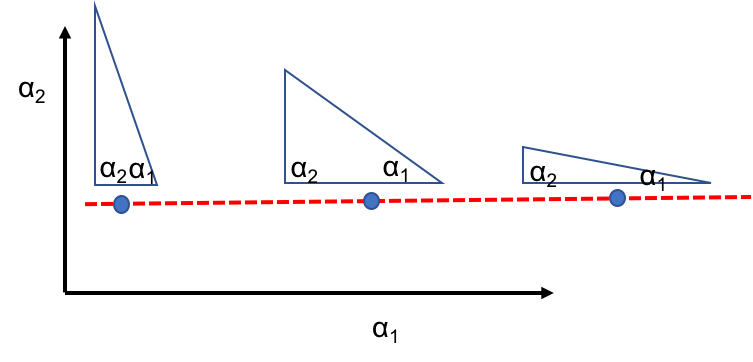}
\end{tabular}
\caption{\label{fig:shapespace} This diagram shows a simplified representation of the neighborhood of a point in the shape space $\Sigma_2^3$. Up to rotation, dilation, and translation, the shape of a triangle is determined by two angles, so we can see that this is a two-dimensional space.  The diagram represents the logarithmic map of a region of $\Sigma_2^3$, with the red line indicating the logarithmic map of the subspace of right triangles, in a coordinate system given by $\alpha_1$ and $\alpha_2$, two angles in the triangle.
}
\end{figure}




\newpage
\section{Torsion Computation}
\label{app:dictionary-details}

\begin{figure}[H]
\includegraphics[scale=0.4,valign=m]{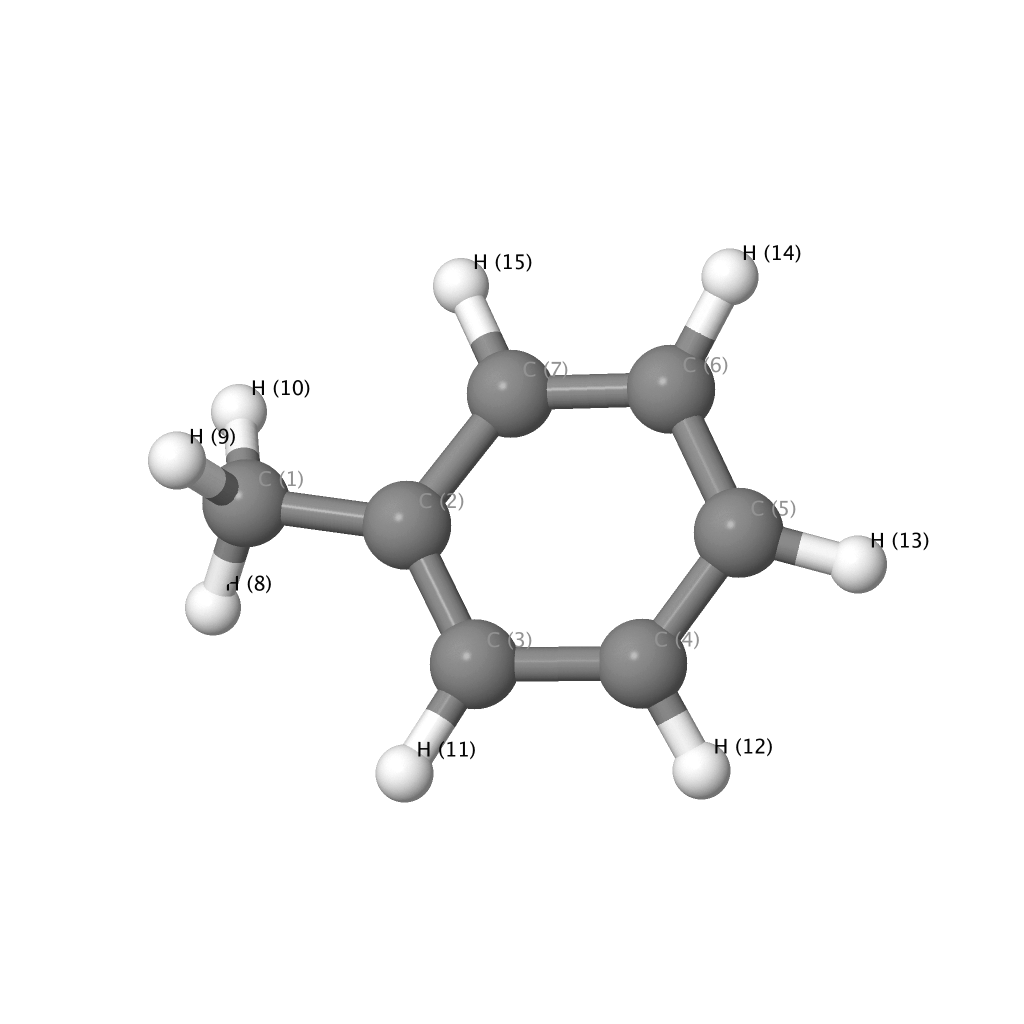}
\caption{Bond diagram of toluene.
}
\label{fig:tol_diagram}
\end{figure}

For molecular dynamics analyses, our dictionary $\G$ consists of bond torsions $g$ (see Figure \ref{fig:molecs}), which are computed from planar angles of the faces of the circumscribing triangles.
These gradients are obtained using automatic differentiation in Pytorch.
Diagrams \ref{fig:eth_diagram} -  \ref{fig:tol_diagram} imply the identities of the bonds input as functional dictionaries to \ouralg~ as a priori-known dictionaries: for these a priori dictionaries, only torsions explicitly shown by $3$ ordered line segments are included.
For example, in Figure \label{fig:eth_diagram}, the ordered atom 4-tuple $[9 ,3,1, 5]$ describes a torsion corresponding to the hydroxyl rotor containing the red oxygen.

As described in Section \ref{sec:md}, association of an ordered atom 4-tuple $(A,B,C,D)$ to a torsion $g(A,B,C,D)$ (where $B$ and $C$ are central, and $A$ and $D$ distal) is not unique.
This is a separate issue from that of merely collinear torsions, and reflects the basic geometric properties of the analysis.
There is an equivalence
\begin{eqnarray*}
g (A,B,C,D) = g (A,C,B,D)  = g (D,C,B,A) = g (D,B,C,A).
\end{eqnarray*}
For example, if $[9 ,3,1, 5]$ is explicitly included in our dictionary, then $[5 ,1,3, 9]$ is not, since these are in fact the same function.
Thus, each set of $4$ atoms defines $6$ torsions upon ordering, since we have ${4 \choose 4}$ ordered $4$-tuples, and equivalences of groups of $4$.
This is understandable geometrically by the fact that a tetrahedron (the shape defined by $4$ points) has $6$ edges, and therefore $6$ torsions.

The following figures show zoomed in versions of the molecules with atomic numberings.
  
\begin{figure}[H]
\includegraphics[scale=0.4,valign=m]{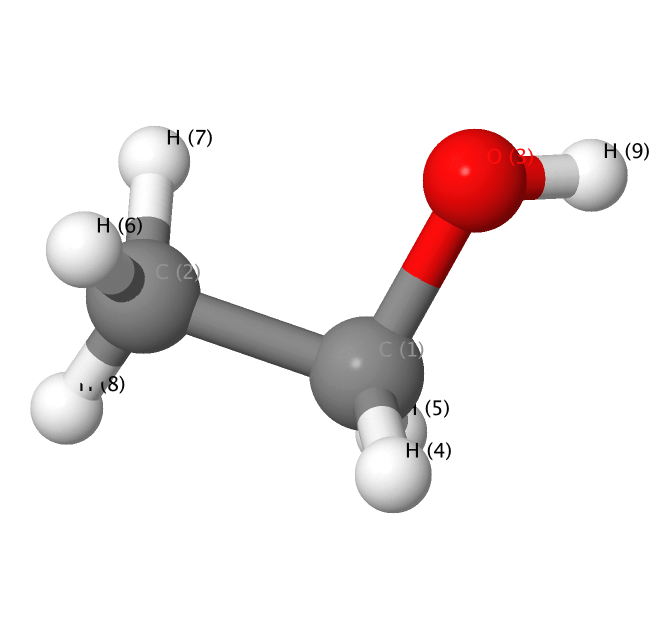}
\caption{Bond diagram of ethanol.
}
\label{fig:eth_diagram}
\end{figure}

\begin{figure}[H]
\includegraphics[scale=0.4,valign=m]{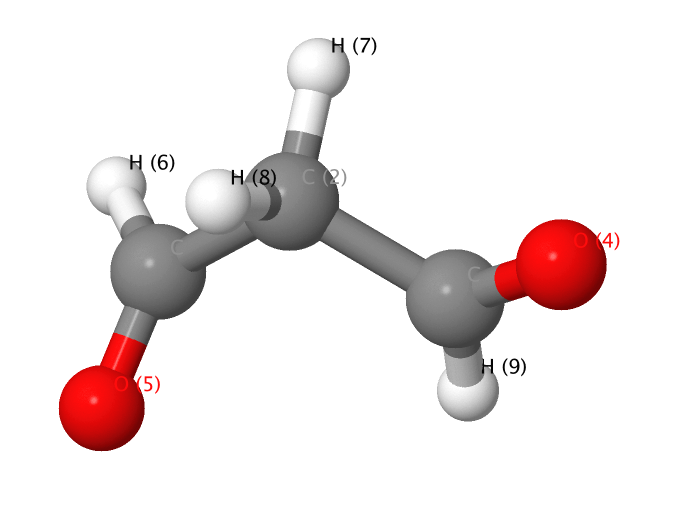}
\caption{Bond diagram of malonaldehyde
}
\label{fig:mal_diagram}
\end{figure}

\newpage

\section{Feature space}
\label{app:feature-space}

In order to demonstrate the multiscale non-i.i.d. noise and non-trivial topology and geometry of our data in the PCA feature space, we display scatterplots of pairs of the top features in our feature space $\mathbb R^D$ containing data points $\xi$.
Recall that PCA is applied as a preprocessing step prior to \ouralg, and so the PCA coordinates therefore form our feature space.
PCA coordinates have a natural ordering given by their corresponding eigenvalues, and so we are able to plot the 'top' coordinates.
Note also that the manifolds are relatively thin in comparison to some noise dimensions; in other words the manifold {\em reach} is of the same scale as the noise.

\begin{figure}[H]
\includegraphics[scale=0.2,valign=m]{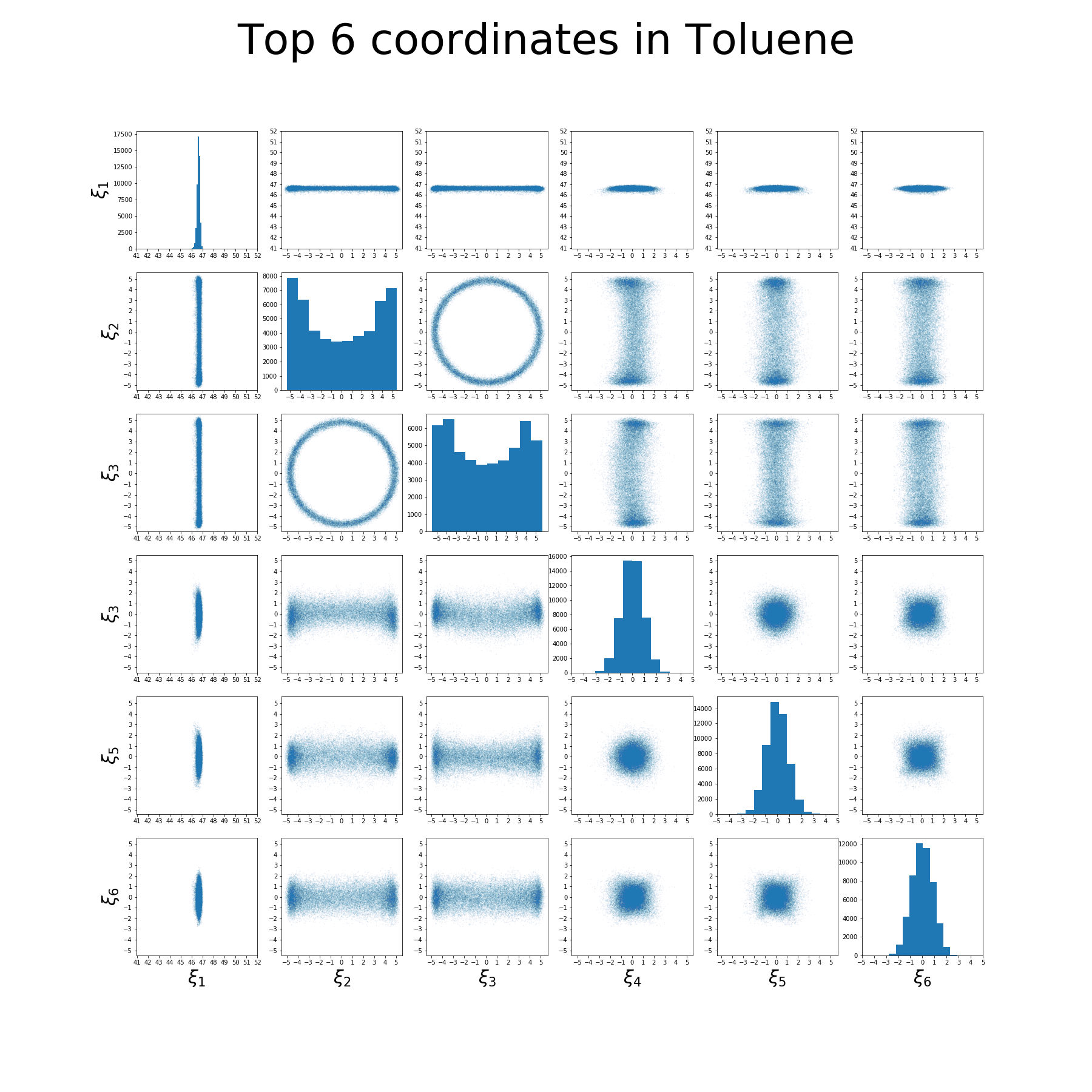}
\caption{First $6$ coordinates in $\mathbb R^D$ output by PCA for Toluene.
}
\label{fig:tol_feature}
\end{figure}

\begin{figure}[H]
\includegraphics[scale=0.2,valign=m]{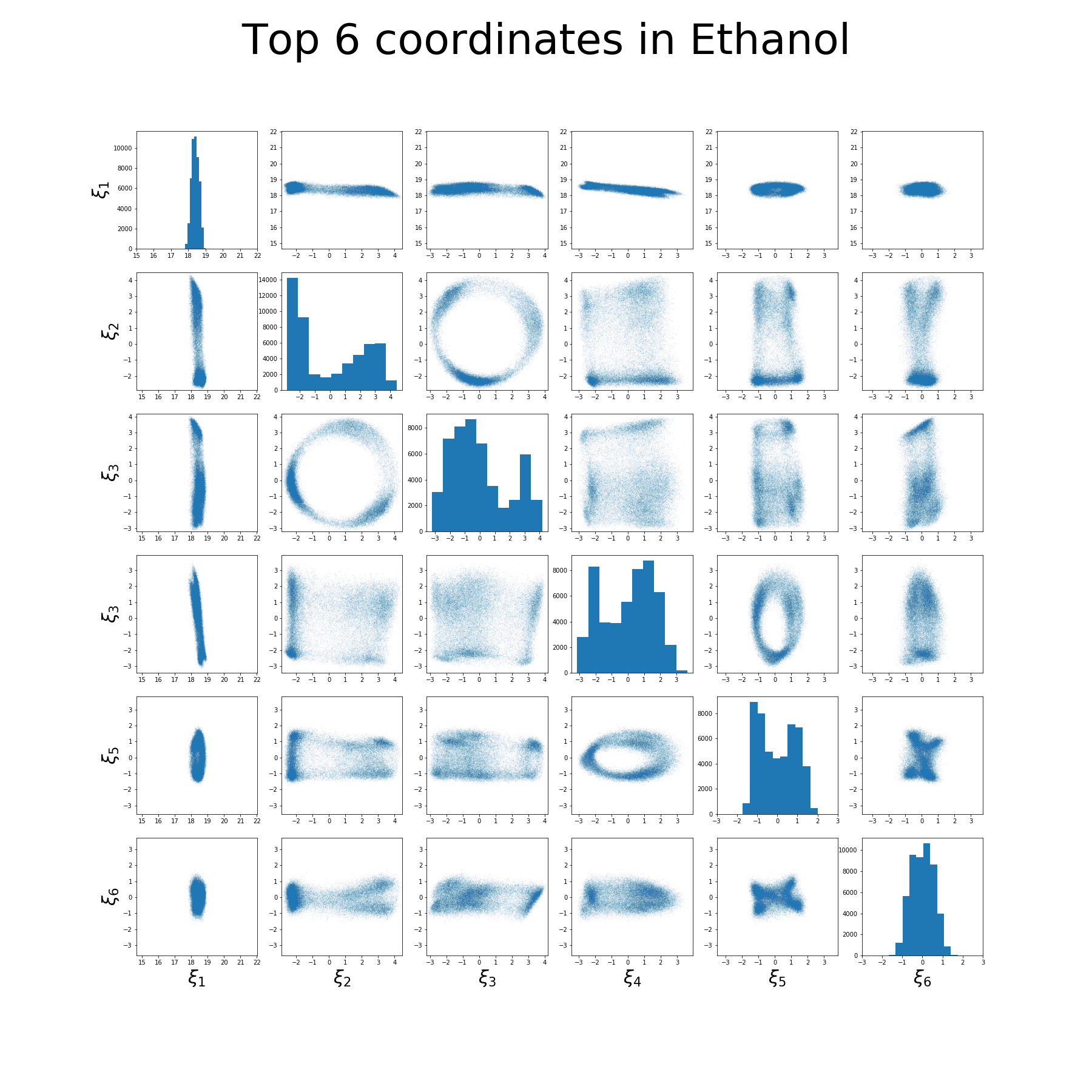}
\caption{First $6$ coordinates in $\mathbb R^D $output by PCA for Ethanol.
}
\label{fig:eth_feature}
\end{figure}

\begin{figure}[H]
\includegraphics[scale=0.2,valign=m]{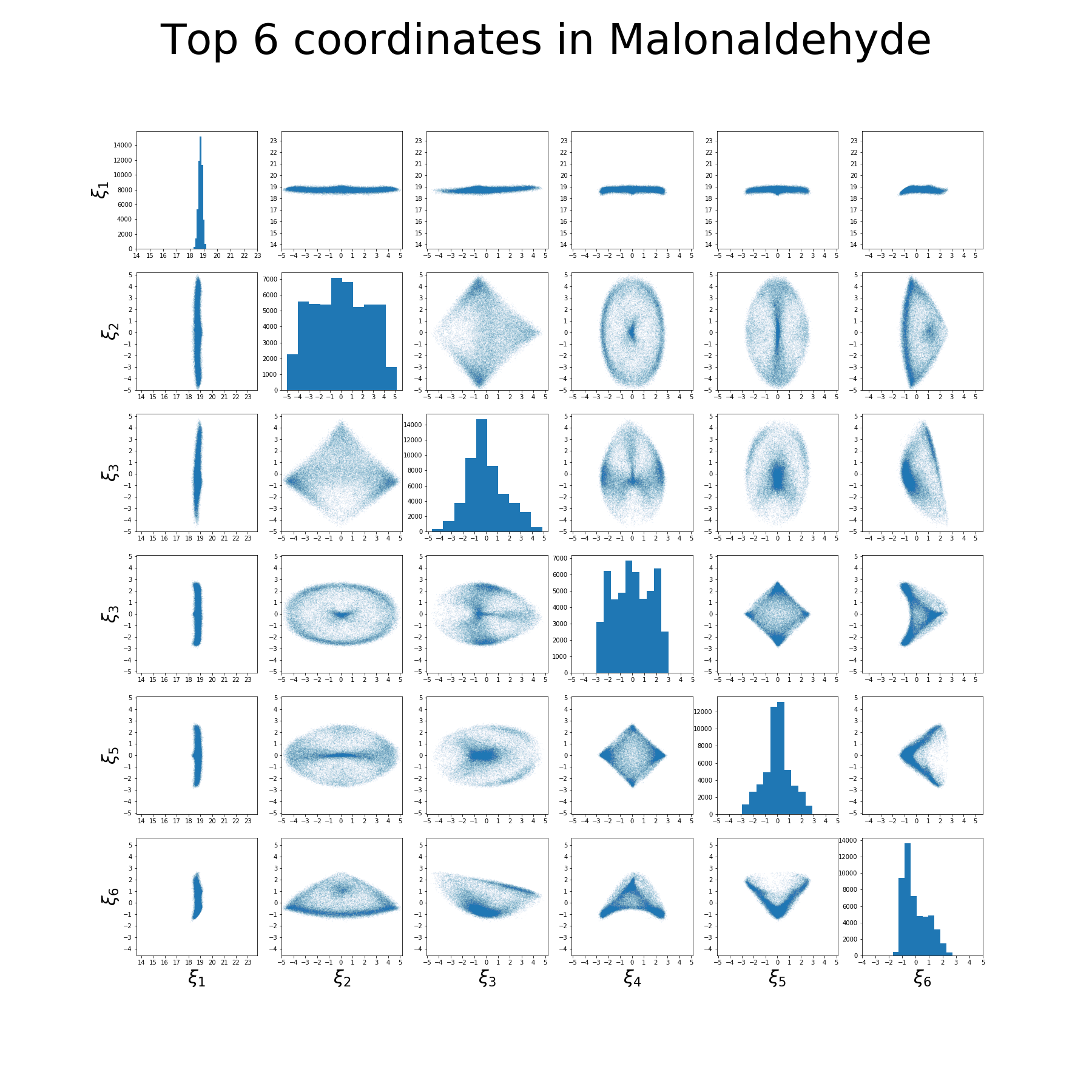}
\caption{First $6$ coordinates in $\mathbb R^D$ output by PCA for Malonaldehyde.
}
\label{fig:mal_feature}
\end{figure}

\newpage

\section{Group sparse basis pursuit}
\label{app:gsbp}

As mentioned in section \ref{sec:variants}, the combinatorial \textit{group sparse basis pursuit problem}
\beq
\label{eq:bp2}
\arg \min_{\beta :s = d } \sum_{j = 1}^p \|\beta_j\| s.t. \grad \phi_k (\xi_i) = \sum_{j = 1}^p \beta_{ijk}  \grad_{T_i^\M} g_j (\xi_i) \quad \text{for all $i=1:n$, and $k=1:m$.}
\eeq
has a natural duality with our approach.
That is, for each value of $\lambda$, there is a corresponding constraint ball of radius $\epsilon$ such the solution of the lasso problem is also the solution of the 
\beq
\label{eq:bpdn}
\arg \min_{\beta} \sum_{j = 1}^p \|\beta_j\| s.t.  \sum_{i = 1}^n \sum_{k = 1}^m \| \grad \phi_k (\xi_i) - \sum_{j = 1}^p \beta_{ijk}  \grad_{T_i^\M} g_j (\xi_i)\|_2^2 < \epsilon.
\eeq
This is a combinatorial problem without restriction on the cardinality of the selected support.
It can be exactly solved using the convex regularized approach.

This cardinality-unrestricted version of Program \eqref{eq:bp2} has several interesting properties \citep{Candes2007-ie}.
First, it favors gradients that are orthogonal and evenly varying.
This matches our intuitively notion of what is a ``good'' explanation and mathematically corresponds to the notion of isometry.
Second, it has a clear but not entirely obvious relation to the $l2$ error of our method applied to gradients in $\mathbb R^D$ rather than $\mathbb R^d$, in the sense that dictionary functions which are non-tangent to $\M$ will accrue a higher penalty.
Third, expected minimizers of such dual problems are used to define optima for sparse estimation \citet{Meinshausen2008-ez}.
We can thus use the empirical estimate of this minimizer in the cardinality-restricted setting to provide a useful notion of what is a ``good''  support beyond simply having low pairwise collinearity.

Compared with the standard duality, the major distinguishing feature of our support recovery setup is our theoretical knowledge of the cardinality of the desired support.
Unfortunately, we sometimes observe that the shrinkage caused by using a high $\lambda$ to restrict support size causes the recovered support to not be close to the sample optimum of \eqref{eq:bp2}.
Dictionary functions with large projection $ ( \sum_{i=1}^n \sum_{k = 1}^m (\grad_{T_i^\M} g_j (\xi_i)) ^T (\grad_{T_i^\M} \phi_m (\xi_i)) )^{1/2}$ tend to appear early the regularization path, regardless of their orthogonality or consistency of variation.
Problems with shrinkage including variable selection inconsistency at large $\lambda$, as well as the desirable properties of an intermediate value, are well-established in the support recovery and sparse coding literature \citep{Chen2001-hh, Hesterberg2008-hm, Breheny2011-ci, Lederer2015-pr, Hastie2015-sl}.

We can empirically adapt our method to respond to this problem while still leveraging the advantages of the convex algorithm.
We follow \citet{Hesterberg2008-hm} in using an intermediate $\lambda$ as a variable filtering step prior to variable selection, in our case, using \eqref{eq:bp2}.
In this adapted approach, we initially prune the dictionary using a \ouralg-type step at an intermediate $\lambda$ value.
We then run Program \eqref{eq:bp2} on the pruned dictionary.
Initial selection of $d' << p$ using our approach prior to selection of $d$ variables using \eqref{eq:bp2} is often more effective in obtaining the empirical minimizer of \eqref{eq:bp2} than \ouralg~ on its own, and is much more computationally feasible than running \eqref{eq:bp2} on the entire dictionary.
The $\lambda$ at which these $d'$ functions are obtained is somewhat arbitrary, since a fully data-driven approach would require computation of Program \eqref{eq:bp2} on the entire dictionary, but relatively generic theoretical arguments provide blanket arguments in favor of $\lambda > O(\log p)$ \citep{Chen2001-hh}.
We in general find relatively wide regions of relatively low cardinality, and substantial improvements in the combinatorial loss with minimal computational burden at $\lambda = \lambda_{\text{max}} / 2$.
Results for this two-stage method for \ethdata~ and \maldata~ are displayed in Figure \ref{fig:sup-brute}.

\newpage

\section{Supplemental Experiments} \label{app:sup_exp}

\paragraph{Coordinate-association in a priori dictionaries}

We show the association of individual embedding coordinates to dictionary functions in \ethdata~ and \maldata.
In contrast to \maldata, but similar to \srdata, \ethdata~ has a distinct association of embedding coordinates with dictionary functions.
In particular, $\phi_3$ is associated with different torsions from $\phi_1$ and $\phi_2$.
This is clearly evident in Figure \ref{fig:molecs}.
In \maldata, there is no clear association with embeddings coordinates.
Note that this would also be true for \toldata, as Figure \ref{fig:molecs} clearly shows a circular manifold symmetric in $\phi_1$ and $\phi_2$.

\begin{figure}[H]
\includegraphics[scale=0.15,valign=m]{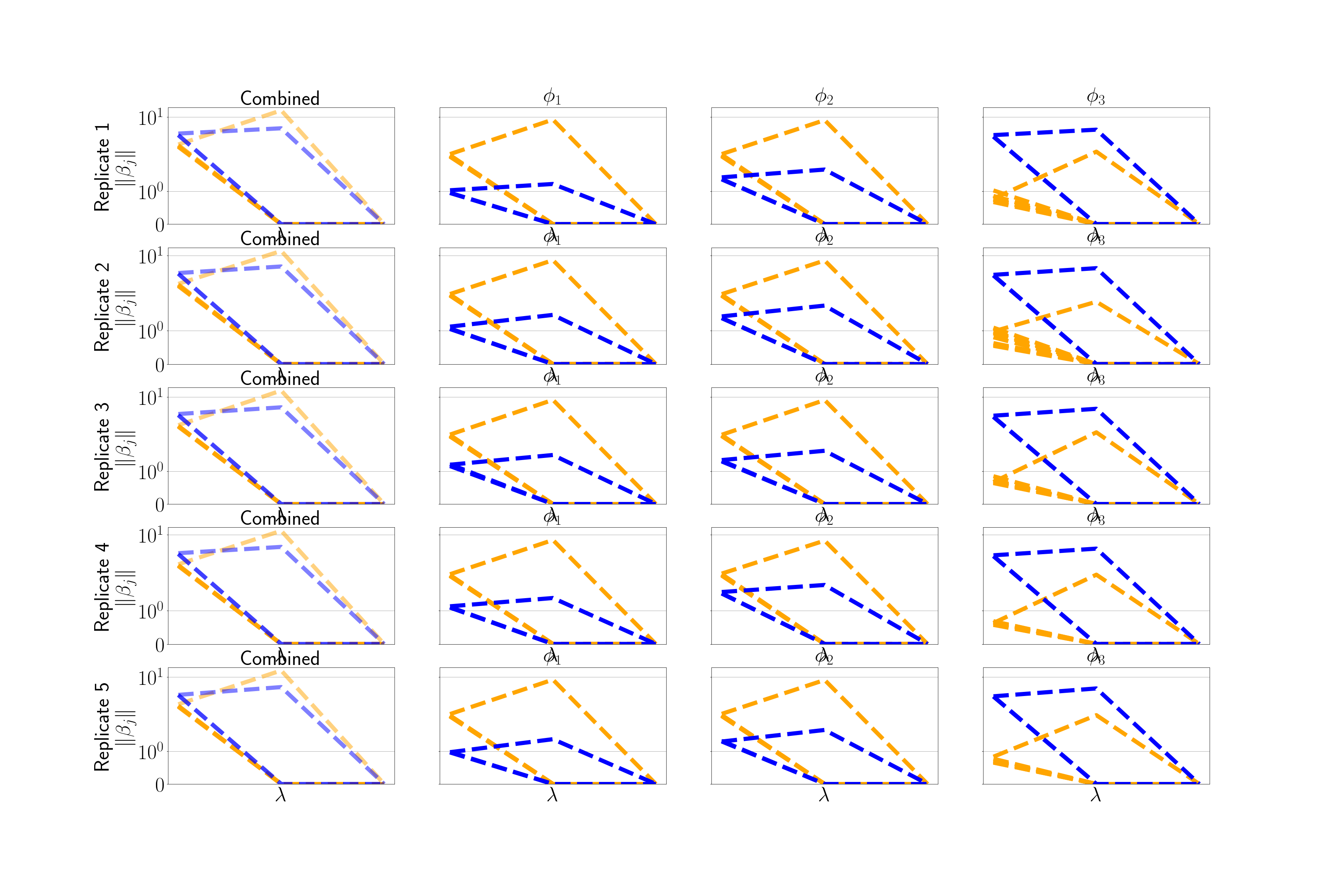}
\caption{Combined and coordinate-specific regularization paths in five replicates of \ouralg~ for \ethdata~ with dictionary given by the bond diagram.  There is a clear association of the blue torsion with $\phi_3$, and orange with $\phi_{1,2}$.
}
\label{fig:eth_phis_regpath}
\end{figure}

\begin{figure}[H]
\includegraphics[scale=0.15,valign=m]{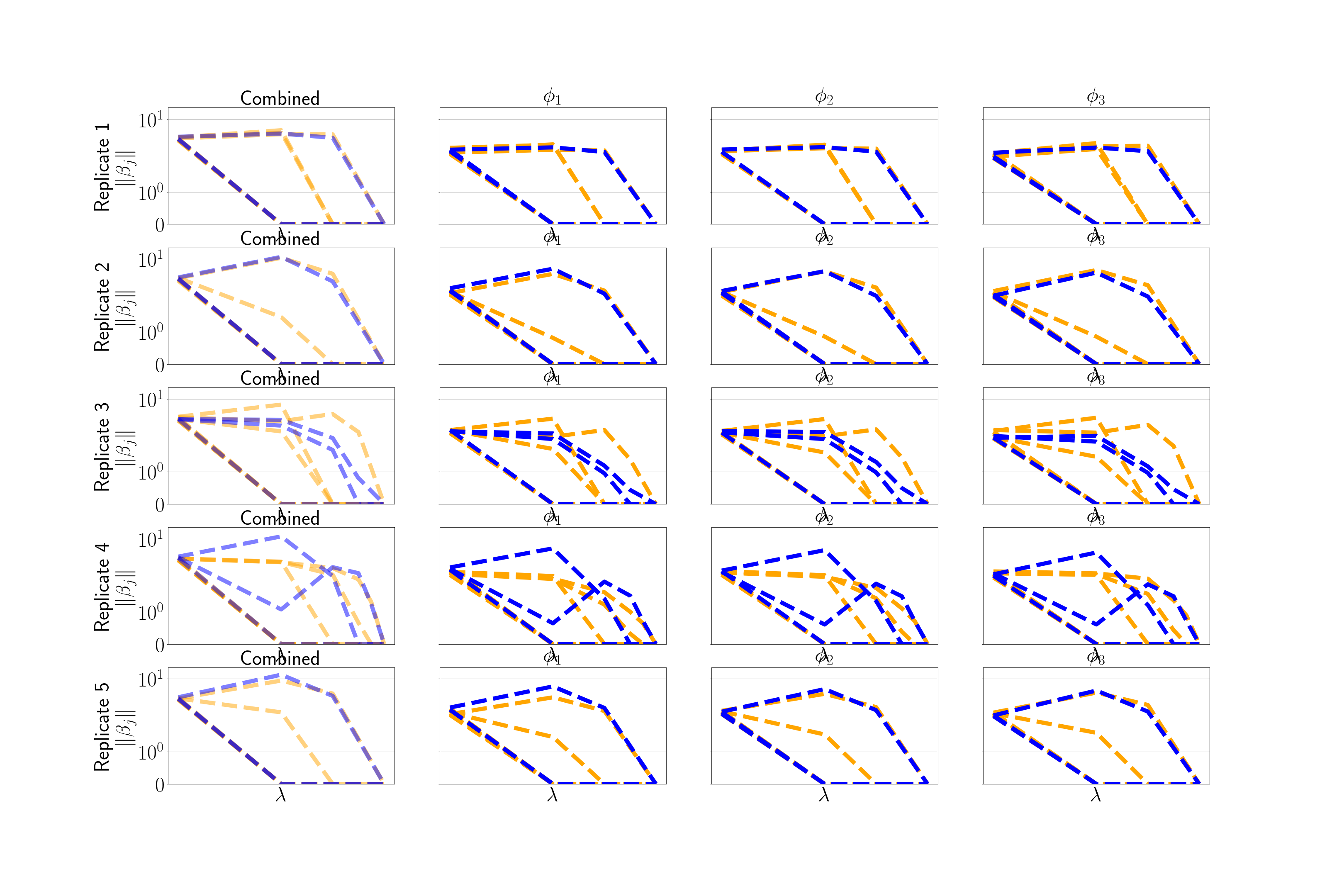}
\caption{Combined and coordinate-specific regularization paths in five replicates of \ouralg~ for \maldata~ with dictionary given by the bond diagram.  There is no clear association of embedding coordinates and covariates.
}
\label{fig:mal_phis_regpath}
\end{figure}

\newpage

\paragraph{Two-stage method results}

The two-stage approach has great success at obtaining highly orthogonal solutions collinear with the true support at minimal computational cost.
Selected functions can be compared with visualizations in Appendix \ref{app:ts_full_visual}.
This shows a clear correspondence between variable selection using the two-stage method and our visual intuition (based on the colored embedding) about what is a good support.

\begin{figure}[htb]
\subfloat[]{\includegraphics[width=5cm]{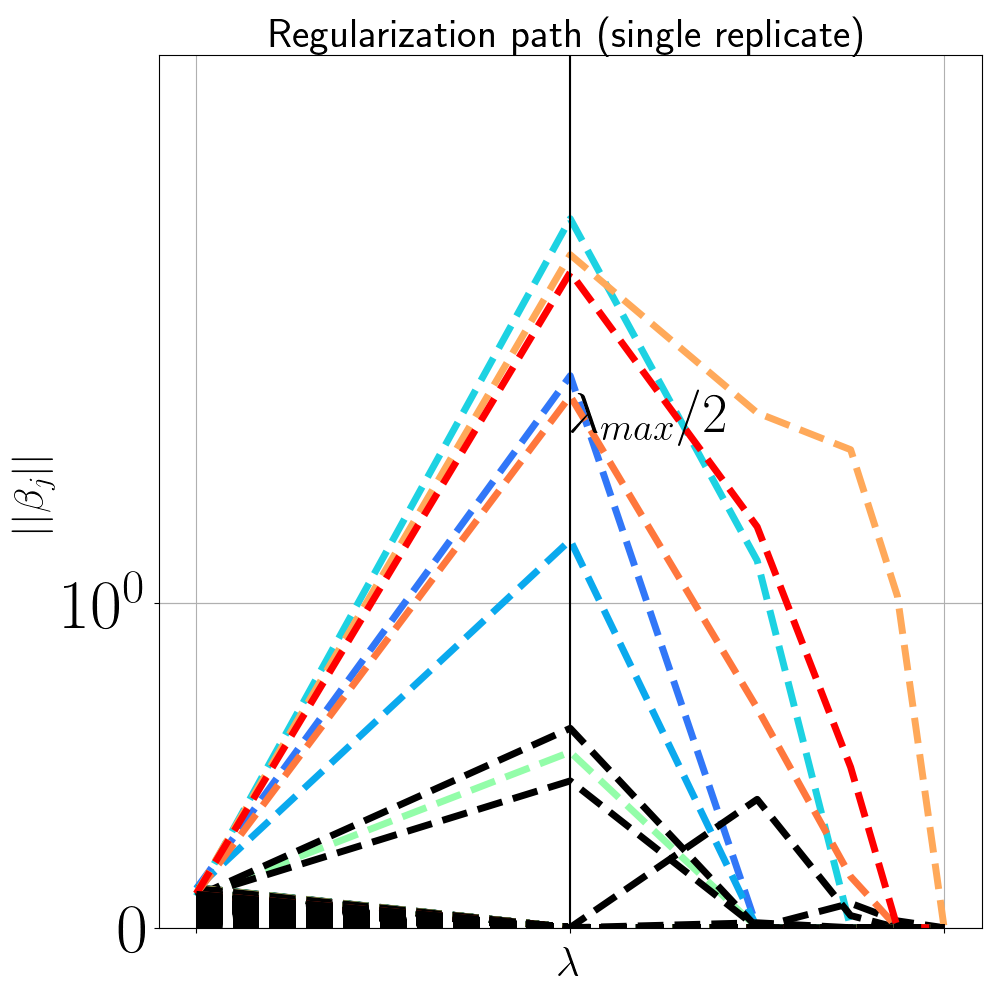}\label{fig:sup-eth-coeff}}\hfill
\subfloat[]{\includegraphics[width=5cm]{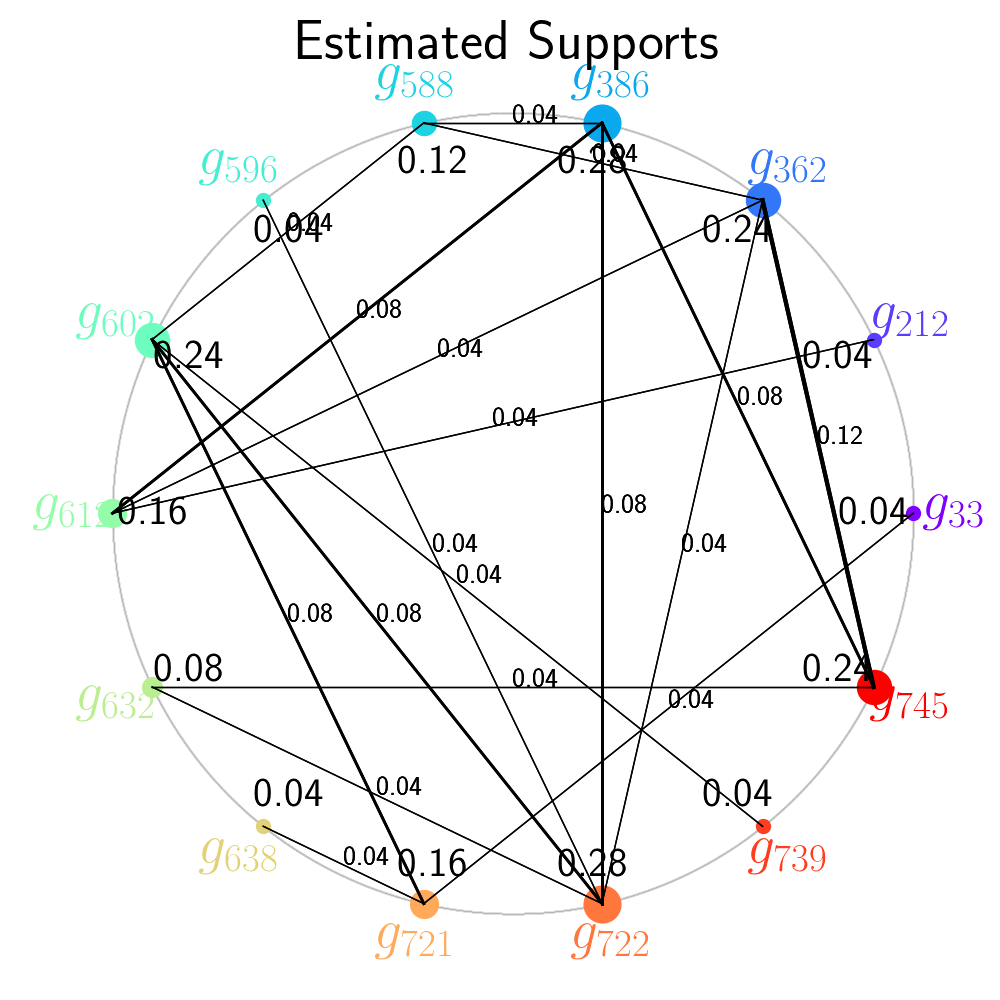}\label{fig:sup-eth-sel}}\hfill 
\subfloat[]{\includegraphics[width=5cm]{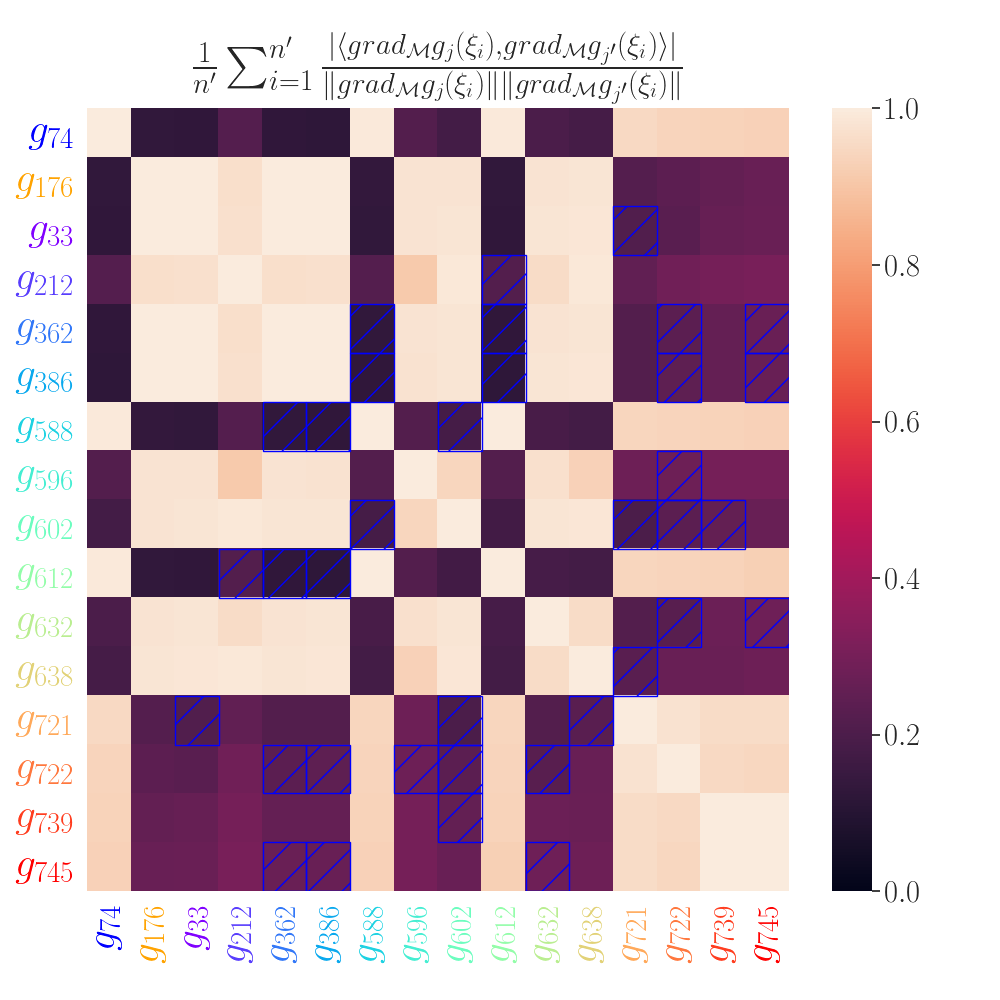}\label{fig:sup-eth-cosel}}\hfill
\newline
\subfloat[]{\includegraphics[width=5cm]{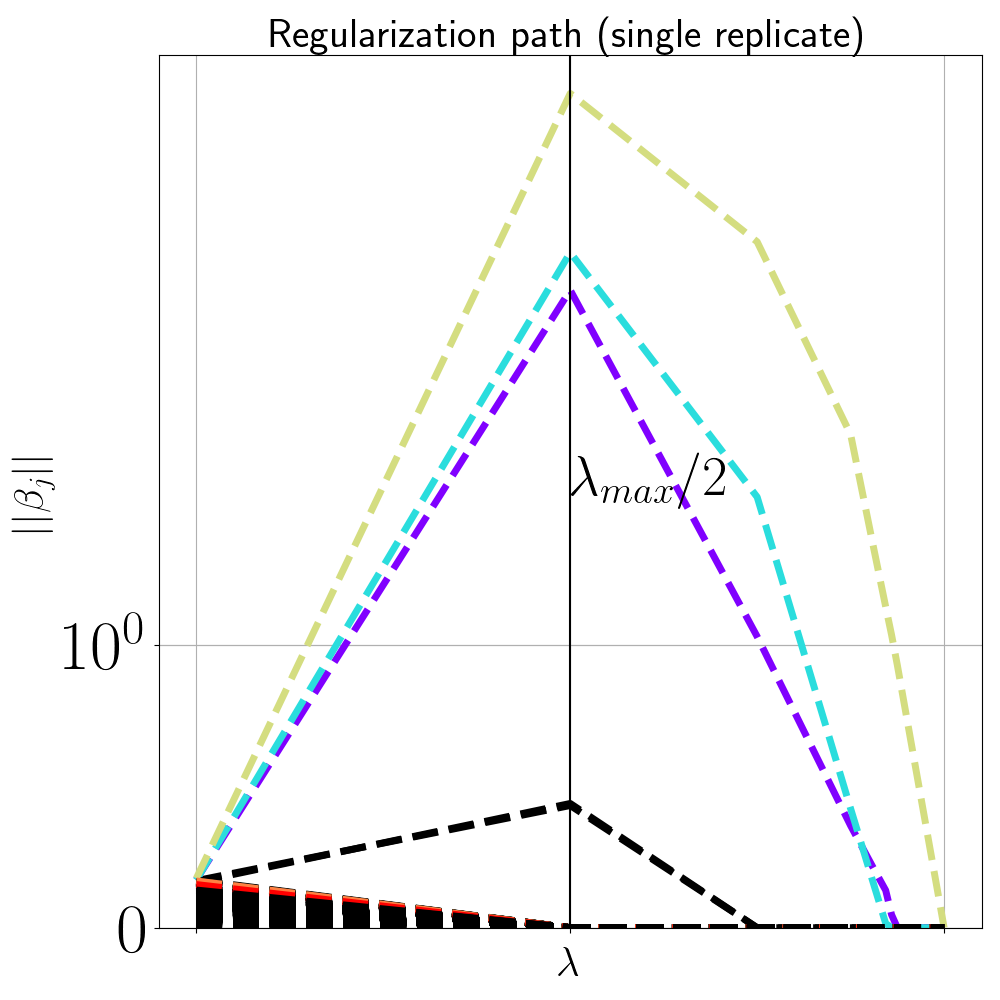}\label{fig:sup-mal-coeff}}\hfill
\subfloat[]{\includegraphics[width=5cm]{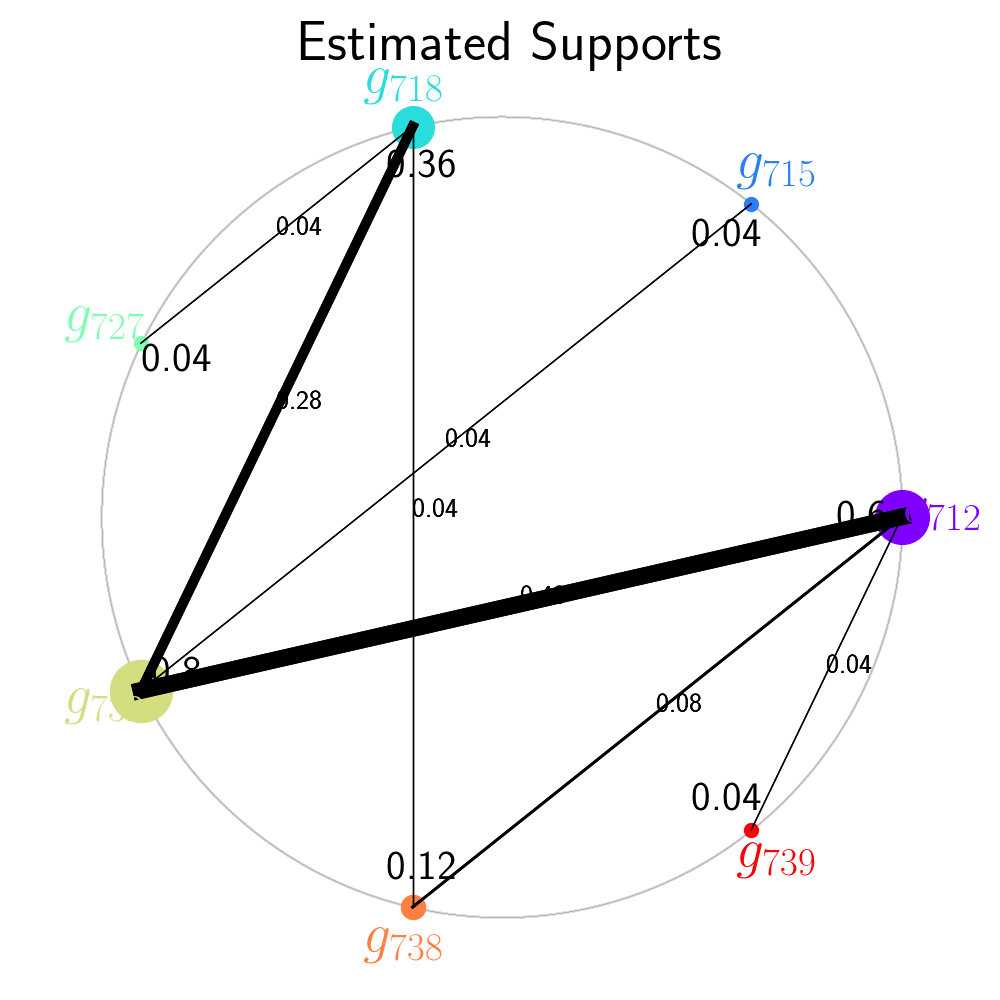}\label{fig:sup-eth-sel}}\hfill
\subfloat[]{\includegraphics[width=5cm]{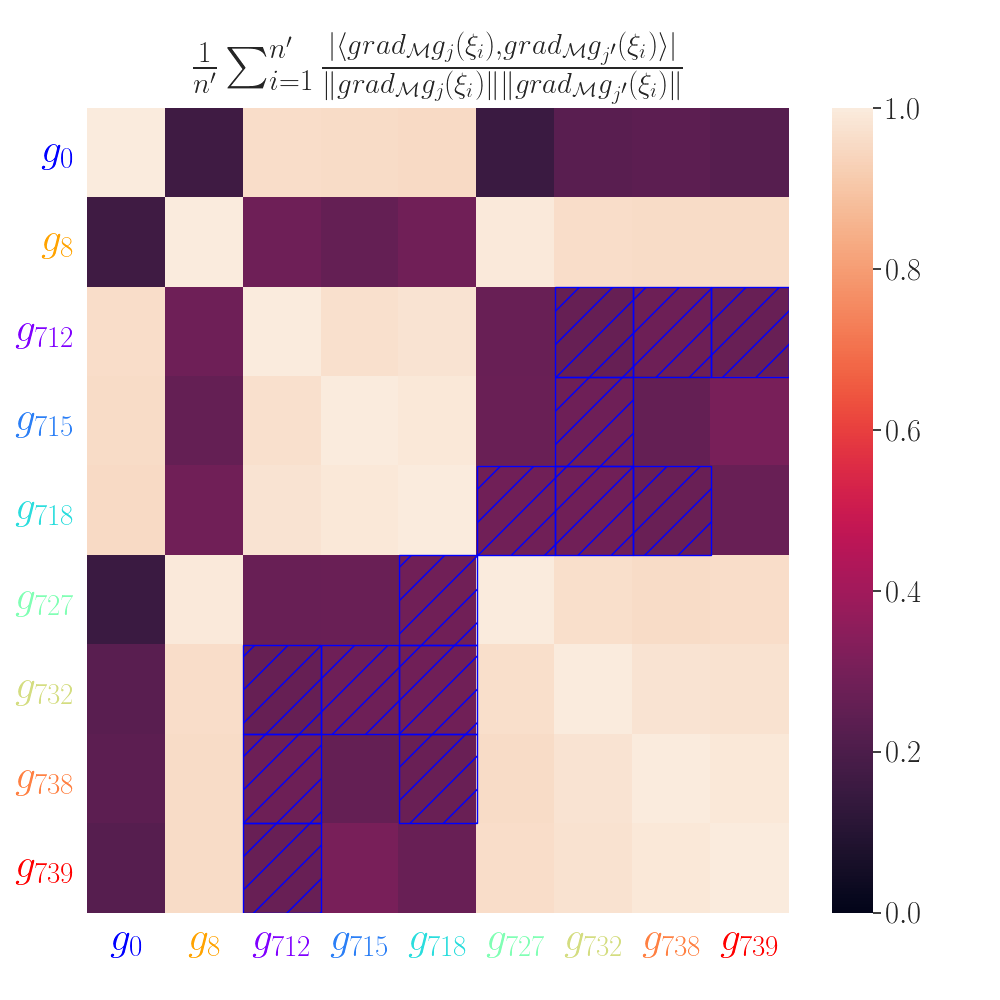}\label{fig:sup-mal-cosel}}\hfill
\caption{Two-stage results for \ethdata~ and \maldata, respectively, with dictionaries given by all possible torsions.
Figures \ref{fig:sup-eth-coeff} and \ref{fig:sup-mal-coeff} show individual replicates, with intermediate tuning parameter value $\lambda_{\text{max}} / 2$.  
Colors are plotted for functions selected by subsequent combinatorial analysis.
Figure \ref{fig:sup-eth-cosel} and  \ref{fig:mal-full-mflasso-support} shows support recoveries given by subset selection using group lasso at $\lambda_{\text{max}} / 2$ followed by Program \eqref{eq:bp2} over $\omega = 25$ different replications.
Figure \ref{fig:eth-full-mflasso-support-cos} and \ref{fig:mal-full-mflasso-support-cos} and shows mean cosine collinearity of selected supports.
$g_{74,176}$ and $g_{0,8}$ are representative torsions from the true support, while the others are selected in any replicate.
Pairs that are selected in any replicate are marked with a blue box.}
\label{fig:sup-brute}
\end{figure}

\newpage

\paragraph{Visualizing functions selected by \ouralg~ from full dictionaries}

We can visualize the selected torsions in the manifold embedding coordinates. The identities of the selected torsions can be compared with the bond diagrams in Appendix \ref{app:dictionary-details}.  We first visualize the functions selected using \ouralg~ from Figure \ref{fig:agnostic-results}.

\label{app:mf_full_visual}

\begin{figure}[H]
\includegraphics[width = 6.5in,valign=m]{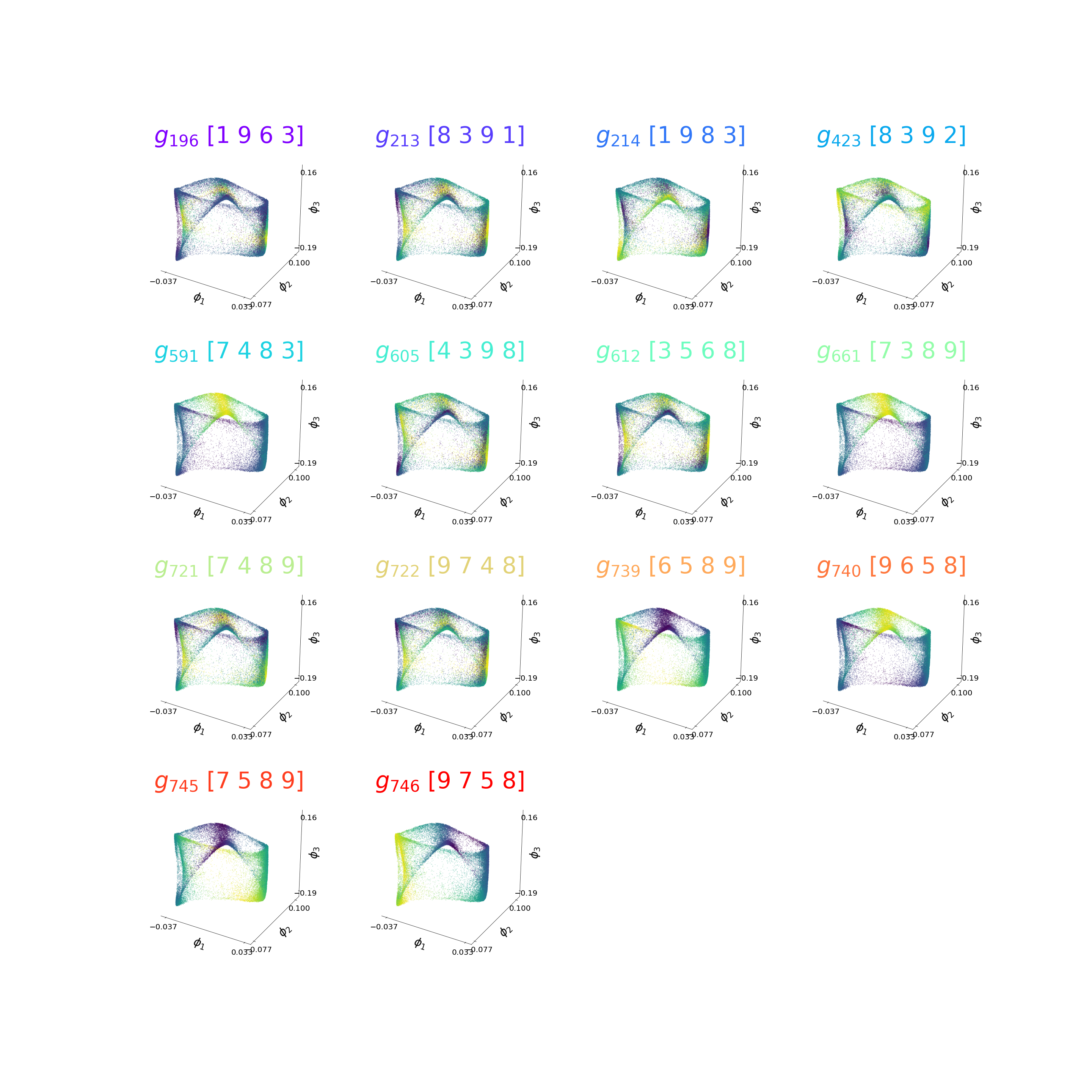}
\caption{\ethdata~ support estimated using \ouralg~ with full dictionary.  Colors should be compared with Figure \ref{fig:agnostic-results}.
}
\label{fig:eth_full_sup_mflasso}
\end{figure}

\begin{figure}[H]
\includegraphics[width = 6.5in,valign=m]{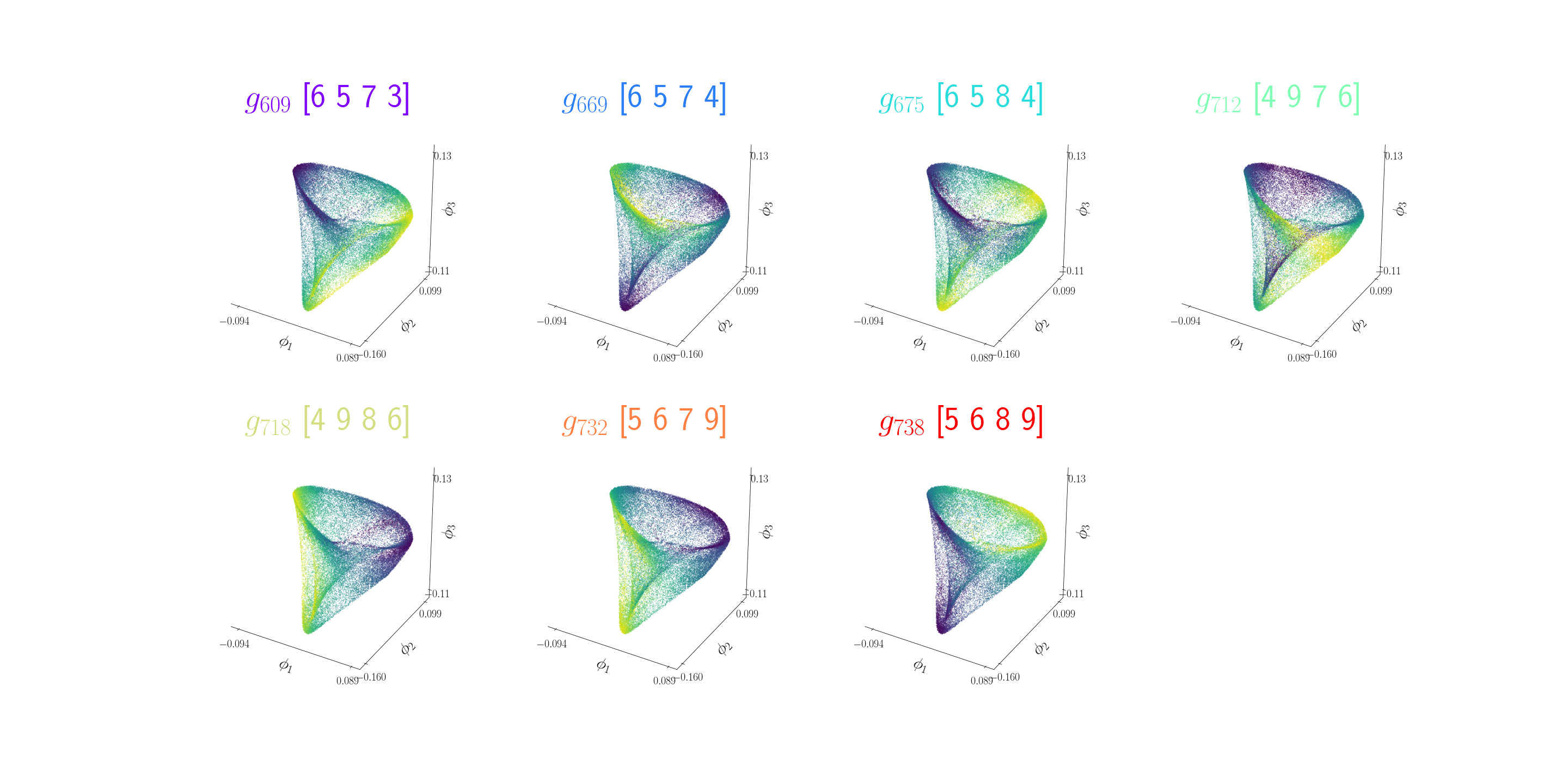}
\caption{\maldata~ support using using \ouralg~ with full dictionary. Colors should be compared with Figure \ref{fig:agnostic-results}.}
\label{fig:mal_full_sup_mflasso}
\end{figure}

\newpage

\paragraph{Visualizing functions selected by the two-stage method from full dictionaries}

We also visualize the functions selected using \ouralg~ from Figure \ref{fig:sup-brute}.  Selected pairs of functions are in \ethdata~ more orthogonal than found using \ouralg.
\label{app:ts_full_visual}

\begin{figure}[H]
\includegraphics[width = 6.5in,valign=m]{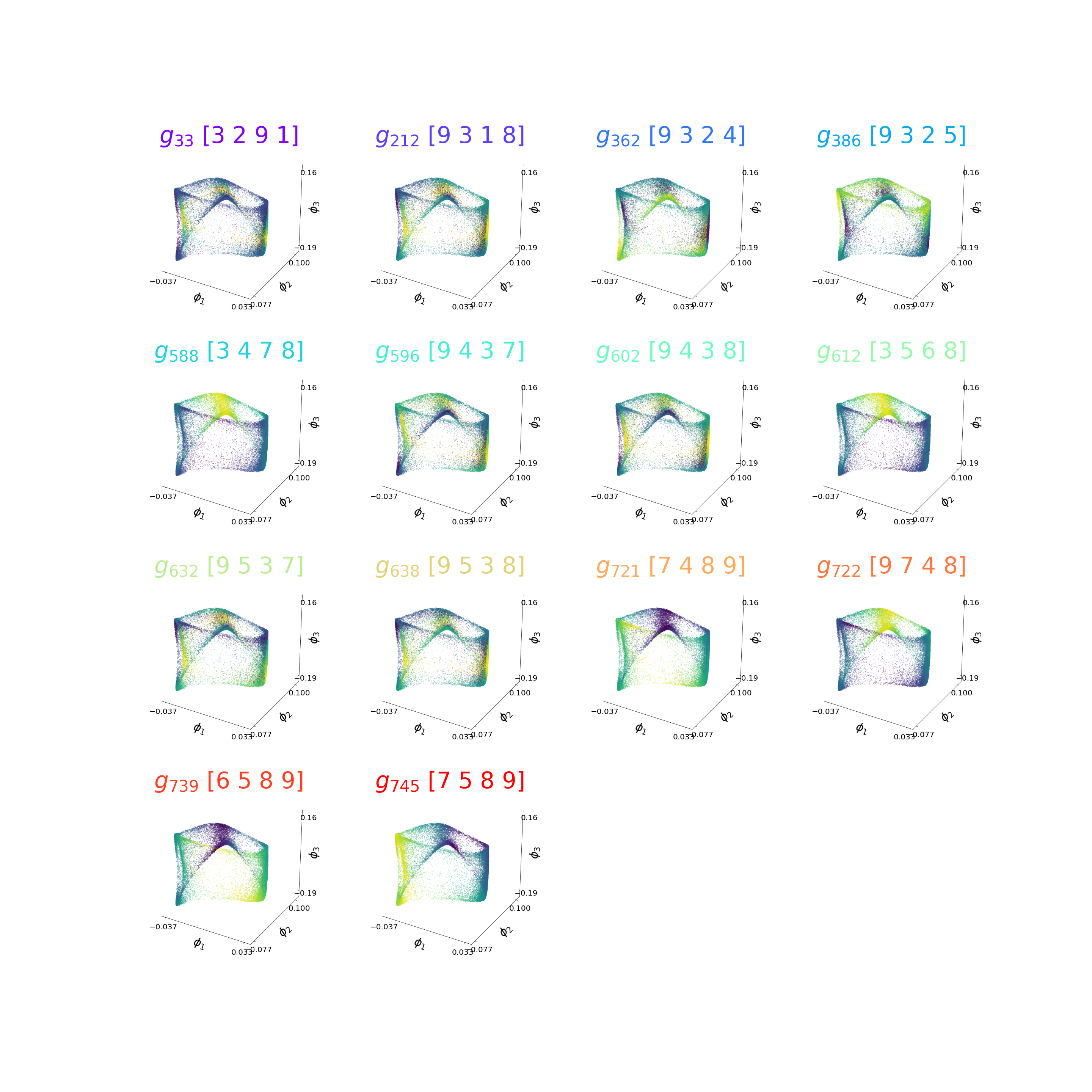}
\caption{\ethdata~ support using basis pursuit on superset obtained using \ouralg. Colors should be compared with Figure \ref{fig:sup-brute}. }
\label{fig:eth_full_sup_brute}
\end{figure}

\begin{figure}[H]
\includegraphics[width = 6.5in,valign=m]{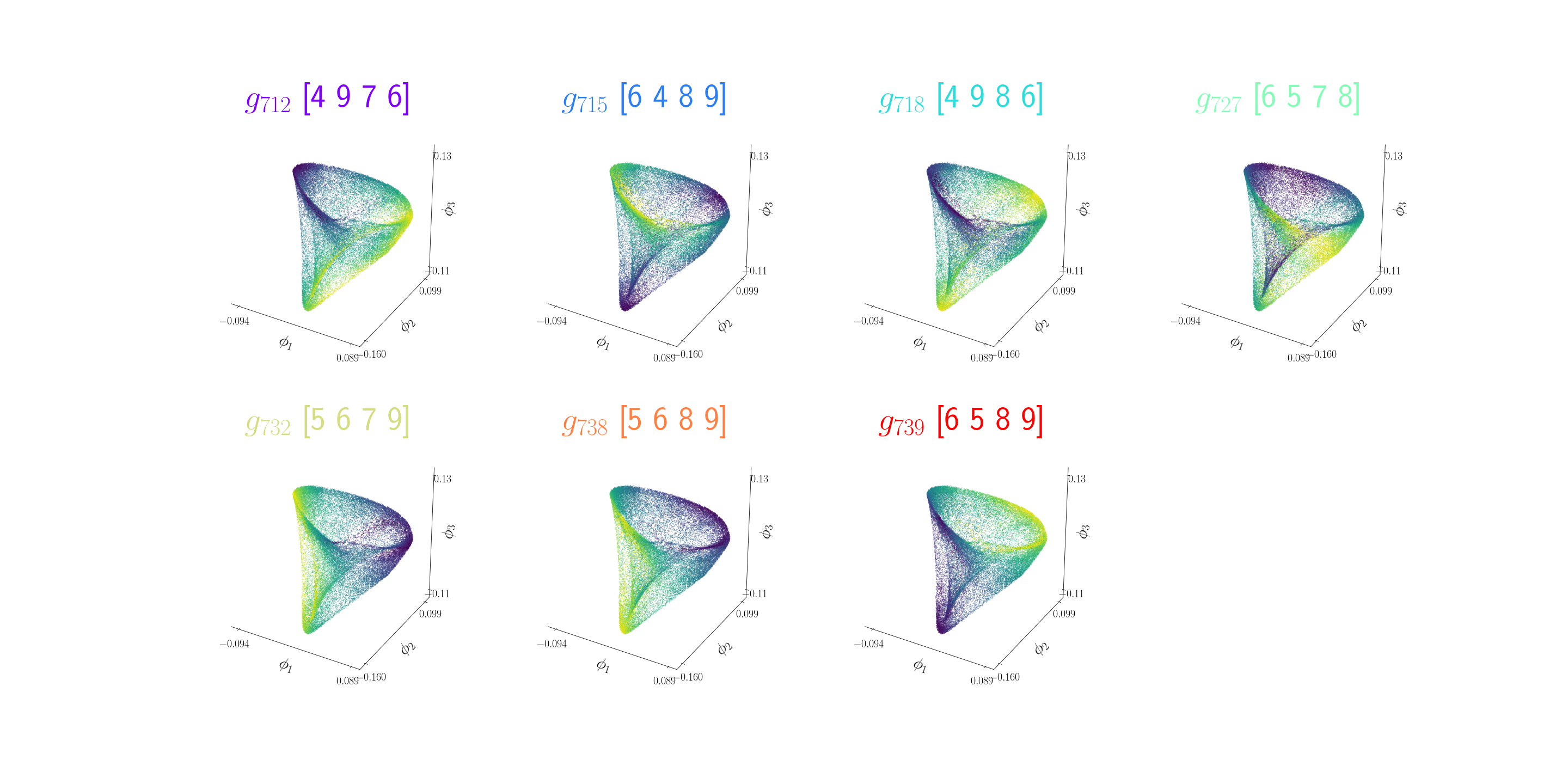}
\caption{\maldata~ support using basis pursuit on superset obtained using \ouralg.  Colors should be compared with Figure \ref{fig:sup-brute}. 
}
\label{fig:mal_full_sup_brute}
\end{figure}

\newpage

\paragraph{Calculated theoretical quantities}

In practice, we do not know theoretical quantities like $\mu$, $\gamma_{\text{max}}$,  $\kappa_S$, and $\min_{i = 1}^n \min_{j' \in S} \|x_{ij}\|$ since we do not have access to $S$.
However, we are able to calculate these quantities using the putative true support.
These are listed in the following table.

\begin{table}[H]
\resizebox{\textwidth}{!}{%
\begin{tabular}{lrrrrrrrr}
\toprule
{} &    $\bar \mu $ &    $ \sigma_\mu $        &        $  \bar {\kappa_S} $&   $      \sigma_{\kappa_S} $&   $       {\gamma_{\text{max}}} $&         $\sigma_{\gamma_{\text{max}}} $&        $ \bar {\min_{i = 1}^{n} \min_{j \in S } ||x_{ij}|| }$ &      $   \sigma_ {\min_{i = 1}^{n} \min_{j \in S } ||x_{ij}|| } $\\
\midrule
\text{Ethanol (a priori)} & $ \sim $1.0 &  8.348332e-08 &  10.029410 &  9.815921 &  44.110696 &  1.825407 &  0.970851 &  0.530601 \\
\text{Malonaldehyde (a priori)} & $ \sim$1.0 &  9.752936e-08 &   2.220002 &  0.771986 &  26.189684 &  0.477759 &  2.709132 &  0.464913 \\
\text{Toluene (a priori)} &   &  &   &  &  15.576112 &  0.407799 &  1.449570 &  0.914012 \\
\text{Ethanol (agnostic)}  &  $ \sim$1.0 &  4.801062e-11 &   4.138372 &  2.113210 &  57.300602 &  1.360244 &  1.932001 &  1.605302 \\
\text{Malonaldehyde (agnostic)}  &  $ \sim$1.0 &  2.016285e-09 &   2.204895 &  0.589812 &  66.019168 &  1.044451 &  3.955157 &  0.853646 \\
\bottomrule
\end{tabular}
}
\caption{Mean and standard deviation of theoretical quantities across replicates}
\label{tab:theory_quantities}
\end{table}
